\pdfoutput=1
\PassOptionsToPackage{hyphens}{url}
\PassOptionsToPackage{table}{xcolor}
\documentclass[]{fairmeta}

\usepackage{url}
\usepackage{amsmath,amssymb}
\DeclareMathOperator*{\argmax}{arg\,max}
\usepackage{amsthm}
\usepackage{algorithm}
\usepackage{algpseudocode}
\definecolor{palo}{HTML}{2E7D4F}
\theoremstyle{plain}
\newtheorem{proposition}{Proposition}
\newtheorem{lemma}{Lemma}
\newtheorem{corollary}{Corollary}
\usepackage{tikz}
\usepackage{tabularx}
\usetikzlibrary{arrows.meta,positioning}
\usepackage{pifont}
\usepackage{wrapfig}
\usepackage{needspace}
\usepackage{xcolor}
\usepackage{colortbl}
\usepackage{capt-of}
\usepackage{bm}
\usepackage{multirow}
\usepackage{titletoc}
\usepackage{fontawesome5}
\usepackage{thmtools, thm-restate}

\definecolor{ink}{HTML}{1A2B2F}
\definecolor{paloalto}{HTML}{4FA36E}
\definecolor{abg}{HTML}{ECE4F3}\definecolor{afg}{HTML}{6A519E}
\definecolor{dbg}{HTML}{D5ECE8}\definecolor{dfg}{HTML}{2E7D6E}
\newcommand{\Aibox}[1]{{\setlength{\fboxsep}{0.6pt}\colorbox{abg}{\textcolor{afg}{#1}}}}
\newcommand{\Dbox}[1]{{\setlength{\fboxsep}{0.6pt}\colorbox{dbg}{\textcolor{dfg}{#1}}}}
\definecolor{peachbg}{HTML}{FBEDE4}
\definecolor{peachbd}{HTML}{D89A76}
\definecolor{beliefbg}{HTML}{EEF6EF}
\definecolor{coral}{HTML}{E76F51}
\definecolor{cardgray}{HTML}{EFF5F9}
\definecolor{icebd}{HTML}{86AFC6}
\definecolor{cream}{HTML}{FDF6E3}
\definecolor{butterbd}{HTML}{CFB25F}
\definecolor{lavbd}{HTML}{9E97CC}

\newtcblisting[auto counter]{promptbox}[3][]{enhanced jigsaw, breakable,
  colback=cardgray, colframe=icebd, boxrule=0.8pt, arc=2.5mm,
  colbacktitle=white, coltitle=ink, fonttitle=\bfseries\small,
  titlerule=0.6pt,
  title={#2},
  title after break={\small\itshape #2 (continued)},
  label={#3},
  left=6pt, right=6pt, top=5pt, bottom=5pt,
  pad at break*=3mm,
  listing only,
  listing options={
    basicstyle=\ttfamily\footnotesize\color{ink},
    columns=fullflexible, keepspaces=true,
    breaklines=true, breakindent=0pt,
    postbreak=\mbox{\textcolor{icebd}{$\hookrightarrow$}\,},
    upquote=true,
    tabsize=4,
    literate={—}{{---}}1 {–}{{--}}1 {•}{{\textbullet\ }}1,
  },
  #1}

\newcommand{\lpred}{\mathcal{L}_{\mathrm{pred}}}
\newcommand{\ytst}{\hat{y}^{\mathrm{tst}}}
\newcommand{\yexe}{\hat{y}^{\mathrm{exe}}}
\newcommand{\rtst}{y^{\mathrm{tst}}}
\newcommand{\rexe}{y^{\mathrm{exe}}}
\newcommand{\Vhat}{\widehat{V}}
\newcommand{\Qhat}{\widehat{Q}}
\newcommand{\Ahat}{\widehat{A}}

\title{ANTE: Anticipate, Then Execute -- Learning Internal World Models for Coding Agents from Verified Beliefs}

\author[1,*]{Yi~Ge}
\author[2,\dagger]{Kangrui~Ruan}
\author[2]{Jifan~Chen}
\author[2]{Manish~Prajapat}
\author[2]{Christian~Bock}
\author[2]{Xiaoyang~Wang}
\author[2]{Rein~Houthooft}
\author[2]{Aritra~Sengupta}
\author[2]{Varun~Kumar}

\affiliation[1]{University of Maryland, College Park}
\affiliation[2]{AWS AI Labs}

\contribution[*]{Work done during an internship at AWS AI Labs}
\contribution[\dagger]{Corresponding author}

\abstract{
Rapid advances in large language models have made software engineering agents capable of tackling complex coding tasks. Within a
containerized environment, a coding agent receives tool responses or
execution results, and determines its later actions by those
earlier results. However, most existing coding agents never look ahead:
their policies do not anticipate the consequences of actions, and
the terminal reward barely attributes a failure to certain
earlier steps, so an ill-judged edit can corrupt the
repository state irreversibly before any signal arrives. Recent  code world models aim to address these challenges, but they are either trained and then frozen before reinforcement learning (RL), or reduced to an auxiliary next-token loss on the environment's outputs. We present ANTE
(Anticipate, Then Execute), an agentic RL framework that verifies the
policy's pre-action predictions by real execution and turns them into
per-step credit for its actions and trust-or-abstain decisions. With an
internal belief stated before each action, Belief-Anchored Policy
Optimization (BAPO) turns the settled predictions into per-step
advantages by anchoring credit on the policy's own forecasts. To turn
the trained forecasts into better actions at inference, thereby
facilitating effective test-time scaling, ANTE-TTO (test-time optimization) selects among
candidate actions without an external scorer. ANTE raises Qwen3-32B from 25.0\% to 57.2\% pass@1 on SWE-bench Verified, 6.2\% above a matched GRPO baseline trained without the belief state; its world model's accuracy climbs from
44\% to 69\% during RL, and the learned foresight capability transfers
across harnesses and out-of-distribution domains. In summary, we aim to build a code world model that knows what has happened, what holds now, and what the next action will achieve.
}

\authoremails{\email{yige@umd.edu}, \email{kangruir@amazon.com}}
\metadata[Project Page]{\url{https://ante-agent.com}}

\begin{document}

\maketitle

\section{Introduction}
\label{sec:intro}

\begin{wrapfigure}{r}{0.5\textwidth}
\centering
\includegraphics[width=\linewidth]{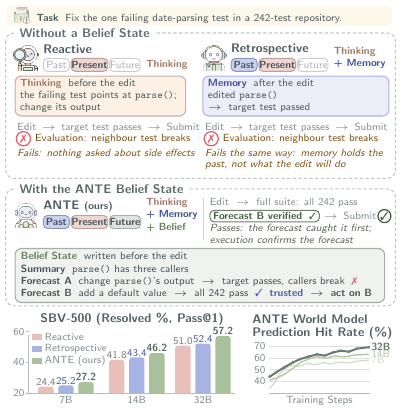}
\caption{\textbf{Anticipation, verified by execution.} Top: the same
task with and without a belief state; only ANTE forecasts the edit's
effect and lets execution settle it before committing. Bottom: under RL,
resolve rate and prediction hit rate rise together.}
\label{fig:intro}
\end{wrapfigure}

Large language models (LLMs) now act as software engineering agents: given a
repository issue or a terminal task, they edit, run, and iterate
inside a container until tests validate the result
\citep{jimenez2024swebench,pan2024swegym,jain2025r2egym}.
Open agents are trained by supervised fine-tuning (SFT) on executed
trajectories \citep{yang2025swesmith} and by reinforcement learning (RL) against that terminal
test \citep{song2026swemaster}, at growing scale. Both SFT and RL are trained on tool-use trajectories, in which the agent observes each tool result and uses it to choose the next action \citep{luo2025deepswe,kimi2025kimidev}. However, outcome-based RL rewards only the final task result \citep{shao2024deepseekmath,fan2025klear,ding2026empowering}; it does not directly train the agent to predict what each action will do. Without anticipating an action's consequences, an agent may solve the immediate problem while introducing errors elsewhere in the repository. Figure~\ref{fig:intro} illustrates this limitation: both the reactive agent (thinking-only) and the retrospective agent (thinking with memory) submit an edit that passes the target test but breaks a previously passing test. The memory records what happened, but neither agent anticipates the edit's side effects before acting.

To anticipate such failures, an agent needs to predict an action's outcome before executing it. A world model provides this prediction, which can then be checked against the execution result, such as whether a test will pass. The observed outcome then serves as a training target, helping the world model make more accurate predictions. Existing approaches learn from execution traces or auxiliary observation losses \citep{meta2025cwm,shrivastava2026echo,lu2026paw}, or refine simulated plans \citep{yu2025dynamind}, but do not turn verified forecasts into per-step policy credit. Other approaches rely on learned simulators or language-model judges to evaluate predictions \citep{sun2026sweworld,wang2025vagen}. Execution-verified reasoning has been explored within individual programs \citep{ni2024next,scr2026,tang2026execverify}, rather than repository-level agent trajectories. The remaining gap is to combine explicit forecasts, verification by real execution, and per-step credit within agentic RL.

We introduce \textbf{ANTE (Anticipate, Then Execute)}, an agentic RL framework that trains coding agents to predict action outcomes and learn from execution feedback. Before each action, the policy writes a \emph{belief state} containing a brief state summary, an outcome prediction, and a \emph{trust} or \emph{abstain} decision. Our code world model maintains an internal belief about what the agent has established, what the code does now, and what the next command will change. 
During RL, ANTE utilizes Belief-Anchored Policy Optimization (BAPO), which compares steps across rollouts that predict the same outcome and assigns per-step credit based on their verified forecasts and trust decisions. At inference, ANTE-TTO samples alternative actions when the policy abstains or predicts failure, ranks the candidates by their own forecasts, and executes the highest-ranked action without an external scorer. 

Figure~\ref{fig:intro} illustrates how this helps: unlike the reactive and retrospective agents, ANTE anticipates that the proposed edit would break other tests, chooses an alternative fix, and confirms that the full test suite passes. This example illustrates the behavior we seek: anticipating an action's consequences rather than reacting to them afterward. 

Our main contributions are: \ding{182} To our knowledge, ANTE is the pioneering agentic RL framework in the code domain that verifies the policy's explicit forecasts against real execution throughout RL. These verified forecasts provide step-level supervision for learning an internal world model within the policy. \ding{183} To
turn verified predictions into per-step credit, we propose BAPO. It groups steps across rollouts by their predicted outcomes and compares their execution feedback, allowing individual steps to receive credit beyond the final task reward. \ding{184} We also propose ANTE-TTO, which uses the learned forecasts to choose among candidate actions before execution at inference time. 
It ranks candidates by their predicted outcomes and executes only the highest-ranked action. 
At 7B scale, it improves task success rate with approximately 31\% additional inference tokens over a single greedy attempt. \ding{185} Our experiments show that ANTE improves both prediction accuracy and task performance across tasks and harnesses. ANTE-32B reaches 57.2\% on SWE-bench Verified \citep{jimenez2024swebench}, outperforming all compared open-weight agents above 30B. ANTE-9B reaches 24.7\% on Terminal-Bench 2.0 \citep{merrill2026terminalbench}. In addition, across eleven out-of-distribution benchmarks, ANTE-32B improves average performance by 6.2 percentage points over a matched RL baseline trained without the belief state.

\makeatletter
\def\paragraph{\@startsection{paragraph}{4}{\z@}{0.6ex plus
0.3ex minus .2ex}{-1em}{\normalsize\bf}}
\makeatother
\section{Related Work}
\label{sec:related}

\paragraph{\bfseries World models for coding agents.}
Code world models could help coding agents anticipate the consequences of their actions. CWM \citep{meta2025cwm} learns to predict program execution during mid-training, while its RL stage does not reward or check these predictions. SWE-World and PatchWorld \citep{sun2026sweworld,bai2026patchworld} instead model the environment outside the policy, as a learned simulator or an executable program. ECHO and PaW \citep{shrivastava2026echo,lu2026paw} add an auxiliary loss on the environment's outputs during RL, yet the policy still acts without first stating a forecast. In contrast, ANTE has the policy forecast each action's outcome before acting, verifies the forecast by real execution throughout RL, and uses it for per-step credit and, at test time, for ranking candidate actions.

\paragraph{\bfseries Structured beliefs in agentic RL.}
Prior work rewards an agent's beliefs during RL by comparing them with the true state, through a language-model judge \citep{wang2025vagen,singh2026brace} or embedding similarity \citep{yu2026rwml}. However, such comparisons only approximate correctness, and judge-based rewards are prone to reward hacking \citep{wang2025vagen,yu2026rwml}. T3 \citep{zou2026t3} keeps the belief implicit and truncates rollouts once proxy signals show that it has stalled, leaving the belief itself unsupervised. ANTE grounds the belief in verifiable action outcomes, which makes the supervision more robust: real execution checks each forecast and the decision to trust it.

\paragraph{\bfseries Credit assignment and execution-grounded reasoning.}
Deciding which steps across rollouts are comparable is central to per-step credit assignment. GiGPO \citep{feng2025gigpo} compares actions taken from the same environment state, which requires identical states to recur across rollouts. ReBel \citep{tang2026rebel} groups steps whose beliefs agree on a set of predefined predicates, while PRIME \citep{cui2025prime} instead learns an implicit process reward model that gives dense rewards. Code-reasoning methods use execution traces within individual programs rather than repository-level agent trajectories \citep{scr2026,tang2026execverify,ni2024next}. Our proposed BAPO also compares steps that make the same forecast in different rollouts of the task.

\section{ANTE: Learning to Predict Before Acting}
\label{sec:method}

\begin{figure}[t]
\centering
\includegraphics[width=1.0\textwidth]{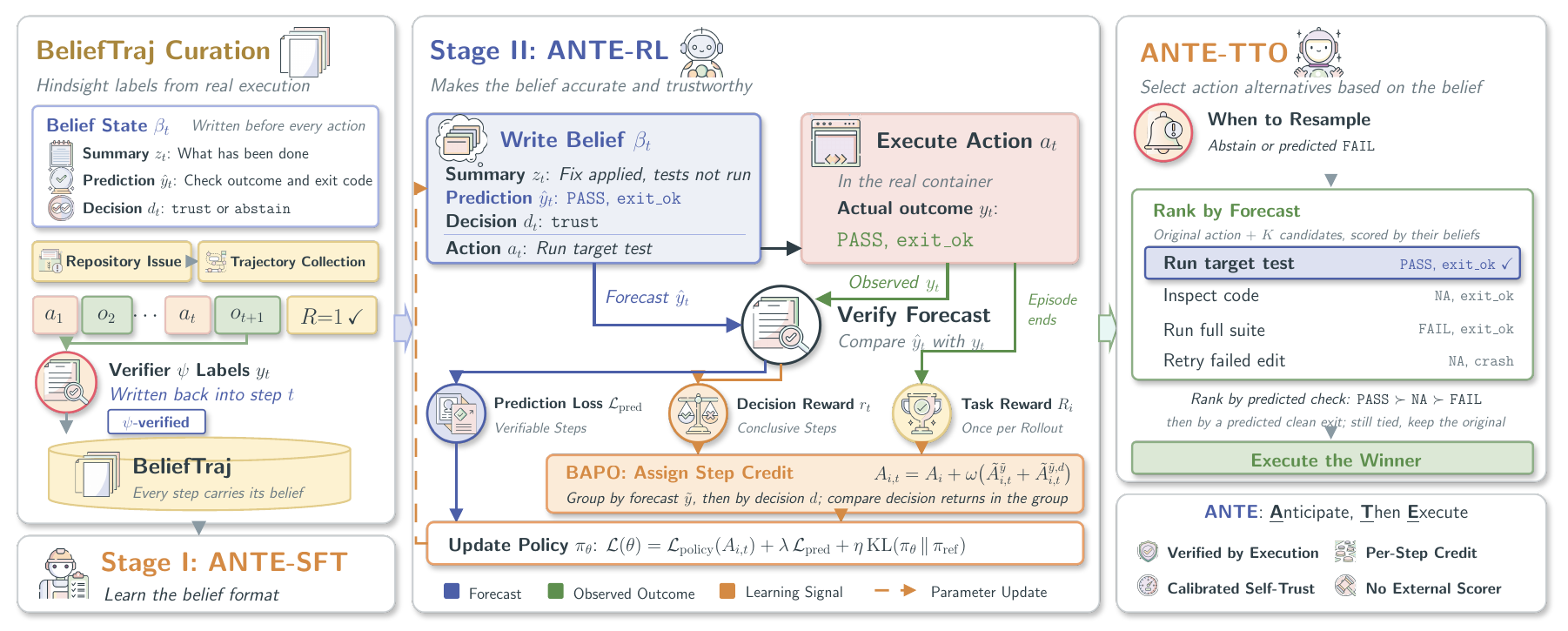}
\caption{\textbf{Overview of ANTE.} In Stage~I, the model learns the belief state from execution-labeled BeliefTraj; Stage~II verifies every forecast against execution and turns it into a prediction loss, a decision reward, and BAPO step credit; ANTE-TTO resamples at an abstention or a predicted \textsc{fail} and executes the best-forecast candidate.}
\label{fig:pipeline}
\end{figure}

\subsection{Problem Formulation}
\label{sec:method:belief}

We formalize a coding agent working on a task inside its container as a partially-observable Markov decision process (POMDP) \citep{astrom1965optimal,kaelbling1998planning} whose latent state includes the repository (Appendix~\ref{sec:prelim}). 
Figure~\ref{fig:pipeline} provides an overview of the ANTE pipeline.
The agent can predict the outcome $y_t$ of the check that its next action runs, and execution verifies that prediction. A deterministic verifier $\psi$ reads it from the execution result as a pair in $\mathcal{Y} = \mathcal{Y}^{\mathrm{tst}} \times \{0,1\}$. The first field is the check's result, \textsc{pass}, \textsc{fail}, or \textsc{na}; the second records whether the command exited cleanly. ANTE's belief is a point forecast of this outcome, written before the tool call.

\paragraph{Belief state.}
\label{def:belief}
Before every action, the policy explicitly articulates what it believes. 
The belief state $\beta_{i,t} = (z_{i,t}, \hat y_{i,t}, d_{i,t})$ has three parts: a one-sentence summary $z_{i,t}$, a forecast $\hat y_{i,t} = (\ytst_{i,t}, \yexe_{i,t}) \in \mathcal{Y}$, and a decision $d_{i,t} \in \{\textsc{trust}, \textsc{abstain}\}$.
The summary $z_{i,t}$ records what is established and what is broken. The forecast takes its values in the outcome space of the verifier, so that it is a statement the harness can confirm or refute. The decision $d_{i,t}$ is a two-level
certainty label in the sense of selective prediction
\citep{davidov2026quit,wu2026bas}. \textsc{trust} commits to the forecast. \textsc{abstain} still writes the forecast and still takes the action, but does not commit to it.
Intuitively, the decision is needed because prediction accuracy alone would reward guessing wherever the repository state is underdetermined. 
A confident forecast of failure is a legitimate \textsc{trust}. Within one autoregressive pass the step
message factorizes into reasoning, summary, prediction, decision, and
action factors,
\begin{equation}
\begin{aligned}
\pi_\theta(u_{i,t} \mid h_{i,t}) \;=\;&\;
\pi_\theta(v_{i,t} \mid h_{i,t})\,
\pi_\theta(z_{i,t} \mid h_{i,t}, v_{i,t})\,
\pi_\theta(\hat y_{i,t} \mid c_{i,t})\\[-2pt]
&\times\;
\pi_\theta(d_{i,t} \mid c_{i,t}, \hat y_{i,t})\,
\pi_\theta(a_{i,t} \mid h_{i,t}, v_{i,t}, \beta_{i,t}),
\end{aligned}
\label{eq:factor}
\end{equation}
where $h_{i,t}$, $v_{i,t}$, and $a_{i,t}$ are the history, the reasoning, and the tool call, respectively, $c_{i,t} := (h_{i,t}, v_{i,t}, z_{i,t})$ is the context in which the forecast is written, and $\pi_\theta(y \mid c_{i,t})$ is the probability the prediction factor assigns to outcome $y$.

\paragraph{Verifiable and conclusive steps.}
\label{def:outcome}
\label{def:chi-psi}
A forecast can be scored only when the action runs something the verifier can read. Before execution, a deterministic action-type map $\chi : \mathcal{A} \to
\{\textsc{epi}, \textsc{prag}\}$ sorts the steps. A step is \emph{epistemic} if its action runs a check the verifier can evaluate, such as a target test or a reproduction script, and \emph{pragmatic} otherwise. After execution, the verifier
$\psi : \mathcal{O} \times \{\textsc{epi}, \textsc{prag}\} \to
\mathcal{Y} \cup \{\bot\}$ evaluates the step. The check field $\rtst_{i,t}$ records the result of the check that an epistemic step ran, or \textsc{na} when the response does not determine one. The exit field $\rexe_{i,t}$ records whether the command ran cleanly. The
realized outcome is $y_{i,t} := \psi(o_{i,t+1}, \chi(a_{i,t}))$. A step is verifiable when its belief block parses and $\psi$ returns a value ($m_{i,t} = 1$). It is \emph{conclusive} when in addition $\rtst_{i,t} \in \{\textsc{pass}, \textsc{fail}\}$; on repository issues these are the verifiable epistemic steps whose check produced a result. We write $\mathcal{C}_i := \{t \in [T_i] : m_{i,t} = 1,\ \rtst_{i,t} \in \{\textsc{pass}, \textsc{fail}\}\}$ for the conclusive steps of a rollout and $\mathcal{C} := \{(i, t) : t \in \mathcal{C}_i\}$ for those of the task.

\subsection{Stage~I (ANTE-SFT): Supervised Fine-Tuning on BeliefTraj}
\label{sec:method:corpus}
\label{sec:method:sft}

Our training procedure consists of two stages: supervised fine-tuning (SFT) followed by reinforcement learning (RL).
In Stage~I, we teach the policy to generate structured
belief states alongside its reasoning and actions through SFT
on BeliefTraj, a corpus of solving trajectories collected
by a capable agent in the execution environment.
Each step is annotated with a belief state $\beta_t = (z_t, \hat y_t, d_t)$.
The forecast $\hat y_t$ is the outcome the verifier $\psi$ reads from the logged next observation. 
The decision $d_t$ is a label assigned when the corpus is built. 
Stage~II subsequently refines the policy's trust/abstain behavior through decision rewards.
Construction, decision labeling, the leakage filter, and statistics are in Appendix~\ref{app:corpus}.
Stage~I jointly optimizes all policy factors in Eq.~\ref{eq:factor} by maximum likelihood on the corpus $\mathcal{D}_{\mathrm{BT}}$, $\mathcal{L}_{\mathrm{ANTE\text{-}SFT}}(\theta) = -\,\mathbb{E}_{\tau \sim \mathcal{D}_{\mathrm{BT}}} \sum_{t} \log \pi_\theta(u_t \mid h_t)$; observation and prompt tokens carry no loss. After Stage~I, belief is part of the policy's normal output, and the prediction factor is correct in hindsight on corpus histories.
Building on prior work on imitation learning and SFT \citep{gangwani2020bmil,ruan2024causal,chu2025sft}, we formulate Stage~I as belief-augmented imitation in a POMDP and derive the following bound:

\begin{restatable}[Belief-augmented imitation bound]{proposition}{propbeliefimitation}
\label{prop:imitation}
Let $P_E$ be the law of BeliefTraj trajectories over the corpus task
distribution, and let $\pi_\theta$ act in the real environment on the same
tasks. Write $J$ for the task success probability, $T$ for the step budget, and $H_E := \sum_{t=1}^{T} \mathbb{E}_{P_E}\big[H(P_E(\cdot \mid
h_t))\big]$ for the conditional entropy of the demonstration process's
step messages $u_t = (v_t, \beta_t, a_t)$, reasoning included,
$\varepsilon_{\mathrm{SFT}}(\theta) :=
\mathcal{L}_{\mathrm{ANTE\text{-}SFT}}(\theta) - H_E \ge 0$ for the excess Stage~I
loss, and $\Lambda := \sum_{t=1}^{T} \mathbb{E}_{P_E}\Big[\mathrm{TV}\big(P_E(o_{t+1} \mid h_t, u_t),\; P_{\mathrm{env}}(o_{t+1} \mid h_t, a_t)\big)\Big]$
for the non-causal part of the demonstration process. Then 
\begin{equation}
J(P_E) - J(\pi_\theta) \le \sqrt{T\,\varepsilon_{\mathrm{SFT}}(\theta)/2} + \Lambda .
\end{equation}
\end{restatable}
Proposition~\ref{prop:imitation} states what Stage~I guarantees. Minimizing the cross-entropy drives the first term down, so the fine-tuned policy approaches the success rate of the demonstrations up to $\Lambda$. This remainder comes from the labels written in hindsight, which imitation cannot recover. Stage~II addresses the hindsight-induced mismatch captured by $\Lambda$.

\subsection{Stage~II (ANTE-RL): Belief-Anchored Policy Optimization}
\label{sec:method:objective}
\label{sec:method:bapo}

In Stage~II, the policy learns from its own rollouts in the execution environment, using execution feedback to refine its forecasts and decisions. These rollouts provide step-level execution outcomes and episode-level task rewards, which support a joint objective combining prediction supervision with belief-anchored policy optimization (BAPO).

\paragraph{RL rollouts.}
\label{def:rollout}
For each training task $x \sim \mathcal{D}$, the policy $\pi_\theta$ produces $N$ rollouts $\tau_1, \dots, \tau_N \sim \pi_\theta(\cdot \mid x)$. At step $t$ of rollout $i$, the policy reads the history $h_{i,t} := (x,\ u_{i,1}, o_{i,2},\ \dots,\ u_{i,t-1}, o_{i,t})$ and emits one message $u_{i,t} = \langle v_{i,t}, \beta_{i,t}, a_{i,t} \rangle$: its reasoning, a belief state, and one tool call. The environment replies with the observation $o_{i,t+1}$. A rollout ends after $T_i$ steps.

Given these rollouts, the prediction loss supervises forecasts with execution-derived labels, while the decision reward scores trust/abstain decisions on conclusive steps.
The policy-gradient update supplements episode-level task credit with belief-anchored adjustments on eligible steps, computed from decision returns within forecast and forecast--decision groups. The decision reward and belief-anchored advantage construction together form BAPO.
The policy gradient trains the reasoning, decision, and action tokens of Eq.~\ref{eq:factor}; the prediction loss alone trains the two prediction tokens; the summary tokens receive no gradient, since nothing checks a summary. The task reward, received once per rollout, is $R_i :=
\mathbf{1}[\tau_i \text{ resolves } x] + r^{\mathrm{fmt}}_i$ with a
bounded bookkeeping term for format and length
(Appendix~\ref{app:hparams}).
To motivate Stage~II's prediction supervision, we consider a selector that ranks candidate actions by their predicted expected utility. The following proposition bounds its one-step selection regret in terms of candidate-specific forecast divergences.

\begin{restatable}[Prediction-to-decision regret bound]{proposition}{proppredregret}
\label{prop:selection}
Let candidates $k = 0, \dots, K$ at one step have outcome laws $p_k$
under execution and belief predictives $q_k$, with $\delta_k :=
\mathrm{KL}(p_k \,\|\, q_k)$ the divergence of each forecast from its
executed outcome law. For a score $u : \mathcal{Y} \to [0, U]$, let
$k^\ast$ maximize the executed value $v_k := \mathbb{E}_{p_k}[u]$ and
$\hat k$ the believed value $\hat v_k := \mathbb{E}_{q_k}[u]$. Then
\begin{equation}
v_{k^\ast} - v_{\hat k} \le U\big(\sqrt{\delta_{k^\ast}/2} + \sqrt{\delta_{\hat k}/2}\big) \le U \sqrt{2 \max_k \delta_k}.
\end{equation}
\end{restatable}
Proposition~\ref{prop:selection} shows that smaller candidate-specific forecast divergences tighten an upper bound on selection regret. Detailed proofs can be found in Appendix~\ref{app:proofs}.

\paragraph{Execution-grounded prediction loss.} 
For a fixed conditional outcome distribution, expected forecast cross-entropy equals KL divergence plus an entropy term independent of the predictive distribution. This motivates supervising forecasts with the realized outcomes of the policy's own actions.
The prediction factor occupies the value-token positions $\mathcal{V}_{i,t}$ of a step. The reasoning, decision, and action factors occupy the positions $\mathcal{K}_{i,t}$. Both losses are averaged over the same token count $Z := \sum_{i,t} |\mathcal{K}_{i,t}|$. Every verifiable step receives a cross-entropy against the realized outcome at its value positions,
\begin{equation}
\lpred(\theta) \;=\; -\,\mathbb{E}_{x \sim \mathcal{D}}\;
\mathbb{E}_{\{\tau_i\}_{i=1}^{N} \sim \pi_\theta(\cdot \mid x)}
\left[ \frac{1}{Z} \sum_{i=1}^{N} \sum_{t=1}^{T_i} m_{i,t}\,
\log \pi_\theta\big(y_{i,t} \,\big|\, c_{i,t}\big) \right].
\label{eq:lpred}
\end{equation}

\paragraph{Decision reward.}
The decision reward encourages accurate forecast commitments and penalizes incorrect ones.
On conclusive steps, the scoring rule $\sigma$ (Eq.~\ref{eq:settle}) assigns a reward of $+\alpha$ to a trusted forecast that is correct in both fields with a clean exit, and a penalty of $-\gamma$ to an incorrect trusted forecast. Abstention incurs a fixed cost $\kappa$, contributing $-\kappa$ to the reward.
We define the decision reward of step $t$ as $r_{i,t} := \sigma(d_{i,t}, \hat y_{i,t}, y_{i,t})$ for $t \in \mathcal{C}_i$ and $r_{i,t} := 0$ otherwise, and its decision return as $W_{i,t} := \sum_{s \ge t} r_{i,s}$, the sum of the rewards from step $t$ on. We set $\gamma = \alpha q^{\ast}/(1 - q^{\ast})$, so that trusting a forecast has positive expected reward only when the forecast is right with probability above the target accuracy $q^{\ast}$. 
As its forecasts improve, the policy therefore drops the corpus's abstention before its accuracy reaches the target.

\begin{figure}[t]
\centering
\includegraphics[width=1.0\textwidth]{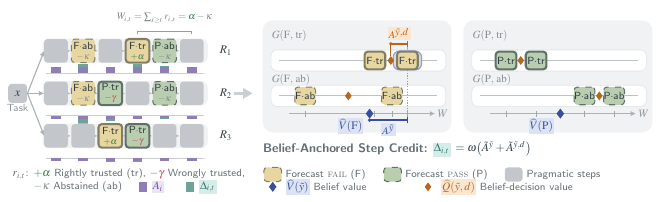}
\caption{\textbf{BAPO credit.} Left: three rollouts of one task; a
conclusive step shows its forecast (F or P), its decision (tr or ab),
and its decision reward $r_{i,t}$; a bracket marks one decision return
$W_{i,t}$, and the bar under every step is its credit $A_{i,t} =
\Aibox{$A_i$} + \Dbox{$\Delta_{i,t}$}$ (Eq.~\ref{eq:bapo-adv});
pragmatic steps carry $A_i$ alone. Right: each white strip is a
belief-decision group $G(\tilde y, d)$ inside its belief group
$G(\tilde y)$; for the ringed step, the gap to the belief value is its belief-relative advantage and the gap to the belief-decision value its decision-relative advantage.}
\label{fig:bapo-overview}
\end{figure}

\Needspace*{16\baselineskip}
\begin{wrapfigure}[12]{r}{0.5\textwidth}
\begin{minipage}{\linewidth}
\begin{algorithm}[H]
\caption{ANTE-RL: BAPO iteration}
\label{alg:bapo-compact}
\scriptsize
\begin{algorithmic}[1]
\Require $\pi_\theta$ (Stage~I), $\pi_{\mathrm{ref}}$, $N$, $\psi$, $\chi$, $(\alpha,\gamma,\kappa)$, $\omega$, $\lambda$, $\eta$, $\eta_{\mathrm{lr}}$
\Statex \textcolor{palo!55!gray}{\textit{// Rollout with the belief}}
  \State $x\!\sim\!\mathcal{D}$;\, $\tau_{1:N}\!\sim\!\pi_\theta(\cdot\mid x)$, each step writing $\beta_{i,t}$, then acting
  \State $y_{i,t}\gets\psi(o_{i,t+1},\chi(a_{i,t}))$; mark $m_{i,t}$, $\mathcal{C}_i$
\Statex \textcolor{palo!55!gray}{\textit{// Settle the forecasts}}
  \State $r_{i,t}\gets\sigma(d_{i,t},\hat y_{i,t},y_{i,t})$ on $\mathcal{C}_i$, else $0$ \Comment{Eq.~\ref{eq:settle}}
  \State $W_{i,t}\gets\sum_{s\ge t}r_{i,s}$;\, $A_i\gets R_i-\tfrac{1}{N}\sum_j R_j$
\Statex \textcolor{palo!55!gray}{\textit{// Belief-anchored credit}}
  \State $\Vhat,\Qhat\gets$ means of $W$ on $G(\tilde y)$, $G(\tilde y,d)$ \Comment{Eq.~\ref{eq:values}}
  \State $A_{i,t}\gets A_i+\Delta_{i,t}$ \Comment{Eq.~\ref{eq:bapo-adv}}
\Statex \textcolor{palo!55!gray}{\textit{// One update; full pipeline in Alg.~\ref{alg:bapo}}}
  \State $\theta\gets\theta-\eta_{\mathrm{lr}}\nabla_\theta\mathcal{L}(\theta)$ \Comment{Eqs.~\ref{eq:lpred},~\ref{eq:objective}}
\end{algorithmic}
\end{algorithm}
\end{minipage}
\end{wrapfigure}

\paragraph{Belief-anchored advantages.}
Crediting a step on its own needs an anchor to compare it against, and the existing approaches \citep{scr2026,feng2025gigpo} are not readily applicable in our setting.
BAPO instead uses the policy's predicted outcome as a recurring grouping key, comparing steps with the same forecast across rollouts of the same task, regardless of trajectory position (Figure~\ref{fig:bapo-overview}).
The episode advantage is
mean-only \citep{liu2025drgrpo}, $A_i := R_i - \tfrac{1}{N}\sum_{j=1}^{N}
R_j$. An anchor value $\tilde{y}$ is \textsc{pass} or \textsc{fail} on repository issues. We define the belief group of $\tilde{y}$ as $G(\tilde{y}) := \{(j, s) \in \mathcal{C} : \ytst_{j,s} = \tilde{y}\}$, excluding the trusted crash forecasts that score zero above, its belief-decision groups as $G(\tilde{y}, d) := \{(j, s) \in G(\tilde{y}) : d_{j,s} = d\}$ for $d \in \{\textsc{trust}, \textsc{abstain}\}$, and the anchored steps as $\mathcal{G} := G(\textsc{pass}) \cup G(\textsc{fail})$. The key is
the predicted check field on repository issues and the predicted exit
field on the terminal harness, where the check is evaluated once per
rollout. The resulting baselines are conditional on the belief: trusting a predicted \textsc{fail} and trusting a predicted \textsc{pass} have different values, and a single global baseline would mix them. For an anchored step $(i,t) \in \mathcal{G}$ with predicted key field $\tilde y$ and decision $d := d_{i,t}$, the \emph{belief value},
the \emph{belief-decision value}, and the \emph{belief-relative} and
\emph{decision-relative advantages} are
\begin{equation}
\begin{aligned}
\Vhat(\tilde{y}) \;&:=\; \mathbb{E}_{G(\tilde{y})}[W],
&\qquad A^{\tilde y}_{i,t} \;&:=\; W_{i,t} - \Vhat(\tilde{y}),\\
\Qhat(\tilde{y}, d) \;&:=\; \mathbb{E}_{G(\tilde{y}, d)}[W],
&\qquad A^{\tilde y, d}_{i,t} \;&:=\; W_{i,t} - \Qhat(\tilde{y}, d),
\end{aligned}
\label{eq:values}
\end{equation}
where $\mathbb{E}_F$ is the uniform mean over a set $F$ of steps, and $\Vhat$ and $\Qhat$ play the role of a value and an action value, computed on belief groups instead of on states. The anchor and the decision are the policy's own output. With the group values and the anchor groups held fixed, subtracting $\Vhat$ or $\Qhat$ is therefore an ordinary baseline for the tokens written after them. For the reasoning tokens, each stream adds a shaping term bounded by the range of its group values times the expected score norm (Appendix~\ref{rem:anchor-bias}). The two streams are standardized within the task, with the divisor floored at the reward scale $\alpha + \gamma$; the standardized advantages are written $\tilde A$. A tilde on $y$ marks a group key, and a tilde on $A$ marks standardization. The combined step term is clipped, a degenerate group falls back to the task mean, and a task with no trusted conclusive step gets no step term. Before that clipping and those fallbacks, BAPO assigns
\begin{equation}
A_{i,t} := A_i + \Delta_{i,t},\qquad
\Delta_{i,t} := \omega\big(\tilde A^{\tilde y}_{i,t} + \tilde
A^{\tilde y,d}_{i,t}\big)\ \text{for } (i,t) \in \mathcal{G},\qquad
\Delta_{i,t} := 0\ \text{otherwise},
\label{eq:bapo-adv}
\end{equation}
with $\omega = \tfrac{1}{2}$. Before standardization, the step term is the decision return minus one baseline, $\omega(A^{\tilde y} + A^{\tilde y,d}) = W_{i,t} - \tfrac{1}{2}\big(\Vhat(\tilde y) + \Qhat(\tilde y, d)\big)$: the belief-decision value shrunk halfway toward the belief value, which stabilizes it. This equals $A^{\tilde y,d}_{i,t} + \tfrac{1}{2}\Ahat(\tilde y, d)$ with $\Ahat := \Qhat - \Vhat$ the decision advantage: the step's deviation within its decision group counts at full weight, and the credit for the decision itself at half (Proposition~\ref{prop:decomp}).

\paragraph{RL objective.}
The three signals combine into one objective. For the $k$-th token $w_{i,t,k}$ of step $t$, the importance ratio $\varrho_{i,t,k}(\theta) := \pi_\theta(w_{i,t,k} \mid h_{i,t}, w_{i,t,<k}) \,/\, \pi_{\mathrm{old}}(w_{i,t,k} \mid h_{i,t}, w_{i,t,<k})$ compares the current policy with $\pi_{\mathrm{old}}$, the policy that generated the batch. With the expectation over $x \sim \mathcal{D}$ and $\tau_{1:N} \sim \pi_{\mathrm{old}}(\cdot \mid x)$, the training objective is
\begin{equation}
\begin{aligned}
\mathcal{L}(\theta) \;&=\; -\,
\mathbb{E}_{x,\,\tau_{1:N}}
\!\left[ \frac{1}{Z} \sum_{i=1}^{N} \sum_{t=1}^{T_i}
\sum_{k \in \mathcal{K}_{i,t}} \ell_{i,t,k}(\theta) \right]
\;+\; \eta\, \widehat{D}_{\mathrm{KL}}\!\left(\pi_\theta \,\|\,
\pi_{\mathrm{ref}}\right) \;+\; \lambda\, \lpred(\theta),\\
\ell_{i,t,k}(\theta) \;&:=\;
\min\!\Big(\varrho_{i,t,k}\, A_{i,t},\
\mathrm{clip}\big(\varrho_{i,t,k},\, 1{-}\epsilon_{\mathrm{lo}},\,
1{+}\epsilon_{\mathrm{hi}}\big)\, A_{i,t}\Big).
\end{aligned}
\label{eq:objective}
\end{equation}
Here $\eta$ and $\lambda$ weight the KL and prediction terms, $(\epsilon_{\mathrm{lo}}, \epsilon_{\mathrm{hi}})$ is the asymmetric clip \citep{yu2025dapo}, and $\widehat{D}_{\mathrm{KL}}$ is the $k_3$ estimator \citep{schulman2020kl} anchored at the frozen warm start $\pi_{\mathrm{ref}}$. Reasoning tokens receive gradient from both terms. The policy term weights them by $A_{i,t}$, and $\lpred$ back-propagates through them as part of $c_{i,t}$. Through these shared parameters, the prediction loss shapes the representation that the action is conditioned on \citep{gangwani2020bmil}. Algorithm~\ref{alg:bapo} gives the full pipeline; notation, outcome channels, implementation rules, and prompts are in Appendices~\ref{app:notation}, \ref{app:format}, \ref{app:hparams}, and~\ref{app:prompts}.

\subsection{ANTE-TTO: Action Selection by the World-Model Prediction}
\label{sec:method:validator}

The trained foresight can also be used at inference. The forecast and the decision already tell which candidate the policy expects to succeed, so test-time foresight needs neither the extra executions of best-of-$n$ \citep{cobbe2021verifiers} nor a second model. When the policy abstains or predicts \textsc{fail}, ANTE-TTO samples $K$ alternative messages and ranks them together with the original by their own beliefs. The score is $s(\beta) = \phi(\ytst) + \mathbf{1}[\yexe = 1]$ with $\phi = 3, 1, 0$ for \textsc{pass}, \textsc{na}, and \textsc{fail}. It executes the highest-ranked candidate, keeps the original on ties, and changes no weights (Algorithm~\ref{alg:tto}).
Its regret follows from Proposition~\ref{prop:selection} with $K + 1$ candidates, plus the spread of a point forecast at uncertain steps (Corollary~\ref{cor:pointsel}).

\makeatletter
\def\paragraph{\@startsection{paragraph}{4}{\z@}{0.6ex plus
0.3ex minus .2ex}{-1em}{\normalsize\bf}}
\makeatother
\setlength{\textfloatsep}{9pt plus 2pt minus 3pt}
\setlength{\floatsep}{6pt plus 2pt minus 2pt}
\setlength{\intextsep}{6pt plus 2pt minus 2pt}
\section{Experiments}
\label{sec:experiments}

\subsection{Experimental Setup}
\label{sec:exp:setup}

\paragraph{Model and Environment.}
We train and evaluate ANTE on two agent harnesses: repository issues under R2E-Gym \citep{jain2025r2egym} at 7B, 14B, and 32B, and terminal tasks at 9B with the Tmax agent~\citep{ivison2026tmax}. The 7B and 14B models build on Qwen2.5-Coder-Instruct \citep{hui2024qwen25coder}, the 32B model on Qwen3-32B \citep{yang2025qwen3}, and the 9B model on Qwen3.5-9B \citep{qwen2026qwen35}. The belief schema and the advantage are the same on both harnesses; the outcome space, the anchor key, and the decision reward are set per harness (\S\ref{sec:method:belief}, Appendix~\ref{app:terminal-config}).

\paragraph{Data and Comparisons.}

SFT uses paired corpora containing 12,500 verified successful trajectories over 6,729 repository tasks and 30,161 over 10,823 terminal tasks from the tmax-15k pool \citep{ivison2026tmax}. 
We compare ANTE with two controlled baselines: \textit{Reactive}, which follows a conventional agent-training paradigm in which the policy interleaves thinking and tool calls; and \textit{Retrospective}, which augments \textit{Reactive} with an explicit memory state summarizing the trajectory so far.
Neither includes ANTE’s outcome forecasts or trust/abstain decisions. Their paired SFT corpora share source trajectories and tool calls. Each variant then generates fresh on-policy rollouts from its own SFT checkpoint using matched RL pools of 6,566 repository tasks and 3,558 terminal tasks. To ablate BAPO, we additionally evaluate ANTE (w/o BAPO), which retains the belief state and prediction loss but uses episode-level policy advantages without decision rewards or belief-anchored step adjustments.

\paragraph{Evaluation.}
We evaluate on SWE-bench Verified \citep{jimenez2024swebench}, on three terminal benchmarks (TB-Dev, TB-Lite \citep{openthoughts2026tblite}, and Terminal-Bench 2.0 \citep{merrill2026terminalbench}), and on eleven out-of-distribution reasoning benchmarks. The RL baselines train the same corrected group-relative objective in the same setting, with the belief components removed. Each starts from the SFT checkpoint of its own corpus variant, so every matched pair isolates the belief. ANTE-TTO also runs one trajectory per task: at each trigger step it samples $K = 3$ alternatives at temperature 1.0, scores them by their beliefs, and executes only the winner (\S\ref{sec:method:validator}, Appendix~\ref{app:tto}). All other entries are single greedy attempts (temperature 0). Implementation details and run-to-run variation are in
Appendices~\ref{app:hparams}, \ref{app:concurrent}, and~\ref{app:term}.

\subsection{Main Results}
\label{sec:exp:main}\label{sec:exp:signals}\label{sec:exp:analysis}

\providecommand{\sepm}[1]{{\scriptsize\textcolor{black!55}{\textpm#1}}}
\begin{table}[t]
\vspace{10pt}
\setlength{\abovecaptionskip}{0pt}\setlength{\belowcaptionskip}{3pt}
\begin{minipage}[t]{0.51\textwidth}
\centering
\caption{\textbf{ANTE outperforms both baselines at every scale on SWE-bench Verified} (pass@1, \%). 
$^{-}$: without protocol annotated (Appendix~\ref{app:prompts});
$\uparrow$: gain over the variant's SFT checkpoint, and for ANTE-TTO over ANTE; $\pm$: s.e.\ over 3 evaluation runs of one checkpoint, for the rows run three times.}
\label{tab:main}
\footnotesize
\setlength{\tabcolsep}{2.5pt}
\renewcommand{\arraystretch}{1.15}
\begin{tabular}{@{}llccc@{}}
\toprule
 & Method & 7B & 14B & 32B \\
\midrule
\multirow{4}{*}{\rotatebox[origin=c]{90}{\scriptsize SFT}}
 & R2E-Gym Agent & 19.0 & 26.8 & 25.0 \\
 & Reactive      & 23.4 & 40.4 & 45.8 \\
 & Retrospective & 23.0$^{-}$\,/\,22.4 & 41.0$^{-}$\,/\,40.2 & 47.2$^{-}$\,/\,48.2 \\
 & ANTE          & 22.2$^{-}$\,/\,24.8 & 40.2$^{-}$\,/\,41.4 & 48.8$^{-}$\,/\,46.4 \\
\midrule
\multirow{4}{*}{\rotatebox[origin=c]{90}{\scriptsize RL}}
 & Reactive-RL          & 24.4$^{\uparrow 1.0}$\sepm{0.1} & 41.8$^{\uparrow 1.4}$\sepm{0.4} & 51.0$^{\uparrow 5.2}$\sepm{0.6} \\
 & Retrospective-RL     & 25.2$^{\uparrow 2.8}$\sepm{0.3} & 43.4$^{\uparrow 3.2}$\sepm{0.3} & 52.4$^{\uparrow 4.2}$\sepm{0.4} \\
 & ANTE (w/o BAPO)      & \underline{25.4}$^{\uparrow 0.6}$ & \underline{43.6}$^{\uparrow 2.2}$ & \underline{54.2}$^{\uparrow 7.8}$ \\
\rowcolor{beliefbg}\cellcolor{white}
 & ANTE                 & \textbf{27.2}$^{\uparrow 2.4}$\sepm{1.7} & \textbf{46.2}$^{\uparrow 4.8}$\sepm{0.9} & \textbf{57.2}$^{\uparrow 10.8}$\sepm{1.1} \\
\midrule
 & $\Delta$ Reactive baseline & $+2.8$ & $+4.4$ & $+6.2$ \\
\midrule
 & $+$ ANTE-TTO & 28.6$^{\uparrow 1.4}$ & 48.6$^{\uparrow 2.4}$ & 57.8$^{\uparrow 0.6}$ \\
\bottomrule
\multicolumn{5}{@{}p{226pt}@{}}{\scriptsize R2E-Gym Agent: at 7B and 14B, the agents released by R2E-Gym (fine-tuned Qwen2.5-Coder); at 32B, the Qwen3-32B backbone on the R2E-Gym scaffold.}
\end{tabular}
\end{minipage}\hfill
\begin{minipage}[t]{0.47\textwidth}
\centering
\caption{\textbf{ANTE outperforms both baselines on every terminal benchmark} (resolved, \%). All variants build on Qwen3.5-9B and run with the Tmax agent \citep{ivison2026tmax} on Harbor \citep{harbor2025}, with the matched setup of Table~\ref{tab:main} (protocol in Appendix~\ref{app:term}). The terminal SFT checkpoints, which install the belief protocol before RL, are not evaluated separately, so $\uparrow$ is the gain over the base model rather than over each variant's SFT checkpoint; $\pm$: s.e.\ over 3 evaluation runs of one checkpoint.}
\label{tab:term-results}
\footnotesize
\setlength{\tabcolsep}{1.8pt}
\renewcommand{\arraystretch}{1.15}
\begin{tabularx}{\linewidth}{@{}l*{3}{>{\centering\arraybackslash}X}@{}}
\toprule
  & TB-Dev & TB-Lite & TB-2.0 \\
\midrule
Qwen3.5-9B (base)             & 32.9\sepm{0.8} & 32\sepm{2.0} & 16.9\sepm{0.7} \\
\midrule
Reactive-RL                   & 35.6$^{\uparrow 2.7}$\sepm{1.6} & 35$^{\uparrow 3}$\sepm{0.9} & 18.0$^{\uparrow 1.1}$\sepm{2.6} \\
Retrospective-RL              & 37.0$^{\uparrow 4.1}$\sepm{0.9} & \underline{43}$^{\uparrow 11}$\sepm{2.1} & 20.2$^{\uparrow 3.3}$\sepm{1.1} \\
ANTE (w/o BAPO)               & \underline{41.1}$^{\uparrow 8.2}$\sepm{1.2} & 39$^{\uparrow 7}$\sepm{2.2} & \underline{22.5}$^{\uparrow 5.6}$\sepm{0.4} \\
\rowcolor{beliefbg}
ANTE                          & \textbf{42.5}$^{\uparrow 9.6}$\sepm{3.3} & \textbf{45}$^{\uparrow 13}$\sepm{1.3} & \textbf{24.7}$^{\uparrow 7.8}$\sepm{1.3} \\
\midrule
$\Delta$ Reactive baseline   & $+6.9$ & $+10.0$ & $+6.7$ \\
\bottomrule
\end{tabularx}

\par\vspace{5pt}
{\scriptsize\raggedright
\par}
\end{minipage}

\end{table}

Our main results are shown in Tables~\ref{tab:main}
and~\ref{tab:term-results}. Adding the belief state and training with BAPO improves over both baselines at every scale and on every benchmark. Concretely, on SWE-bench Verified, ANTE leads both baselines
at every scale, by 2.8, 4.4, and 6.2 points over Reactive and by 2.0,
2.8, and 4.8 over Retrospective at 7B, 14B, and 32B, and reaches
57.2\% pass@1 at 32B (Table~\ref{tab:main}). On the terminal tasks, it leads both
baselines on all three benchmarks, by 6.7 to 10.0 points over
Reactive and by 2.0 to 5.5 over Retrospective, and resolves 24.7\% of
Terminal-Bench 2.0 (Table~\ref{tab:term-results}). 

Besides, we find that without BAPO, the belief state alone already improves over Reactive, by 1.0 to 3.2 points on SWE-bench Verified and 4.0 to 5.5 on the terminal tasks, and matches or exceeds
Retrospective except on TB-Lite. Furthermore, training with BAPO adds 1.8, 2.6, and 3.0 points on top of the belief state at 7B, 14B, and 32B, about half or more of ANTE’s lead over Reactive, and 1.4 to 6.0 points on the terminal tasks,  so a large part of the gain comes from BAPO's step-level credit (\S\ref{sec:exp:why}).
Furthermore, our experiments corroborate the theoretical analysis: declining SFT loss supports belief-augmented initialization (Proposition~\ref{prop:imitation}; Figure~\ref{fig:sft9}), while improving forecast accuracy and consistent ANTE-TTO gains show that learned beliefs are both predictive and actionable (Proposition~\ref{prop:selection}; Figure~\ref{fig:hit}; Table~\ref{tab:main}).

We also compare ANTE with published open-weight agents. This comparison asks
whether the resulting agents are competitive with existing
systems in the literature at comparable model scales. On
SWE-bench Verified, ANTE leads the task-trained agents at every
scale; above 30B it is ahead of SWE-World, FrogBoss \citep{sonwane2025bugpilot}, and CWM, and of
every 72B method (Table~\ref{tab:sota-full},
Appendix~\ref{app:concurrent}). On Terminal-Bench 2.0, ANTE-9B leads
every terminal agent below 20B that we compare with, including ECHO,
which predicts observations but never commits to a scored forecast
(Table~\ref{tab:term-market}, Appendix~\ref{app:term}).

\subsection{Out-of-Domain Generalization}
\label{sec:exp:gen}\label{sec:exp:ood}

\providecommand{\oodup}[1]{\textcolor{paloalto}{$\uparrow$#1}}
\providecolor{lavbg}{HTML}{EFEDF8}
\providecommand{\oodchip}[4]{\tikz[baseline=(c.base)]\node[rounded corners=2.5pt, fill=#1, draw=#2, line width=0.4pt, inner xsep=3.5pt, inner ysep=1.4pt] (c) {\textcolor{#2}{#3}\hspace{0.35em}{\itshape #4}};}
\providecommand{\ooddom}[4]{\multicolumn{4}{@{}l@{}}{\cellcolor{#1}\oodchip{#1}{#2}{#3}{#4}}\\[-1pt]}
Tables~\ref{tab:main} and~\ref{tab:term-results} measure ANTE inside
the harnesses it was trained on; we next ask whether the foresight it
learns there transfers beyond agentic coding. Table~\ref{tab:ood}
shows ANTE-32B and Reactive-RL on eleven benchmarks
spanning code execution prediction \citep{gu2024cruxeval,jain2024livecodebench}, state tracking (BIG-Bench Hard, BBH) \citep{suzgun2022bbh}, mathematics \citep{hendrycks2021math,lightman2023verify,mathai2025aime25,mathai2025amc23,lewkowycz2022minerva}, and science \citep{wang2024mmlupro,wang2023scibench}, answered with a plain chain-of-thought prompt \citep{wei2022chain}. The two methods share the backbone, trajectories, and training
procedure and differ only in the belief and its signals, and therefore we can study what the belief changes in general reasoning. ANTE leads on all datasets by 6.2 points
on average and most on AIME'25 and MMLU-Pro, and the traces in Appendix~\ref{app:case-ood} show the foresight also works outside the harness.

\begin{table}[!htb]
\setlength{\abovecaptionskip}{0pt}\setlength{\belowcaptionskip}{3pt}
\begin{minipage}[t]{0.47\textwidth}\vspace{0pt}
\centering
\caption{Out-of-distribution generalization of ANTE-32B.}
\label{tab:ood}
\small
\setlength{\tabcolsep}{5pt}
\begin{tabular}{@{}lr>{\columncolor{beliefbg}}rr@{}}
\toprule
\textbf{Benchmark} & Reactive & \multicolumn{1}{c}{ANTE} & \multicolumn{1}{c}{$\Delta$} \\
\midrule
\ooddom{cardgray}{icebd}{\faCode}{Execution Prediction}
CruxEval-O & 83.6 & \textbf{85.4} & \oodup{1.8} \\
CruxEval-I & 72.2 & \textbf{78.8} & \oodup{6.6} \\
LCB-exec & 91.4 & \textbf{94.2} & \oodup{2.8} \\
\ooddom{cream}{butterbd}{\faProjectDiagram}{State Tracking}
BBH-hard & 87.2 & \textbf{90.9} & \oodup{3.7} \\
BBH-state & 88.6 & \textbf{91.7} & \oodup{3.1} \\
\ooddom{peachbg}{peachbd}{\faCalculator}{Math Reasoning}
MATH500 & 94.8 & \textbf{97.2} & \oodup{2.4} \\
AIME'25 & 63.3 & \textbf{80.0} & \oodup{16.7} \\
AMC23 & 90.0 & \textbf{97.5} & \oodup{7.5} \\
Minerva & 38.6 & \textbf{46.0} & \oodup{7.4} \\
\ooddom{lavbg}{lavbd}{\faFlask}{Scientific Reasoning}
MMLU-Pro & 72.8 & \textbf{83.2} & \oodup{10.4} \\
SciBench & 66.0 & \textbf{71.4} & \oodup{5.4} \\
\midrule
\textit{Mean of 11} & 77.1 & \textbf{83.3} & \oodup{6.2} \\
\bottomrule
\end{tabular}
\end{minipage}\hfill
\begin{minipage}[t]{0.49\textwidth}\vspace{0pt}
\centering
\includegraphics[width=0.96\linewidth]{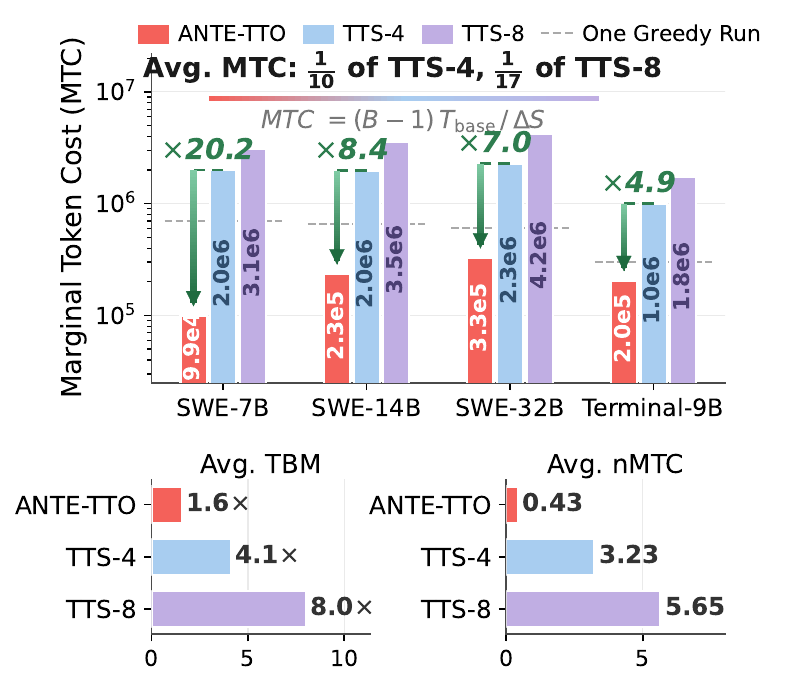}
\captionof{figure}{\textbf{ANTE-TTO cost analysis.} $B_{\mathrm{tok}}$ (TBM in the panels): inference tokens per task relative to one greedy attempt; $\mathrm{nMTC} = (B_{\mathrm{tok}}{-}1)/\Delta S$: greedy attempts spent per point of score gain; axes show inference tokens as $10^{x}$.}
\label{fig:mtc}
\end{minipage}
\vspace{2pt}
\end{table}

\subsection{ANTE-TTO: Gains and Overhead of Test-Time Foresight}
\label{sec:exp:tto}
We evaluate ANTE-TTO as a test-time scaling strategy, measuring its gains in task success alongside the additional inference cost.
ANTE-TTO improves task success at every scale (Table~\ref{tab:main}) while resampling only at trigger steps, scoring the candidates by the beliefs the policy already emits, and executing exactly one (\S\ref{sec:method:validator}). Best-of-$n$ sampling (TTS-$n$), execution-based reranking \citep{chen2022codet}, and process-reward reranking \citep{lightman2023verify} instead multiply full rollouts, execute every candidate, or query an external scorer. Measured from call counts, ANTE-TTO runs at $B_{\mathrm{tok}}
= 1.31\times$ at 7B against $4.42\times$ for TTS-4, and at
$1.6\times$ against $4.1\times$ averaged over its four settings
(Figure~\ref{fig:mtc}); Figure~\ref{fig:case} shows ANTE-TTO case studies on different harnesses end to end.

\subsection{Why ANTE Works}
\label{sec:exp:why}\label{sec:exp:decision}

\begin{figure}[t]
\setlength{\abovecaptionskip}{2pt}
\centering
\includegraphics[width=0.75\textwidth]{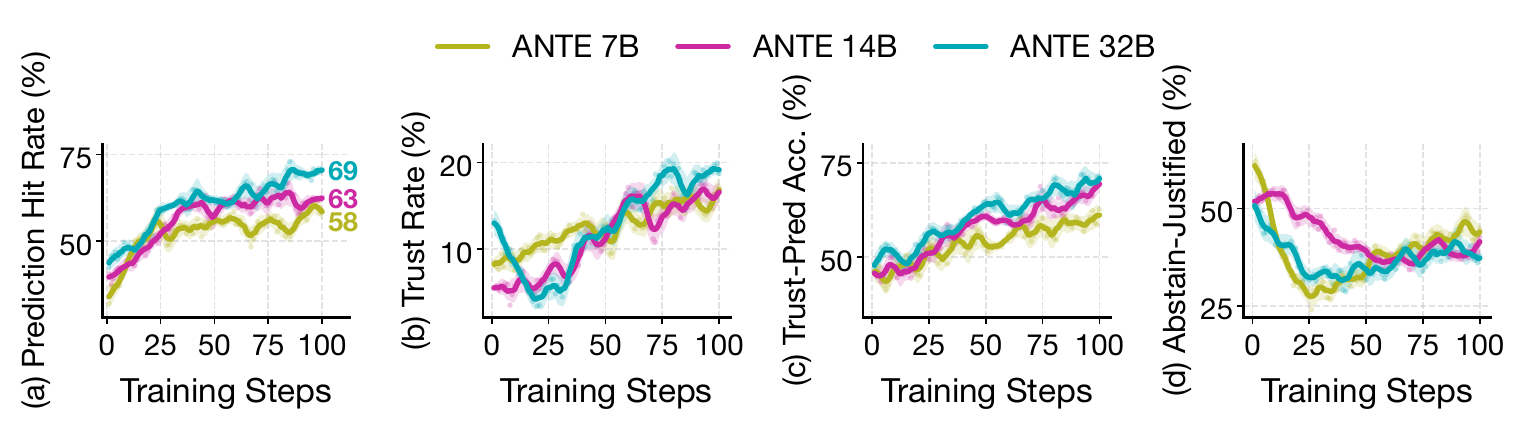}
\caption{\textbf{The world model sharpens and the trust decision is
learned.} (a) The prediction hit rate of ANTE's world model on
conclusive steps rises throughout RL at every scale. (b)--(d) BAPO
analysis: (b) the trust rate doubles, (c) the accuracy of trusted predictions rises, and (d)
abstention concentrates on the steps that remain uncertain.}
\label{fig:hit}\label{fig:trust}
\vspace{2pt}
\end{figure}

\begin{figure}[t]
\setlength{\abovecaptionskip}{3pt}
\centering
\includegraphics[width=0.92\textwidth]{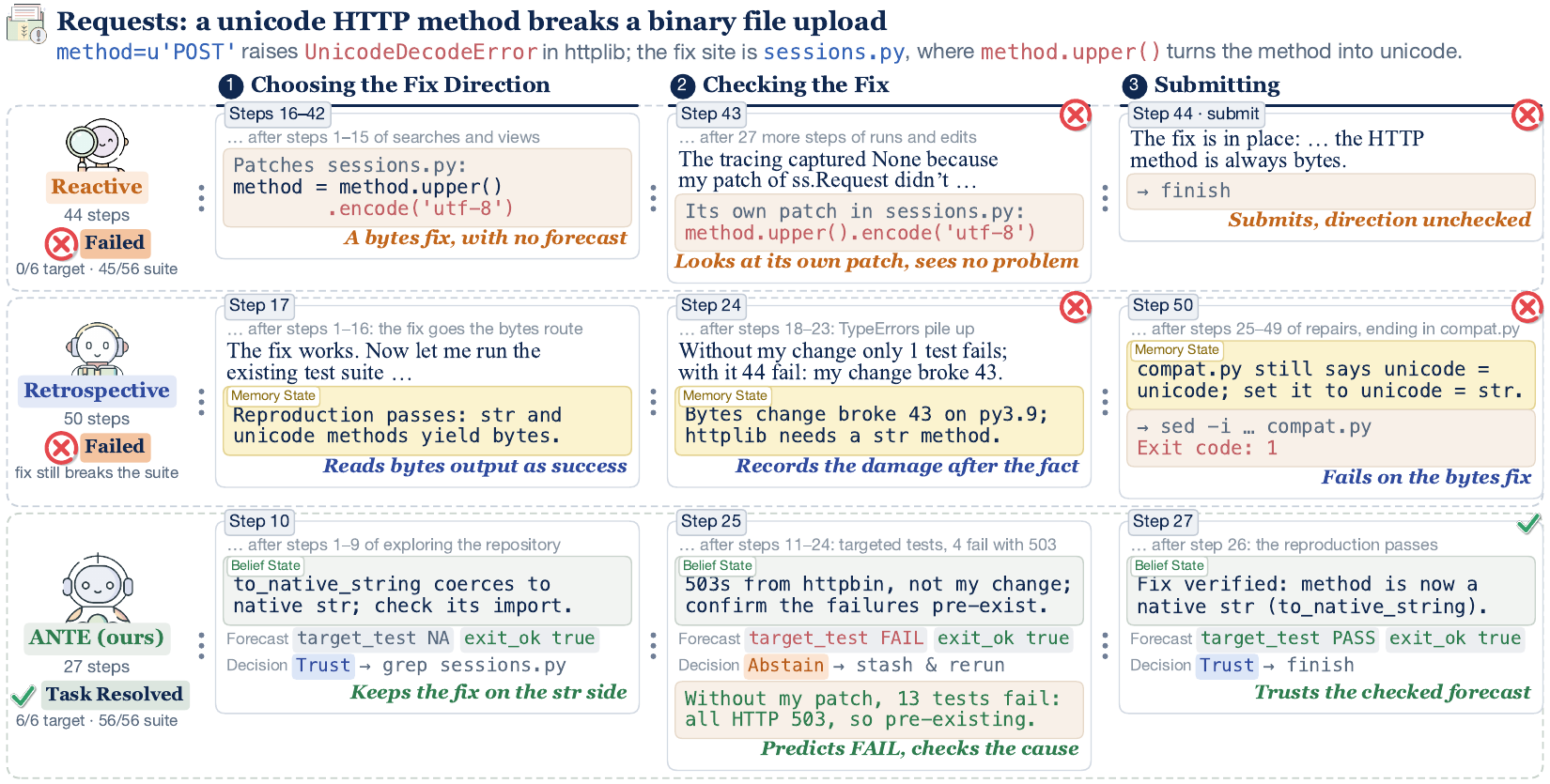}
\caption{\textbf{Case study.} The internal belief helps reinforce ANTE's world model reasoning for coding agents (task: instance \texttt{psf\_\_requests-1724} of SWE-bench Verified).}
\label{fig:case2b} 
\end{figure}
\paragraph{The world model sharpens and the trust decision is learned.}
To understand the behavior that yields these results, we analyze
ANTE's trust decisions, the relationship between its beliefs and
task outcomes, and representative trajectories.
Figure~\ref{fig:trust} traces the belief through RL. The hit rate of
conclusive predictions rises at every scale (a), from 44\% to 69\% at
32B, so the forecasts the policy acts on become more accurate as
training proceeds. The trust rate doubles (b), the accuracy of trusted predictions rises past $q^{\ast}$, the level above which trusting is rewarded (c), and abstention concentrates on the steps that remain uncertain (d). The belief also outperforms SWE-World's dedicated 32B reward model \citep{sun2026sweworld} as a judge of its trajectories (Table~\ref{tab:swr}). Together with the ANTE-TTO gains of Table~\ref{tab:main}, this is the empirical side of Proposition~\ref{prop:selection}. RL lowers the forecast error that bounds the selection regret, the forecasts carry outcome signal, and selecting by them raises success. The foresight is therefore not only predictive but also actionable.

\paragraph{Case study.} Figure~\ref{fig:case2b} shows how the three models handle a repository issue. In this case, uploading a file with the method \texttt{u'POST'}, a Unicode string, raises a \texttt{UnicodeDecodeError}. The fix is to convert the method to the native string type, which is bytes on Python 2 and str on Python 3. Reactive reasons only about the immediate error and encodes the method to bytes. It is not aware of the environment the fix must run in, so when Python 3 rejects the bytes method, it dismisses this as outside the issue and submits the fix. For Retrospective, the memory records only past steps and indicates nothing about what its bytes edit will do. ANTE's belief predicts what the next edit will change. It scopes the bug to Python 2 (``force bytes in py2''), so it uses the existing native-string helper instead of forcing bytes everywhere. This foresight ability makes ANTE resolve tasks more robustly (full trace in Appendix~\ref{app:case-studies}).

\section{Conclusion}
\label{sec:conclusion}

With ANTE, we want coding agents to foresee the consequences of their actions and to let real execution correct their beliefs. During training, ANTE's world model forms an internal belief within the policy: what the agent has established, what the code does now, and what the next command will change. The foresight ability lets ANTE improve across harnesses and model scales, and generalize to tasks in other domains. In summary, we envision agents whose beliefs span their environment's past, present, and future, acting on what they anticipate and learning from what execution reveals.

\bibliographystyle{assets/plainnat}
\bibliography{references}

\clearpage
\beginappendix
\startcontents[appendix]
\section*{Appendix Contents}
{\footnotesize\setlength{\parskip}{0pt}\printcontents[appendix]{}{1}{\setcounter{tocdepth}{2}}}
\clearpage
\section{Preliminaries}
\label{sec:prelim}

\paragraph{Coding agents as POMDPs.}
\begingroup\sloppy
We formalize a coding agent working on a task inside its container as a partially-observable Markov decision process (POMDP) $(\mathcal{S}, \mathcal{A}, \mathcal{O}, \mathcal{T}, \Omega, \mathcal{R})$ \citep{astrom1965optimal,kaelbling1998planning}. The latent state $s_t$ is the full repository and container, an action $a_t$ is one tool call, $\mathcal{T}(s_{t+1} \mid s_t, a_t)$ executes it, the observation $o_{t+1} \sim \Omega(\cdot \mid s_{t+1}, a_t)$ is its textual output, and the reward $\mathcal{R}$ is a terminal outcome in $\{0, 1\}$. The agent therefore acts on the history $h_t$ of everything it has written and observed. Over histories the process is Markov with kernel $P_{\mathrm{env}}(o_{t+1} \mid h_t, a_t)$, so policy gradients apply unchanged. We use the corrected group-relative policy gradient of Dr.GRPO and DAPO \citep{shao2024deepseekmath,liu2025drgrpo,yu2025dapo} for the policy term of Eq.~\ref{eq:objective} (Appendix~\ref{app:hparams}). The belief $b_t = P(s_t \mid h_t)$ suffices for optimal control but cannot be computed or checked for a repository, so learning agents keep a belief representation instead, a function of the history trained to predict what the latent state determines \citep{gangwani2020bmil}.\par
\endgroup

\paragraph{Coding tasks and execution-based verification.}
A repository issue \citep{jimenez2024swebench,jain2025r2egym} consists of a containerized repository at a fixed commit, an issue description, and a hidden test suite $\mathcal{U}$. It is resolved when the submitted patch passes every fail-to-pass and pass-to-pass test in $\mathcal{U}$. A terminal task states a goal in natural language inside an isolated container, and a hidden checker grades the final state. In both harnesses, each step is one assistant message that reasons and then makes exactly one tool call, and the environment replies only in text. Execution can check one part of the belief exactly: the deterministic verifier $\psi$ of \S\ref{def:chi-psi} \citep{jain2025r2egym} maps each response to an outcome in $\mathcal{Y}$, or to $\bot$ when it determines none. The realized outcome $y_t := \psi(o_{t+1}, \chi(a_t))$ therefore needs no judge, learned reward model, or simulator, and its distribution given $(h_t, a_t)$ is fixed by $P_{\mathrm{env}}$. ANTE trains the policy to predict $y_t$ before $a_t$ runs and scores the forecast against the realized outcome (\S\ref{sec:method}). The belief state of \S\ref{sec:method:belief} is the written form of this belief representation, and its forecast is the part that execution checks.

\definecolor{obsbg}{HTML}{F4F5F4}
\definecolor{trajred}{HTML}{C25450}
\definecolor{trajamber}{HTML}{B08430}
\newtcolorbox{trajstep}[2]{colback=white, colframe=#1, arc=1.5pt,
boxrule=0.7pt, left=5pt, right=5pt, top=3pt, bottom=3pt,
fonttitle=\bfseries\footnotesize, coltitle=white, title={#2}, breakable,
before skip=6pt, after skip=2pt}
\newtcolorbox{trajbelief}{colback=beliefbg, colframe=beliefbg, arc=1pt,
boxrule=0pt, left=4pt, right=4pt, top=2pt, bottom=2pt,
fontupper=\ttfamily\fontsize{7.3}{9.2}\selectfont,
before skip=3pt, after skip=3pt}
\newtcolorbox{trajobs}{colback=obsbg, colframe=obsbg, arc=1pt,
boxrule=0pt, left=4pt, right=4pt, top=2pt, bottom=2pt,
fontupper=\ttfamily\fontsize{7.3}{9.2}\selectfont, breakable,
before skip=2pt, after skip=6pt}
\newtcolorbox{trajmem}{colback=cream, colframe=cream, arc=1pt,
boxrule=0pt, left=4pt, right=4pt, top=2pt, bottom=2pt,
fontupper=\ttfamily\fontsize{7.3}{9.2}\selectfont,
before skip=3pt, after skip=3pt}
\newtcolorbox{trajnone}{colback=cardgray, colframe=cardgray, arc=1pt,
boxrule=0pt, left=4pt, right=4pt, top=2pt, bottom=2pt,
fontupper=\itshape\fontsize{7.3}{9.2}\selectfont,
before skip=3pt, after skip=3pt}

\FloatBarrier
\section{Notation}
\label{app:notation}

\begin{table}[H]
\centering
\scriptsize
\caption{\textbf{Notation.} Hats mark estimates; $\tilde y$ denotes a value of the predicted key field (a group key) and $\tilde A$ a standardized advantage.}
\label{tab:notation}
\setlength{\tabcolsep}{3pt}
\renewcommand{\arraystretch}{1.0}
\begin{tabular}{@{}>{\raggedright\arraybackslash}p{3.4cm} >{\raggedright\arraybackslash}p{4.9cm} >{\raggedright\arraybackslash}p{5.2cm}@{}}
\toprule
\textbf{Symbol} & \textbf{Meaning} & \textbf{Definition or reference} \\
\midrule
\multicolumn{3}{@{}l}{\textit{Environment and rollouts}} \\
$(\mathcal{S}, \mathcal{A}, \mathcal{O}, \mathcal{T}, \Omega, \mathcal{R})$ & POMDP of a harness & Appendix~\ref{sec:prelim} \\
$P_{\mathrm{env}}(o_{t+1} \mid h_t, a_t)$ & Kernel of the history process & Appendix~\ref{sec:prelim} \\
$b_t$ & Posterior belief & $P(s_t \mid h_t)$ \\
$\mathcal{Y} = \mathcal{Y}^{\mathrm{tst}} \times \{0,1\}$ & Outcome space & $\mathcal{Y}^{\mathrm{tst}} = \{\textsc{pass}, \textsc{fail}, \textsc{na}\}$; \S\ref{def:outcome} \\
$x \sim \mathcal{D}$; $\mathcal{D}_{\mathrm{BT}}$ & Task; BeliefTraj corpus & \S\ref{def:rollout}; \S\ref{sec:method:sft} \\
$\pi_\theta$; $\pi_{\mathrm{old}}$; $\pi_{\mathrm{ref}}$ & Policy; policy that generated the batch; reference policy & Eq.~\ref{eq:objective} \\
$N$; $\tau_i$; $T_i$; $h_{i,t}$ & Rollouts per task; rollout $i$; its length; history before step $t$ & \S\ref{def:rollout} \\
$u_{i,t} = \langle v_{i,t}, \beta_{i,t}, a_{i,t} \rangle$ & Step message & Reasoning, belief state, action; \S\ref{def:rollout} \\
$c_{i,t} = (h_{i,t}, v_{i,t}, z_{i,t})$ & Writing context of the forecast & \S\ref{def:belief} \\
$\mathbb{E}_F[f]$ & Mean of $f$ over a step set $F$ & \S\ref{sec:method:bapo} \\
\multicolumn{3}{@{}l}{\textit{Outcomes and belief state}} \\
$\chi$; $\psi$ & Action type; execution-based verifier & $\mathcal{A} \to \{\textsc{epi}, \textsc{prag}\}$; \S\ref{def:chi-psi} \\
$y_{i,t} = (y^{\mathrm{tst}}_{i,t}, y^{\mathrm{exe}}_{i,t})$ & Realized outcome (check field, exit field) & $\psi(o_{i,t+1}, \chi(a_{i,t})) \in \mathcal{Y} \cup \{\bot\}$ \\
$m_{i,t}$ & Verifiable-step indicator & $1$ iff the block parses and $\psi \neq \bot$ \\
$\mathcal{C}_i$; $\mathcal{C}$ & Conclusive steps (rollout; task) & \S\ref{def:chi-psi} \\
$R_i$; $r^{\mathrm{fmt}}_i$ & Task reward; format term & $\mathbf{1}[\text{resolved}] + r^{\mathrm{fmt}}_i$; \S\ref{sec:method:objective} \\
$A_i$ & Episode advantage & $R_i - \tfrac{1}{N}\sum_j R_j$ \\
$\beta_{i,t} = (z_{i,t}, \hat y_{i,t}, d_{i,t})$ & Belief state & Summary, forecast, decision; \S\ref{def:belief} \\
$\hat y_{i,t} = (\ytst_{i,t}, \yexe_{i,t})$ & Forecast (check field, exit field) & $\in \mathcal{Y}$ \\
$d_{i,t}$ & Decision & $\{\textsc{trust}, \textsc{abstain}\}$; corpus labels in Appendix~\ref{app:corpus} \\
\multicolumn{3}{@{}l}{\textit{Decision reward and credit}} \\
$\sigma$; $r_{i,t}$ & Scoring rule; decision reward & $r_{i,t} = \sigma(d_{i,t}, \hat y_{i,t}, y_{i,t})$ on $\mathcal{C}_i$; Eq.~\ref{eq:settle} \\
$\alpha$, $\gamma$, $\kappa$ & Bonus, penalty, abstention cost & Eq.~\ref{eq:settle} \\
$q^{\ast}$; $q^{\circ}$ & Target accuracy; indifference point & $\gamma/(\alpha+\gamma)$; $q^{\ast} - \kappa/(\alpha+\gamma)$; clean-exit forecasts \\
$W_{i,t}$ & Decision return & $\sum_{s \ge t} r_{i,s}$ \\
$G(\tilde y)$; $G(\tilde y, d)$ & Belief group; belief-decision group & \S\ref{sec:method:bapo} \\
$\mathcal{G}$ & Anchored steps & Union of the belief groups (key: predicted check field; exit field on terminal tasks) \\
$\Vhat(\tilde y)$; $\Qhat(\tilde y, d)$ & Belief value; belief-decision value & Means of $W$ over $G(\tilde y)$, $G(\tilde y, d)$; Eq.~\ref{eq:values} \\
$A^{\tilde y}_{i,t}$; $A^{\tilde y, d}_{i,t}$ & Belief-relative; decision-relative advantage & $W_{i,t} - \Vhat(\tilde y)$; $W_{i,t} - \Qhat(\tilde y, d)$ \\
$\Ahat(\tilde y, d)$ & Decision advantage & $\Qhat(\tilde y, d) - \Vhat(\tilde y)$ \\
$\tilde A^{\tilde y}_{i,t}$; $\tilde A^{\tilde y, d}_{i,t}$ & Standardized streams & Per-task divisor $\max(\mathrm{std}, \alpha + \gamma)$; Appendix~\ref{app:hparams} \\
$\Delta_{i,t}$; $\omega$ & Step credit; its weight & $\omega(\tilde A^{\tilde y}_{i,t} + \tilde A^{\tilde y, d}_{i,t})$ on $\mathcal{G}$, else $0$; Eq.~\ref{eq:bapo-adv} \\
$A_{i,t}$ & Per-step advantage & $A_i + \Delta_{i,t}$ \\
\multicolumn{3}{@{}l}{\textit{Objective, corpus, and selection}} \\
$\lpred$ & Prediction loss & Eq.~\ref{eq:lpred} \\
$\mathcal{K}_{i,t}$; $\mathcal{V}_{i,t}$; $Z$ & Policy tokens; value tokens; token-mean denominator & Eqs.~\ref{eq:lpred}, \ref{eq:objective} \\
$\varrho_{i,t,k}$ & Importance ratio & $\pi_\theta / \pi_{\mathrm{old}}$ per policy token \\
$\epsilon_{\mathrm{lo}}, \epsilon_{\mathrm{hi}}$; $\eta$; $\lambda$; $\eta_{\mathrm{lr}}$ & Clip range; KL coefficient; prediction weight; learning rate & Eq.~\ref{eq:objective}; Alg.~\ref{alg:bapo-compact} \\
$s(\beta)$; $\phi(\cdot)$ & Candidate score; check-field rank & Eq.~\ref{eq:tto-score} \\
$K$ & Alternatives per trigger & $K + 1$ candidates with the draft; Alg.~\ref{alg:tto} \\
\multicolumn{3}{@{}l}{\textit{Analysis (Appendix~\ref{app:proofs})}} \\
$P_E$, $J$, $T$, $H_E$, $\varepsilon_{\mathrm{SFT}}$, $\Lambda$; $p_k$, $q_k$, $\delta_k$, $u$ & Symbols of Propositions~\ref{prop:imitation} and~\ref{prop:selection} & Defined in their statements \\
$p(\cdot \mid c_t)$; $q_\theta(\cdot \mid c_t)$ & Outcome law; policy predictive over $\mathcal{Y}$ & Appendix~\ref{app:proofs:setting}, under $m_t = 1$ \\
$\delta(c_t)$; $\bar{\mathcal{L}}_t$ & Forecast divergence; expected prediction loss & $\mathrm{KL}(p \,\|\, q_\theta)$; $\mathbb{E}[-\log q_\theta(y_t \mid c_t)]$; Eq.~\ref{eq:excess-risk} \\
\bottomrule\end{tabular}
\end{table}

\FloatBarrier
\section{Theoretical analysis of the ANTE objective}
\label{app:proofs}

(i) The belief-anchored advantage splits credit for a decision from
credit for a step (Proposition~\ref{prop:decomp},
Appendix~\ref{app:proofs:decomp}). (ii) Read as imitation, Stage~I lowers
an upper bound on the gap to the demonstration process
(Proposition~\ref{prop:imitation}). (iii) An idealized form of the
policy term of Eq.~\ref{eq:objective} lower-bounds task
performance, and the prediction weight $\lambda$ indexes a monotone
family of accuracy budgets (Lemma~\ref{lem:surrogate},
Proposition~\ref{prop:lagrangian}). (iv) The expected prediction loss upper-bounds the uncertainty left about the checkable outcome, and $H(y_t \mid a_t) - \bar{\mathcal L}_t$ lower-bounds the information the writing context carries about the next latent state. The forecast divergence upper-bounds the cost of acting on a forecast, in training and at test time (Proposition~\ref{prop:variational},
Corollaries~\ref{cor:pricing}--\ref{cor:pointsel},
Proposition~\ref{prop:selection}, Appendix~\ref{app:proofs:information}).

\subsection{Setting and the quantities the analysis controls}
\label{app:proofs:setting}

\begin{figure}[!t]
\centering
\includegraphics[width=0.9\textwidth]{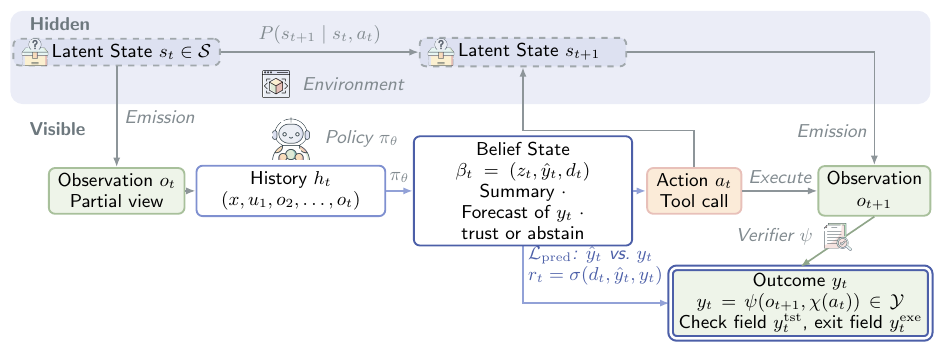}
\caption{\textbf{Belief as a point summary of the history.} The forecast $\hat y_t$ predicts the one coordinate of the next observation that the verifier $\psi$ reads; the prediction loss checks it against $y_t$, and the decision reward scores $d_t$ against the same outcome.}
\label{fig:app-pomdp}
\end{figure}

We work in the partially observed process of Appendix~\ref{sec:prelim} and
\S\ref{def:rollout}. A step policy $\pi(u \mid h)$ emits the message $u = (v, \beta, a)$ (Figure~\ref{fig:app-pomdp}) given the history $h$, which holds the earlier messages and observations but not the current reasoning; where only the belief and the action matter we abbreviate the emission as $(\beta, a)$, the reasoning being carried inside $u$; the environment kernel $P_{\mathrm{env}}(o_{t+1} \mid h_t, a_t) = \sum_{s, s'} P(s_t = s
\mid h_t)\, \mathcal{T}(s' \mid s, a_t)\, \Omega(o_{t+1} \mid s', a_t)$ depends on the step message only through the action. Every rollout ends by a horizon $T$
(a terminated rollout emits null steps up to $T$ with probability one
under every policy considered), the terminal outcome is recorded as part
of the last observation, so that $R \in [0, 1]$ is a function of the
trajectory (binary in both harnesses; only boundedness is used), and the analysis uses the task outcome alone, without the
bounded format term $r^{\mathrm{fmt}}_i$ of \S\ref{sec:method:objective}. $P_\pi$
denotes the law of trajectories under $\pi$ and the real environment,
and the performance measure is $J(\pi) := \mathbb{E}_{P_\pi}[R]$. For a
verifiable step ($m_t = 1$, \S\ref{def:chi-psi}), called a
supervised step below, $p(\cdot \mid c_t)$ is
the true conditional law of the realized outcome $y_t$ given the writing
context $c_t = (h_t, v_t, z_t)$ of \S\ref{def:belief}, $q_\theta(\cdot
\mid c_t)$ the policy's predictive at the value positions renormalized
over $\mathcal{Y}$, and
\begin{equation}
\delta(c_t) := \mathrm{KL}\big(p(\cdot \mid c_t)\,\big\|\,q_\theta(\cdot \mid c_t)\big),
\qquad
\bar{\mathcal L}_t(\theta) := \mathbb{E}\big[-\log q_\theta(y_t \mid c_t)\big]
= H(y_t \mid c_t) + \mathbb{E}[\delta(c_t)],
\label{eq:excess-risk}
\end{equation}
the pointwise forecast divergence and the step-$t$ expected prediction loss; the identity
is the decomposition of cross-entropy into entropy plus divergence. The
training loss $\lpred$ of Eq.~\ref{eq:lpred} is a token-normalized
empirical surrogate for both. It scores the unnormalized value-token likelihood, which is at most the renormalized one, and weights supervised steps by $1/Z$. The statements below control the per-step expected prediction loss $\bar{\mathcal L}_t$. Quantities attached to supervised steps are taken under the rollout law conditioned on $m_t = 1$, written $\mathcal{M}$ in the prediction and information statements of Appendix~\ref{app:proofs:information}. The trajectory-level statements of Appendices~\ref{app:proofs:imitation} and~\ref{app:proofs:objective} use the unconditioned trajectory laws. Candidates whose outcome is $\bot$ are excluded from the selection statements, and logarithms are natural. Total variation is $\mathrm{TV}(P, Q) := \sup_E |P(E) -
Q(E)|$; hence $|\mathbb{E}_P[f] - \mathbb{E}_Q[f]| \le \mathrm{TV}(P, Q)$ for $f \in [0, 1]$, and $\le 2\|f\|_\infty \mathrm{TV}(P,Q)$ in general. Pinsker's inequality reads $\mathrm{TV}(P, Q) \le
\sqrt{\mathrm{KL}(P \| Q)/2}$.

\subsection{Decomposition of the anchored advantage}
\label{app:proofs:decomp}

BAPO subtracts two baselines from one decision return. The statement
below shows that they are not redundant: together they split the return into the credit for choosing a decision and the credit for the step inside that decision.

\begin{proposition}[Anchored-advantage decomposition]
\label{prop:decomp}
Fix a task with rollouts $\tau_1, \dots, \tau_N$ and an anchor value
$\tilde y \in \{\textsc{pass}, \textsc{fail}\}$ whose belief group
$G(\tilde y)$ is non-empty, and let the belief-decision groups $G(\tilde y, d)$,
$d \in \{\textsc{trust}, \textsc{abstain}\}$, partition it, both
non-empty for the final claim. Write $\mu_d := |G(\tilde y, d)| / |G(\tilde y)|$ for the decision weights, and let
$\Ahat(\tilde y, d) := \Qhat(\tilde y, d) - \Vhat(\tilde y)$ denote the
decision advantage of $d$, a quantity shared by every member of
$G(\tilde y, d)$. Then, identically on $G(\tilde y)$,
\begin{equation}
A^{\tilde y}_{i,t} \;=\; \Ahat(\tilde y, d_{i,t}) + A^{\tilde y, d}_{i,t},
\label{eq:decomp}
\end{equation}
and under the empirical measure on $G(\tilde y)$ the two terms are
mean-zero and orthogonal,
\begin{equation}
\mathbb{E}_{G(\tilde y)}\big[\Ahat\big] = \mathbb{E}_{G(\tilde y)}\big[A^{\tilde y, d}\big] = 0,
\qquad
\mathbb{E}_{G(\tilde y)}\big[\Ahat\, A^{\tilde y, d}\big] = 0,
\label{eq:decomp-orth}
\end{equation}
so that $\mathrm{Var}_{G(\tilde y)}(W) = \mathrm{Var}_{G(\tilde y)}(\Ahat)
+ \mathrm{Var}_{G(\tilde y)}(A^{\tilde y, d})$ splits the variance of $W$
over the belief group into a between-decision and a within-decision part.
Consequently, for the step term of the BAPO assignment before
standardization, $S_{i,t} := \tfrac{1}{2}\big(A^{\tilde y}_{i,t} +
A^{\tilde y, d}_{i,t}\big) = W_{i,t} - \tfrac{1}{2}\big(\Vhat(\tilde y)
+ \Qhat(\tilde y, d_{i,t})\big) = A^{\tilde y, d}_{i,t} +
\tfrac{1}{2}\Ahat(\tilde y, d_{i,t})$ (Eq.~\ref{eq:bapo-adv} with
$\omega = \tfrac{1}{2}$), the difference of its means over the two decisions is
\begin{equation}
\mathbb{E}_{G(\tilde y, \textsc{trust})}[S] -
\mathbb{E}_{G(\tilde y, \textsc{abstain})}[S] \;=\;
\tfrac{1}{2}\big(\Qhat(\tilde y, \textsc{trust}) - \Qhat(\tilde y,
\textsc{abstain})\big),
\label{eq:decomp-choice}
\end{equation}
so the between-decision signal is carried entirely by the decision advantage $\Ahat$.
\end{proposition}

The identities hold before standardization and clipping. A baseline conditioned on the decision as well as the belief, $\Qhat$ itself, would leave the two decisions with equal group means under any fixed belief. The baseline of the BAPO step term (Eq.~\ref{eq:bapo-adv}) is the average of the belief value and the belief-decision value, $\tfrac{1}{2}(\Vhat + \Qhat)$. The within-decision deviation enters once through each of the two streams, and the credit for the decision enters through the belief-relative stream alone. The average shrinks the belief-decision value toward the belief value: the belief-decision group is the sharper comparison, but with $N$ rollouts per task a belief-decision group holds few conclusive steps, and the belief value stabilizes it. The two streams are standardized separately (Appendix~\ref{app:hparams}), which rescales each without changing which steps it compares.

\begin{proof}[Proof of Proposition~\ref{prop:decomp}]
The identity follows by adding and subtracting $\Qhat(\tilde y, d_{i,t})$ in
the definition $A^{\tilde y}_{i,t} = W_{i,t} - \Vhat(\tilde y)$. With $\mu_d$ the decision weights of the statement, mean-zero:
$\mathbb{E}_{G(\tilde y)}[\Ahat] = \sum_d \mu_d (\Qhat(\tilde y, d) -
\Vhat(\tilde y)) = 0$ since the belief-decision groups partition the belief group, and
$\mathbb{E}_{G(\tilde y)}[A^{\tilde y, d}] = \sum_d \mu_d\,
\mathbb{E}_{G(\tilde y, d)}[W - \Qhat(\tilde y, d)] = 0$. Orthogonality:
$\Ahat$ is constant on each belief-decision group, so
$\mathbb{E}_{G(\tilde y)}[\Ahat\, A^{\tilde y, d}] = \sum_d \mu_d
(\Qhat(\tilde y, d) - \Vhat(\tilde y))\, \mathbb{E}_{G(\tilde y, d)}[W -
\Qhat(\tilde y, d)] = 0$. For the variance claim, square the identity and
average over $G(\tilde y)$; the cross term vanishes by orthogonality. For
the separation claim, restricting the mean-zero property of $A^{\tilde y,
d}$ to a single belief-decision group gives $\mathbb{E}_{G(\tilde y, d)}[A^{\tilde y, d}] =
0$, so the mean of $S$ over $G(\tilde y, d)$ is $\tfrac{1}{2}\Ahat(\tilde y, d) =
\tfrac{1}{2}(\Qhat(\tilde y, d) - \Vhat(\tilde y))$; subtracting the two
belief-decision means cancels $\Vhat$.
\end{proof}

\paragraph{Anchor bias.}\label{rem:anchor-bias}
The baselines $\Vhat(\tilde y)$ and $\Qhat(\tilde y, d)$ of
Eq.~\ref{eq:values} are indexed by the step's own forecast and decision,
so they act differently on the tokens written before and after the anchor. The reading below is the idealization in which the group values and the anchor groups are held fixed, rollouts are sampled on-policy without masks, and every forecast lies in $\{\textsc{pass}, \textsc{fail}\}$. The empirical group means of Proposition~\ref{prop:decomp} are their batch estimates. For the action, written after the belief, the anchor is part of the token's context, and a baseline that is a function of the context leaves the expected score-function gradient unchanged. The decision token is baselined by $\Vhat(\tilde y)$ in the same way, and the difference between its two group means is carried by $\Ahat$ (Eq.~\ref{eq:decomp-choice}). For the reasoning $v$, written before the anchor, freeze the downstream kernels and let the forecast be drawn from
$\pi_\theta(\cdot \mid c)$ with $c$ containing $v$; then
\[
\mathbb{E}\Big[\nabla_\theta \log \pi_\theta(v \mid h)\,\big(W -
\Vhat(\ytst)\big)\Big] \;=\; \mathbb{E}\Big[\nabla_\theta \log
\pi_\theta(v \mid h)\, W\Big] \;-\; \nabla_\theta\,
\mathbb{E}\big[\Vhat(\ytst)\big],
\]
the gradient on the right taken through $\pi_\theta(v \mid h)$ alone.
The second term is a shaping term rather than a baseline. It favors reasoning that leads into the belief group of lower value, it vanishes
when the two group values coincide, and its size is bounded by
$|\Vhat(\textsc{pass}) - \Vhat(\textsc{fail})|$ times the expected score norm. The scoring rule of Eq.~\ref{eq:settle} gives correct trusted \textsc{pass} and \textsc{fail} forecasts with clean exits the same bonus, so
the gap between the two group values is set by the batch rather than by the rule. The same reading applies to the decision-relative stream with the belief-decision values. Empirically, the trust rate rises during Stage~II and the forecasts grow more accurate (\S\ref{sec:exp:why}).

\subsection{Belief-augmented imitation}
\label{app:proofs:imitation}

Stage~I maximizes the likelihood of BeliefTraj under the step policy.
Read as imitation, it controls the gap between the policy and the
demonstration process.

\propbeliefimitation*

\begin{proof}
Since $R \in [0,1]$ is a function of the trajectory, $J(P_E) -
J(\pi_\theta) \le \mathrm{TV}(P_E, P_{\pi_\theta})$. Both laws share the
task distribution and factorize over the $2T$ conditionals $u_t \mid
h_t$ and $o_{t+1} \mid (h_t, u_t)$, with $P_E(\cdot \mid h_t)$ and $P_E(o_{t+1} \mid h_t, u_t)$ on one side and $\pi_\theta(\cdot \mid h_t)$ and $P_{\mathrm{env}}(o_{t+1} \mid h_t, a_t)$ on the other, the environment depending on the step message only through $a_t$. Let $P^{(k)}$ be the law that takes the first $k$
conditionals from $P_E$ and the remaining ones from $\pi_\theta$ and the
environment, so $P^{(2T)} = P_E$ and $P^{(0)} = P_{\pi_\theta}$.
Consecutive hybrids differ in one conditional, share the distribution
of its prefix under $P_E$, and share every later kernel; since
adjoining a common continuation kernel while retaining the prefix
preserves total variation, $\mathrm{TV}(P^{(k)}, P^{(k-1)})$ equals the
expected total variation of the $k$-th conditional under the $P_E$
prefix. The triangle inequality
over $k$ gives
\[
\mathrm{TV}(P_E, P_{\pi_\theta}) \le \sum_{t=1}^{T}
\mathbb{E}_{P_E}\big[\mathrm{TV}(P_E(\cdot \mid h_t), \pi_\theta(\cdot \mid h_t))\big]
+ \Lambda .
\]
For the first sum, Pinsker's inequality and Jensen's inequality for the concave square root give $\mathbb{E}[\mathrm{TV}] \le \mathbb{E}[\sqrt{\mathrm{KL}/2}]
\le \sqrt{\varepsilon_t/2}$ with $\varepsilon_t :=
\mathbb{E}_{P_E}[\mathrm{KL}(P_E(\cdot \mid h_t) \,\|\, \pi_\theta(\cdot
\mid h_t))] = \mathbb{E}_{P_E}[-\log \pi_\theta(u_t \mid h_t)] -
\mathbb{E}_{P_E}[H(P_E(\cdot \mid h_t))]$, the step-$t$ excess
cross-entropy of the Stage~I loss (\S\ref{sec:method:sft}), whose sum
over $t$ is $\varepsilon_{\mathrm{SFT}}(\theta)$. Finally $\sum_t \sqrt{\varepsilon_t/2} \le
\sqrt{T \sum_t \varepsilon_t / 2}$ by the Cauchy--Schwarz inequality.
\end{proof}

The bound of Proposition~\ref{prop:imitation} separates what Stage~I optimizes from what it cannot touch.
The first term is driven down by the likelihood objective, but only on the demonstration process's own histories. The second term is a property of the corpus and of the environment. BeliefTraj writes the parsed future outcome back into $\beta_t$, so under $P_E$ the next observation given $(h_t, u_t)$ is supported only on observations consistent with that label. The environment law given $(h_t, a_t)$ instead mixes over the posterior of the latent repository state. The two agree only when the history and the action already determine the outcome of the step, and $\Lambda$ (Proposition~\ref{prop:imitation}) measures the information the hindsight labels carry beyond the history and the action; it is zero exactly when the two observation kernels agree $P_E$-almost surely at every step. For BeliefTraj, $J(P_E) = 1$ because the
corpus contains solving trajectories only. The bound then reads as a statement about what can separate the fine-tuned policy from that ceiling:
the corpus term, with the compounding term on top of it. This is
the formal counterpart of \S\ref{sec:method:sft}, where Stage~I leaves the prediction factor calibrated to corpus histories. Stage~II, whose data are drawn causally from the environment at the policy's own histories, addresses that gap.

\subsection{Task reward under a world-model accuracy budget}
\label{app:proofs:objective}

The policy term of Eq.~\ref{eq:objective} is the clipped, token-level objective motivated by the standard step-level surrogate
\citep{schulman2015trpo,schulman2017ppo}, with the group baseline of
\S\ref{sec:method:bapo} in place of a learned value function
\citep{shao2024deepseekmath}. Up to the distribution-shift term of the lemma, the surrogate lower-bounds the performance of the updated policy.

\begin{lemma}[Step-level surrogate bound]
\label{lem:surrogate}
For step policies $\pi$ and $\pi'$ with $\pi'(\cdot \mid h) \ll \pi(\cdot
\mid h)$ for every history $h$ (as holds for full-support softmax
policies), define the value functions of $\pi$ recursively by
$V^\pi(h_{T+1}) := R$, $Q^\pi(h_t, \beta, a) := \mathbb{E}_{o_{t+1} \sim
P_{\mathrm{env}}(\cdot \mid h_t, a)}\big[V^\pi(h_t, \beta, a,
o_{t+1})\big]$ and $V^\pi(h_t) := \mathbb{E}_{(\beta, a) \sim \pi(\cdot
\mid h_t)}\big[Q^\pi(h_t, \beta, a)\big]$, so that $Q^\pi$ is defined for
every emission, and let $A^\pi := Q^\pi - V^\pi \in [-1, 1]$. Define the
surrogate
\begin{align}
L_\pi(\pi') &:= J(\pi) + \sum_{t=1}^{T} \mathbb{E}_{h_t \sim P_\pi}\,
\mathbb{E}_{(\beta, a) \sim \pi'(\cdot \mid h_t)}\big[A^\pi(h_t, \beta, a)\big]
\label{eq:surrogate-def}\\
&\;= J(\pi) + \sum_{t=1}^{T} \mathbb{E}_{h_t \sim P_\pi}\,
\mathbb{E}_{(\beta, a) \sim \pi(\cdot \mid h_t)}\Big[\tfrac{\pi'(\beta, a \mid h_t)}{\pi(\beta, a \mid h_t)}\,
A^\pi(h_t, \beta, a)\Big],
\nonumber
\end{align}
the second form being the joint-emission importance-sampling
representation that underlies the token-level clipped estimator in the
policy term of Eq.~\ref{eq:objective}.
With $\Delta_{\mathrm{TV}} := \sup_h \mathrm{TV}(\pi'(\cdot \mid h),
\pi(\cdot \mid h))$ and $\Delta_{\mathrm{KL}} := \sup_h
\mathrm{KL}(\pi(\cdot \mid h) \,\|\, \pi'(\cdot \mid h))$,
\begin{equation}
J(\pi') \;\ge\; L_\pi(\pi') - T(T-1)\,\Delta_{\mathrm{TV}} \;\ge\;
L_\pi(\pi') - T(T-1)\sqrt{\Delta_{\mathrm{KL}}/2}.
\label{eq:surrogate-bound}
\end{equation}
\end{lemma}

\begin{proof}
By the recursive definitions, $\mathbb{E}\big[V^\pi(h_{t+1}) -
Q^\pi(h_t, \beta_t, a_t) \,\big|\, h_t, \beta_t, a_t\big] = 0$ under
either policy, because the observation is drawn from the shared kernel
$P_{\mathrm{env}}$, and $\mathbb{E}_{P_{\pi'}}[V^\pi(h_1)] = J(\pi)$
because $h_1$ is the task alone. Telescoping $R - V^\pi(h_1) =
\sum_{t=1}^{T} \big(V^\pi(h_{t+1}) - V^\pi(h_t)\big)$ along a trajectory
of $\pi'$ and taking expectations gives the performance difference
identity $J(\pi') - J(\pi) = \sum_{t=1}^{T} \mathbb{E}_{h_t \sim
P_{\pi'}} \mathbb{E}_{(\beta, a) \sim \pi'(\cdot \mid h_t)} [A^\pi(h_t,
\beta, a)]$ \citep{kakade2002approximately}. Writing $g_t(h) :=
\mathbb{E}_{(\beta,a) \sim \pi'(\cdot \mid h)}[A^\pi(h, \beta, a)] \in
[-1, 1]$, the surrogate differs from the identity only in the law of
$h_t$, so $|J(\pi') - L_\pi(\pi')| \le \sum_{t=1}^{T} 2\,
\mathrm{TV}(P_{\pi'}(h_t), P_\pi(h_t))$. The two laws of $h_t$ share the
task distribution and all environment kernels and differ in the $t-1$
policy conditionals before step $t$; the hybrid argument of
Proposition~\ref{prop:imitation} bounds their total variation by
$\sum_{s < t} \mathbb{E}_{P_\pi}[\mathrm{TV}(\pi'(\cdot \mid h_s),
\pi(\cdot \mid h_s))] \le (t-1)\Delta_{\mathrm{TV}}$. Summing,
$|J(\pi') - L_\pi(\pi')| \le 2\Delta_{\mathrm{TV}} \sum_{t=1}^{T} (t-1)
= T(T-1)\Delta_{\mathrm{TV}}$, and Pinsker's inequality gives the second
form.
\end{proof}

The policy term differs from the surrogate of Eq.~\ref{eq:surrogate-def}
in three ways. The ratio is factorized over tokens rather than taken per emission. The advantage is a group-baseline estimate from the task's own rollouts rather than $A^\pi$. The return is the shaped one of the repository instantiation, combining task outcome, bookkeeping term, and decision rewards; its width is at most $1.4 + T(\alpha + \max(\gamma, \kappa))$ in place of the unit range in the constant of Eq.~\ref{eq:surrogate-bound}, and the decision return is what BAPO measures against belief-conditional baselines (Proposition~\ref{prop:decomp}).
The clip range $(\epsilon_{\mathrm{lo}}, \epsilon_{\mathrm{hi}})$ of Eq.~\ref{eq:objective} plays the role of the trust
region, as in PPO \citep{schulman2017ppo}. The remaining question is what the prediction term does to the optimum.

\begin{proposition}[Multiplier as accuracy budget]
\label{prop:lagrangian}
Fix $\lambda \ge 0$, let $\bar{\mathcal L}$ be the expected prediction loss of
Eq.~\ref{eq:excess-risk} averaged over supervised steps, and suppose that for every multiplier considered the scalarized objective $J(\theta) - \lambda\, \bar{\mathcal L}(\theta)$ attains a global maximum $\theta_\lambda$ with finite expected prediction loss. Then $\theta_\lambda$ is optimal for the constrained
program
\[
\begin{aligned}
\max_\theta\quad & J(\theta)\\
\mathrm{s.t.}\quad & \bar{\mathcal L}(\theta) \le B(\lambda),
\end{aligned}
\]
where $B(\lambda) := \bar{\mathcal L}(\theta_\lambda)$, and the budget is monotone: for $\lambda' > \lambda$, $B(\lambda') \le
B(\lambda)$ and $J(\theta_{\lambda'}) \le J(\theta_\lambda)$.
\end{proposition}

\begin{proof}
If some $\theta$ had $\bar{\mathcal L}(\theta) \le \bar{\mathcal
L}(\theta_\lambda)$ and $J(\theta) > J(\theta_\lambda)$, then $J(\theta) -
\lambda \bar{\mathcal L}(\theta) > J(\theta_\lambda) - \lambda
\bar{\mathcal L}(\theta_\lambda)$, contradicting the optimality of
$\theta_\lambda$. For monotonicity, optimality of $\theta_\lambda$ and of
$\theta_{\lambda'}$ for their own multipliers gives
$J(\theta_\lambda) - \lambda \bar{\mathcal L}(\theta_\lambda) \ge
J(\theta_{\lambda'}) - \lambda \bar{\mathcal L}(\theta_{\lambda'})$ and
$J(\theta_{\lambda'}) - \lambda' \bar{\mathcal L}(\theta_{\lambda'}) \ge
J(\theta_\lambda) - \lambda' \bar{\mathcal L}(\theta_\lambda)$; adding
them yields $(\lambda' - \lambda)\big(\bar{\mathcal L}(\theta_\lambda) -
\bar{\mathcal L}(\theta_{\lambda'})\big) \ge 0$, hence $\bar{\mathcal
L}(\theta_{\lambda'}) \le \bar{\mathcal L}(\theta_\lambda)$, and
substituting into the first inequality gives $J(\theta_\lambda) -
J(\theta_{\lambda'}) \ge \lambda\big(\bar{\mathcal L}(\theta_\lambda) -
\bar{\mathcal L}(\theta_{\lambda'})\big) \ge 0$.
\end{proof}

Training with a frozen multiplier is the special case of
reward-constrained policy optimization
\citep{altman1999constrained,tessler2019rcpo}: the multiplier indexes a monotone family of accuracy budgets.

\subsection{Information-theoretic guarantees of the prediction loss}
\label{app:proofs:information}

Proposition~\ref{prop:variational} relates the expected prediction loss of Eq.~\ref{eq:excess-risk} to what the belief knows.
Corollaries~\ref{cor:pricing} and~\ref{cor:traj} then show that the forecast divergence bounds the error of a decision reward evaluated under the belief.
Proposition~\ref{prop:selection}, proved here, and Corollary~\ref{cor:pointsel} bound the regret of choosing among candidates by their forecasts.

\begin{proposition}[Prediction-loss information bounds]
\label{prop:variational}
For every supervised step,
\begin{equation}
H(y_t \mid c_t) \;\le\; \bar{\mathcal L}_t(\theta),
\qquad
I(c_t;\, y_t \mid a_t) \;\ge\; H(y_t \mid a_t) - \bar{\mathcal L}_t(\theta),
\label{eq:info-lower}
\end{equation}
and
\begin{equation}
I(c_t;\, y_t \mid a_t) \;\le\; I(c_t;\, o_{t+1} \mid a_t) \;\le\;
I(c_t;\, s_{t+1} \mid a_t),
\label{eq:info-dpi}
\end{equation}
and, for the written part of the context, the reasoning and the summary
together,
\begin{equation}
I((v_t, z_t);\, y_t \mid h_t) \;\ge\; H(y_t \mid h_t) - \bar{\mathcal L}_t(\theta).
\label{eq:info-summary}
\end{equation}
\end{proposition}

\begin{proof}
For any conditional distribution $q(\cdot \mid c)$, $\mathbb{E}[-\log
q(y_t \mid c_t)] = H(y_t \mid c_t) + \mathbb{E}[\mathrm{KL}(p(\cdot \mid
c_t) \,\|\, q(\cdot \mid c_t))] \ge H(y_t \mid c_t)$, the variational
bound of \citet{barber2003im}; with $q = q_\theta$ this is
Eq.~\ref{eq:excess-risk}. Conditioning reduces entropy, so $H(y_t \mid
c_t, a_t) \le H(y_t \mid c_t) \le \bar{\mathcal L}_t(\theta)$, and
$I(c_t; y_t \mid a_t) = H(y_t \mid a_t) - H(y_t \mid c_t, a_t)$ gives
the second inequality of Eq.~\ref{eq:info-lower}. For
Eq.~\ref{eq:info-dpi}: the context $c_t$ is formed before $o_{t+1}$
exists; given $(s_{t+1}, a_t)$ the observation is drawn from $\Omega(\cdot \mid s_{t+1}, a_t)$ independently of $c_t$ (under supervised-step conditioning the kernel is restricted to verifier-valid observations, and since parsing depends on the emitted message and validity on $(o_{t+1}, a_t)$, the independence is preserved), and given
$(o_{t+1}, a_t)$ the outcome $y_t = \psi(o_{t+1}, \chi(a_t))$ is
deterministic, so conditionally on $a_t$ the chain $c_t \to s_{t+1} \to
o_{t+1} \to y_t$ is Markov and both inequalities are the conditional
data processing inequality. For Eq.~\ref{eq:info-summary}, $H(y_t \mid
h_t, v_t, z_t) = H(y_t \mid c_t) \le \bar{\mathcal L}_t(\theta)$ and
$I((v_t, z_t); y_t \mid h_t) = H(y_t \mid h_t) - H(y_t \mid h_t, v_t,
z_t)$.
\end{proof}

The expected prediction loss is an upper bound on $H(y_t \mid c_t)$, the uncertainty about the checkable
outcome that remains at the moment the block is written. Minimizing it therefore lowers the forecast divergence at a fixed outcome law and, through the policy's own choices of context and action, bounds the uncertainty left at writing time. In turn, $H(y_t \mid a_t) - \bar{\mathcal L}_t$ is a lower bound on the information the writing
context carries about the outcome of the committed action, hence, by
Eq.~\ref{eq:info-dpi}, about the latent state that action leads to. The
conditioning on $a_t$ is what makes the statement about the state
rather than about the action: unconditionally, the context is trivially
informative about the action the policy is about to take. Eqs.~\ref{eq:info-lower} and~\ref{eq:info-dpi} are about the context $c_t$ in which the value tokens are generated; the predictor reads the whole history, and the belief is the
policy's own summary of that history rather than a separate module.
Eq.~\ref{eq:info-summary} isolates the written part of the context. Whatever the expected prediction loss leaves of $H(y_t \mid h_t)$, the
uncertainty that the raw history alone leaves about the checkable
outcome, is information carried by the reasoning and the written summary, including what they induce through the chosen action. This is the objective used to
train belief modules for imitation learning in POMDPs
\citep{gangwani2020bmil}, with two changes: the belief lives in
the policy's emitted tokens, and the target is a parsed, verifiable
outcome rather than the raw observation, which removes the need for a
learned decoder. The classical reason for wanting the context to be
informative about the latent state is that optimal POMDP policies act
on beliefs, sufficient statistics of the history
\citep{astrom1965optimal,kaelbling1998planning} (Figure~\ref{fig:app-pomdp}). The forecast divergence
$\mathbb{E}[\delta(c_t)] = \bar{\mathcal L}_t - H(y_t \mid c_t)$ is the
part of the expected prediction loss the objective can remove; the entropy term is set by
the environment and by the policy's own action distribution.

\begin{corollary}[Decision-reward error bound]
\label{cor:pricing}
Fix a supervised step with writing context $c$ and predictive $q =
q_\theta(\cdot \mid c)$. Let $p = p(\cdot \mid c)$ be the outcome law at
the writing context, before the belief is committed, and $p_{\hat y, d}
= p(\cdot \mid c, \hat y, d)$ the law under which the decision reward is realized once the belief $(\hat y, d)$, drawn from the policy's law at $c$ under $\mathcal{M}$, has been written and acted on, all laws taken under the same conditioning as $\delta$ and $I$. Write $\delta = \mathrm{KL}(p \,\|\, q)$ for
the forecast divergence and $I := I\big((\hat y, d);\, y \mid c\big)$ for the
information the committed belief carries about the outcome beyond the
context. For any payoff $u : \mathcal{Y} \to \mathbb{R}$ with
$\|u\|_\infty \le U$,
\[
\bigl|\, \mathbb{E}_p[u] - \mathbb{E}_q[u] \,\bigr| \;\le\; U \sqrt{2\delta},
\qquad
\mathbb{E}_{(\hat y, d)}\,\bigl|\, \mathbb{E}_{p_{\hat y, d}}[u] -
\mathbb{E}_q[u] \,\bigr| \;\le\; U \sqrt{2(\delta + I)}.
\]
In particular, averaged over the committed beliefs, the expected decision reward (Eq.~\ref{eq:settle}, payoffs bounded by $\alpha \vee \gamma \vee \kappa$ and zero on inconclusive outcomes) evaluated under the belief differs from its expectation under the realized law by at most $(\alpha \vee \gamma \vee
\kappa)\sqrt{2(\delta + I)}$.
\end{corollary}
The prediction loss controls $\delta$, while $I$ is zero
when the belief is a deterministic function of the context and positive
whenever the sampled belief steers the action toward different
outcomes; it is at most the action-information term of
Eq.~\ref{eq:candidate-risk}, with equality when the action adds no
information about the outcome beyond the belief.

\begin{proof}
For the marginal law, $|\mathbb{E}_p[u] - \mathbb{E}_q[u]| \le
\|u\|_\infty \sum_y |p(y) - q(y)| \le 2U\,\mathrm{TV}(p, q) \le
U\sqrt{2\,\mathrm{KL}(p \| q)}$ by Pinsker's inequality. For the
realized law, the chain rule of relative entropy gives
$\mathbb{E}_{(\hat y, d)}\big[\mathrm{KL}(p_{\hat y, d} \,\|\, q)\big]
= \mathbb{E}_{(\hat y, d)}\big[\mathrm{KL}(p_{\hat y, d} \,\|\, p)\big]
+ \mathrm{KL}(p \,\|\, q) = I + \delta$, since $q$ does not depend on
$(\hat y, d)$ and $\mathbb{E}_{(\hat y, d)}[p_{\hat y, d}] = p$; Pinsker
applied to each $p_{\hat y, d}$ and Jensen's inequality for the concave
square root give the second bound.
\end{proof}

\begin{corollary}[Trajectory-level cost bound]
\label{cor:traj}
Fix a rollout $i$ and a nonempty set $S$ of
supervised steps at which predictions are consumed, with forecast divergences
$\delta_t$ for $t \in
S$ and average $\bar\delta := |S|^{-1} \sum_{t \in S} \delta_t$. For per-step payoffs with $\|u_t\|_\infty \le U$,
\[
\sum_{t \in S} \bigl|\, \mathbb{E}_{p_t}[u_t] - \mathbb{E}_{q_t}[u_t]
\,\bigr| \;\le\; U \sum_{t \in S} \sqrt{2\delta_t} \;\le\; |S|\, U
\sqrt{2\bar\delta}.
\]
\end{corollary}
The cost of consuming predictions along a rollout is therefore controlled by the average forecast divergence over the steps at which predictions are consumed, rather than by its worst case. Averaged over the commitments at each step, the same bound holds under the committed-belief laws with $\delta_t + I_t$ in place of $\delta_t$, where $I_t$ is the information term of Corollary~\ref{cor:pricing} at step $t$.

\begin{proof}
Apply Corollary~\ref{cor:pricing} at each step of $S$ and sum; the
second inequality is Jensen's inequality for the concave square root.
\end{proof}

\proppredregret*
\begin{proof}[Proof of Proposition~\ref{prop:selection}]
Write $c^{(k)}$ for the writing context of candidate $k$, so that $q_k =
q_\theta(\cdot \mid c^{(k)})$ and $p_k = p(\cdot \mid h, a^{(k)})$.
Since $\hat v_{\hat k} \ge \hat v_{k^\ast}$,
$v_{k^\ast} - v_{\hat k} \le (v_{k^\ast} - \hat v_{k^\ast}) + (\hat
v_{\hat k} - v_{\hat k})$, and since $u$ takes values in $[0, U]$ each bracket is at most $U\,\mathrm{TV}(p_k, q_k) \le U\sqrt{\delta_k/2}$ by Pinsker's inequality, applied to candidate $k^\ast$ and to candidate $\hat k$.
\end{proof}

\begin{corollary}[Point-forecast regret bound]
\label{cor:pointsel}
In the setting of Proposition~\ref{prop:selection}, let each candidate
be scored by a single point forecast, $\tilde v_k := u(\hat y^{(k)})$
with $\hat y^{(k)}$ the forecast the procedure emits for candidate
$k$, and let $\tilde k \in \argmax_k
\tilde v_k$. With $\zeta := \mathbb{E}\big[\max_k
|u(\hat y^{(k)}) - \hat v_k|\big]$ the expected largest deviation of a drawn score from the believed value,
\begin{equation}
\mathbb{E}\big[v_{k^\ast} - v_{\tilde k}\big] \;\le\; U\,\mathbb{E}\Big[\sqrt{2\max_k \delta_k}\Big] \;+\; 2\zeta ,
\label{eq:pointsel-regret}
\end{equation}
the expectation taken over the generation of the candidates, under which the $\delta_k$ are random.
\end{corollary}
The second term is the outcome spread at uncertain steps: it is zero
when every $q_k$ puts all its mass on one score level, and it need not vanish as the forecast divergences $\delta_k$ vanish, since a candidate whose
outcome is genuinely uncertain can be outscored by a single draw even
under a perfect predictive.
\begin{proof}
Write $v_{k^\ast} - v_{\tilde k} = (v_{k^\ast} - \hat v_{k^\ast}) +
(\hat v_{k^\ast} - \tilde v_{k^\ast}) + (\tilde v_{k^\ast} - \tilde
v_{\tilde k}) + (\tilde v_{\tilde k} - \hat v_{\tilde k}) + (\hat
v_{\tilde k} - v_{\tilde k})$. The third bracket is at most zero by the
choice of $\tilde k$; the first and last are bounded as in Proposition~\ref{prop:selection}; the second and fourth are each at most $\max_k |u(\hat y^{(k)}) - \hat v_k|$ pointwise, and taking expectations gives $2\zeta$.
\end{proof}

Two points connect Proposition~\ref{prop:selection} to the training procedure. First,
the forecast divergence of a candidate is measured against the outcome of its
own action, whereas the predictive was written before that action. The two are tied by an identity that holds because the outcome is independent of the written reasoning and summary given the history and the action,
\begin{equation}
\mathbb{E}_{a \sim P(\cdot \mid c, \mathcal{M})}\Big[\mathrm{KL}\big(p(\cdot
\mid h, a)\,\big\|\,q_\theta(\cdot \mid c)\big)\Big] \;=\; \delta(c) +
I(a_t;\, y_t \mid c_t = c),
\label{eq:candidate-risk}
\end{equation}
so, in expectation over the action distribution, the candidates' forecast divergences are small when the trained forecast divergence is small and the writing context already determines what the committed action will do; the second term is the cost of predicting before
acting. Second, ANTE-TTO (\S\ref{sec:method:validator},
Algorithm~\ref{alg:tto}, $U = 4$ for the score of
Eq.~\ref{eq:tto-score}) scores each candidate by the plug-in $\tilde v_k$ of Corollary~\ref{cor:pointsel}, with the decoder greedy for the original continuation and sampling at temperature one for the alternatives, and with $\zeta$ taken under that decoder, so its regret obeys Eq.~\ref{eq:pointsel-regret}. The bound says that the part of the regret attributable to the world model is controlled by the forecast divergence. A smaller candidate divergence tightens the bound.

\paragraph{Instantiation across harnesses.}\label{rem:harness}
Every statement above depends on the environment through the action-type map $\chi$, the verifier $\psi$ of Appendix~\ref{sec:prelim}, and the reward the harness assigns. On
repository issues $\psi$ reads the check field and the exit field, and no constant in the propositions references the
harness otherwise. The terminal instantiation
(Appendix~\ref{app:terminal-config}) keeps the exit field as the
per-step coordinate that $\psi$ reads from the next observation and
anchors the belief groups on its forecast, so the per-step statements apply there to that coordinate and to rewards that depend on it, with the terminal verifier and scoring rule in place.

\FloatBarrier
\section{Corpus curation}\label{app:corpus}

\begin{figure}[!htb]
\centering
\includegraphics[width=\textwidth]{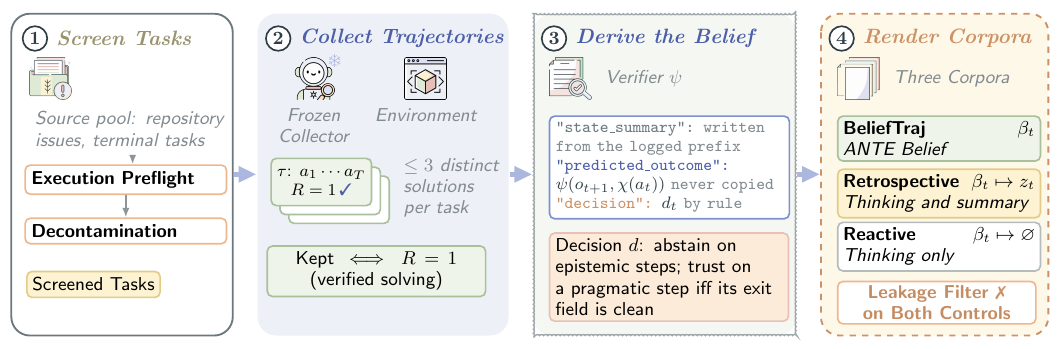}
\caption{\textbf{BeliefTraj curation.} Screening, collection in containers, and label derivation; counts in Table~\ref{tab:funnel}.}
\label{fig:app-curation}
\end{figure}

BeliefTraj is a corpus of solving trajectories collected by a capable agent in the execution environment, with its forecast labels read from execution. The paired corpora for Reactive and Retrospective use the same trajectories, without the belief block or with only its summary. This section covers task screening, label derivation and its checks, the paired construction with its protocol-leakage filter, and the corpus statistics behind Table~\ref{tab:main}. Figure~\ref{fig:app-curation} maps the pipeline, and Table~\ref{tab:funnel} provides the stage counts.

\begin{table}[h]
\centering
\caption{\textbf{Task screening counts.}}
\label{tab:funnel}
\small
\begin{tabular}{@{}lr@{}}
\toprule
Stage & Count \\
\midrule
Source pool of repository-issue tasks & 13{,}558 \\
Reproduce in their containers (execution preflight) & 10{,}662 \\
Solved by the trajectory collector & 6{,}746 \\
Verified solving trajectories (quality gates) & 12{,}500 over 6{,}729 tasks \\
Training corpora (ANTE, Retrospective, Reactive) & 3 $\times$ 12{,}500 trajectories \\
\bottomrule
\end{tabular}
\end{table}

\subsection{Task screening and trajectory collection}
Before any model call, the source pool (Table~\ref{tab:funnel}) passes an execution preflight, in which every task must reproduce in its own container, and decontamination against the evaluation benchmarks. A strong collector (an LLM such as Qwen3-Coder-480B-A35B \citep{qwen2025qwen3coder480b} or DeepSeek-V3.2 \citep{deepseek2025v32}) then attempts the tasks in the containers, and only verified solving trajectories survive: BeliefTraj keeps 12{,}500 of them over 6{,}729 repository tasks and 30{,}161 over 10{,}823 terminal tasks.
Table~\ref{tab:pools} gives the composition of the three source pools, R2E-Gym \citep{jain2025r2egym}, SWE-rebench \citep{badertdinov2025swerebench}, and SWE-Gym \citep{pan2024swegym}, including the RL pool of 6{,}566 unique instances.

\begin{table}[h]
\centering
\caption{\textbf{RL pool composition.} Per-pool composition of the repository-issue tasks after the execution preflight.}
\label{tab:pools}
\small
\begin{tabular}{@{}lrrrr@{}}
\toprule
Pool & Executable tasks & Solved in collection & Solve rate & In RL pool \\
\midrule
R2E-Gym & 4{,}555 & 3{,}919 & 86.0\% & 3{,}880 \\
SWE-rebench & 4{,}456 & 1{,}829 & 41.0\% & 1{,}767 \\
SWE-Gym & 1{,}651 & 998 & 60.4\% & 919 \\
\midrule
Total & 10{,}662 & 6{,}746 & 63.3\% & 6{,}566 \\
\bottomrule
\end{tabular}
\end{table}

\begin{figure}[!t]
\footnotesize
\noindent\textbf{Step 1.}\; Thinking: ``Let me explore the repo structure to find the URL dispatcher code.''\par\vspace{2pt}
\noindent\begin{minipage}[t]{0.16\textwidth}\raggedright
\textcolor{trajred}{\textbf{Reactive}}\par\vspace{1pt}
\begin{trajnone}---\end{trajnone}
\end{minipage}\hfill
\begin{minipage}[t]{0.36\textwidth}\raggedright
\textcolor{trajamber}{\textbf{Retrospective}}\par\vspace{1pt}
\begin{trajmem}<memory\_state>\allowbreak\{"state\_summary": "Starting to investigate the aiohttp URL dispatcher issue where routes with variables after a fixed path segment fail to match."\}\allowbreak</memory\_state>\end{trajmem}
\end{minipage}\hfill
\begin{minipage}[t]{0.44\textwidth}\raggedright
\textcolor{paloalto}{\textbf{ANTE (BeliefTraj)}}\par\vspace{1pt}
\begin{trajbelief}<belief\_state>\allowbreak\{"state\_summary": "Starting to investigate the aiohttp URL dispatcher issue where routes with variables after a fixed path segment fail to match.", "predicted\_outcome": \{"target\_test": "NA", "exit\_ok": true\}, "decision": "trust"\}\allowbreak</belief\_state>\end{trajbelief}
\end{minipage}\par\vspace{2pt}
{\raggedright\noindent Tool call: \texttt{<function=search> def index\_key \textperiodcentered\ /testbed/aiohttp/web\_urldispatcher.py}\par}
{\centering $\cdots$\ Steps 2--4\ $\cdots$\par}\vspace{2pt}
\noindent\textbf{Step 5.}\; Thinking: ``Let me run this to confirm the failure.''\par\vspace{2pt}
\noindent\begin{minipage}[t]{0.16\textwidth}\raggedright
\textcolor{trajred}{\textbf{Reactive}}\par\vspace{1pt}
\begin{trajnone}---\end{trajnone}
\end{minipage}\hfill
\begin{minipage}[t]{0.36\textwidth}\raggedright
\textcolor{trajamber}{\textbf{Retrospective}}\par\vspace{1pt}
\begin{trajmem}<memory\_state>\allowbreak\{"state\_summary": "Created reproduce\_issue.py to confirm the 404 for /core/\allowbreak locations/\allowbreak tail/\allowbreak here."\}\allowbreak</memory\_state>\end{trajmem}
\end{minipage}\hfill
\begin{minipage}[t]{0.44\textwidth}\raggedright
\textcolor{paloalto}{\textbf{ANTE (BeliefTraj)}}\par\vspace{1pt}
\begin{trajbelief}<belief\_state>\allowbreak\{"state\_summary": "Created reproduce\_issue.py to confirm the 404 for /core/\allowbreak locations/\allowbreak tail/\allowbreak here.", "predicted\_outcome": \{"target\_test": "FAIL", "exit\_ok": true\}, "decision": "abstain"\}\allowbreak</belief\_state>\end{trajbelief}
\end{minipage}\par\vspace{2pt}
{\raggedright\noindent Tool call: \texttt{<function=execute\_bash> cd /testbed \&\& python reproduce\_issue.py}\par}
{\centering $\cdots$\ Steps 6--14\ $\cdots$\par}\vspace{2pt}
\noindent\textbf{Step 15 (Submission).}\; Thinking: ``The task is complete. The fix resolves the issue and is verified. \ldots''\par\vspace{2pt}
\noindent\begin{minipage}[t]{0.16\textwidth}\raggedright
\textcolor{trajred}{\textbf{Reactive}}\par\vspace{1pt}
\begin{trajnone}---\end{trajnone}
\end{minipage}\hfill
\begin{minipage}[t]{0.36\textwidth}\raggedright
\textcolor{trajamber}{\textbf{Retrospective}}\par\vspace{1pt}
\begin{trajmem}<memory\_state>\allowbreak\{"state\_summary": "Confirmed the 3 static-permission test failures are pre-existing (fail on unmodified code too, since tests run as root); my index-key fix resolves the issue with all matching tests passing and the reproduce script printing VERDICT: PASS."\}\allowbreak</memory\_state>\end{trajmem}
\end{minipage}\hfill
\begin{minipage}[t]{0.44\textwidth}\raggedright
\textcolor{paloalto}{\textbf{ANTE (BeliefTraj)}}\par\vspace{1pt}
\begin{trajbelief}<belief\_state>\allowbreak\{"state\_summary": "Confirmed the 3 static-permission test failures are pre-existing (fail on unmodified code too, since tests run as root); my index-key fix resolves the issue with all matching tests passing and the reproduce script printing VERDICT: PASS.", "predicted\_outcome": \{"target\_test": "PASS", "exit\_ok": true\}, "decision": "trust"\}\allowbreak</belief\_state>\end{trajbelief}
\end{minipage}\par\vspace{2pt}
{\raggedright\noindent Tool call: \texttt{<function=finish> submit \textperiodcentered\ "Fixed the URL dispatcher's route indexing in /testbed/aiohttp/web\_urldispatcher.py. \ldots"}\par}
\caption{\textbf{Trajectory samples from the SFT corpora.} Steps 1, 5, and 15 (the submission) of one recorded run, with long passages elided.}
\label{fig:ladder}
\end{figure}

\subsection{Execution-derived labels and checks}
\paragraph{Label derivation.}
The verifier $\psi$ re-derives the prediction labels from the logged observation, overriding the collector's own claim on $59{,}212$ check-field and $17{,}407$ exit-field labels.

\paragraph{Decision labels.}
The decision labels follow a fixed rule: $d_t = \textsc{abstain}$ on
every epistemic step, whose check is still open, and \textsc{trust} on pragmatic steps exactly when
the re-derived exit field is clean. A demonstration carries no counterfactual for the untaken decision, so the label fixes the format and leaves calibration
to Stage~II (\S\ref{sec:exp:decision}).

\paragraph{Reward monitoring: test-tampering checks.}
We monitor test-file activity during training with an over-inclusive screen that counts every command touching a test file
or a test entry point. The screen is observational and never enters
the reward. None of the $2{,}442$ commands it catches is reward hacking, and the task reward comes from hidden tests in a fresh container the agent cannot reach.

\subsection{Paired corpus construction and protocol-leakage filtering}
\paragraph{Corpus construction.}
Each corpus trajectory is written three times from one recorded run,
sharing the trajectory's tool calls and, apart from the cleaning below, its reasoning (Figure~\ref{fig:ladder}): Reactive carries no block, Retrospective the summary alone in a \texttt{<memory\_state>} block, and ANTE (BeliefTraj) the summary, the predicted outcome, and the decision.
The two control corpora then pass a protocol-leakage filter: reasoning
that references the belief protocol, its prediction fields, or the
decision vocabulary is removed or rewritten, so that neither
control variant sees the protocol it is meant to ablate. The block is
inserted at the same position in every assistant step, immediately before the tool call, so the
variants differ by the protocol block and by this cleaning alone.

\paragraph{Decontamination.} Decontamination finds no overlap with SWE-bench Verified at the task and repository levels, and, for the RL pool, none with
SWE-bench Lite at the instance, owner/repository, and R2E-Gym-repository
levels.

\subsection{Corpus statistics and evaluation task lists}
\paragraph{Per-task solution counts.}
Up to three distinct verified solutions are kept per task. On the
repository side the split is $1{,}003$ tasks with one solution, $5{,}681$
with two, and $45$ with three; on the terminal side $785$, $738$, and
$9{,}300$ respectively. 

\begin{figure}[h]
\centering
\includegraphics[width=\linewidth]{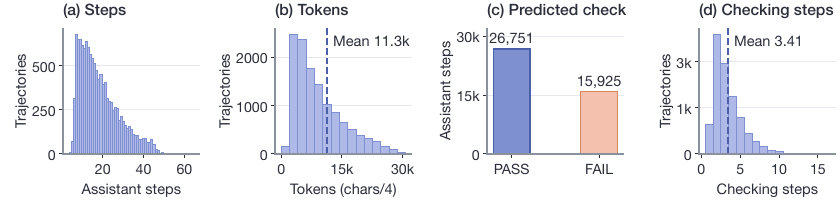}
\caption{\textbf{Training corpus distributions.} (a) Assistant steps
per trajectory. (b) Tokens per trajectory (characters/4 estimate). (c) Predicted \textsc{pass} and \textsc{fail} counts. (d) Steps with a definite predicted check outcome, per trajectory.}
\label{fig:corpus}
\end{figure}

\paragraph{Step statistics.}
Figure~\ref{fig:corpus} profiles the repository training corpus (12{,}500 trajectories).
Trajectories average $11{,}962$ tokens
under the training tokenizer (median $10{,}073$, p95 $26{,}023$, max
$32{,}666$, inside the 32k window). Every assistant step carries exactly
one parsed belief state. Steps that predict a definite check outcome make up 18.5\% of assistant steps (42{,}676: 26{,}751 \textsc{pass} and 15{,}925 \textsc{fail}). Of these, the 30{,}178 test steps whose observation the verifier evaluated (13.1\%) are the conclusive steps, where the decision is scored; the rest are submission steps, whose check field keeps the collector's verdict. The action-type map $\chi$ marks 47{,}997 steps epistemic (20.8\%).

\paragraph{Terminal evaluation sets.}
We build TB-Dev following the Internal-dev recipe of ECHO \citep{shrivastava2026echo}. TB-Dev collects the 73 data-processing, systems-operations, and development-tooling tasks from Terminal-Bench 1.0 \citep{tbench2025} and OpenThoughts-TB-dev \citep{openthoughts2025tbdev} that share neither a name nor a near-duplicate instruction with Terminal-Bench 2.0 or TB-Lite.

\FloatBarrier
\section{Belief state specification and outcome evaluation}\label{app:format}

This section introduces the details of the belief block's schema, the map from a response to an outcome, and the decision reward for each harness.

\subsection{Fields of the belief state}
\begin{figure}[h]
\centering
\includegraphics[width=\textwidth]{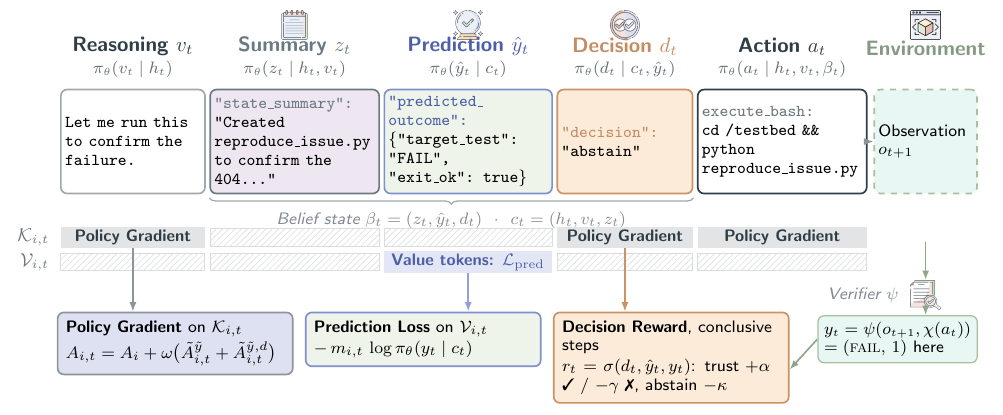}
\caption{\textbf{Anatomy of a step message.} The belief state $\beta_t = (z_t, \hat y_t, d_t)$ is between the reasoning and the tool call in one pass. The prediction loss reaches the forecast's value tokens, and the policy gradient reaches the reasoning, decision, and action tokens. The verifier $\psi$ maps $o_{t+1}$ to the realized outcome $y_t$, which checks the forecast and settles the decision.}
\label{fig:anatomy}
\end{figure}

The belief is emitted as a tagged JSON block immediately before the
tool call, in the same assistant message (Figure~\ref{fig:anatomy}). \texttt{state\_summary} carries the summary $z_t$; \texttt{predicted\_outcome.target\_test} is $\ytst_t$;
\texttt{predicted\_outcome.exit\_ok} is $\yexe_t$; \texttt{decision} is $d_t$.
Figure~\ref{fig:ladder} shows a pragmatic step, $\hat y = (\textsc{na}, 1)$, $d = \textsc{trust}$, and an epistemic step, $\hat y = (\textsc{fail}, 1)$, $d = \textsc{abstain}$.

The epistemic step predicts \textsc{fail} for a reproduction script, since the bug
is expected to be present, and chooses to abstain. Within the block, the two predicted-value tokens receive $\lpred$ and the decision token the policy gradient.

\subsection{Outcome channels, parsing rules, and decision reward per harness}
\paragraph{Action typing and the verifier.}
The action-type map $\chi$ of \S\ref{def:chi-psi} marks as epistemic
the steps that run a check the verifier can evaluate, and as pragmatic all remaining steps. Its syntactic form is the EPISTEMIC/PRAGMATIC rule of the belief protocol in Appendix~\ref{app:prompts}.
The check field is instantiated per scaffold (Table~\ref{tab:parser-rules}). On the repository scaffold it is the outcome of the check the step runs; on pragmatic steps it is \textsc{na} by definition, which leaves the collector's forecast unsupervised there. On the terminal scaffold it is the outcome of the hidden task verifier (the \texttt{expect\_success} target), which every step carries. On the terminal
corpus the predicted-outcome key set is \{\texttt{exit\_ok},
\texttt{expect\_success}\}; steps carrying a predicted
\texttt{target\_test}, for which no grading source exists, are dropped
at corpus construction. The block schema is identical at collection, fine-tuning, reinforcement learning, and evaluation.

\begin{table}[h]
\centering
\caption{\textbf{Parser rules.} The check field and the exit field per scaffold.}
\label{tab:parser-rules}
\small
\setlength{\tabcolsep}{2pt}
\begin{tabular}{@{}llp{3.9cm}@{}}
\toprule
 & Repository issues (R2E-Gym) & Terminal tasks \\
\midrule
Check field predicts & Outcome of the check the step runs & Outcome of the hidden task verifier \\
Values & \textsc{pass} / \textsc{fail} / \textsc{na} & \textsc{pass} / \textsc{fail} \\
\textsc{na} is definitional on & Pragmatic steps & -- \\
Infeasible pair & (\textsc{pass}, exit field $0$): a passing test implies a clean run & -- \\
Exit field predicts & Command exits with code zero & Same \\
Conclusive iff & Check field in \{\textsc{pass}, \textsc{fail}\} & After the last step \\
Anchor key & Predicted check field & Predicted exit field \\
\bottomrule\end{tabular}
\end{table}

\paragraph{Scoring and shaping details.}
On conclusive steps, where $\rtst_{i,t} \in \{\textsc{pass}, \textsc{fail}\}$, the scoring rule $\sigma$ of
\S\ref{sec:method:objective} is
\begin{equation}
r_{i,t} \;=\; \sigma(d_{i,t}, \hat y_{i,t}, y_{i,t}) \;=\;
\begin{cases}
\;+\alpha & d_{i,t} = \textsc{trust},\ \hat y_{i,t} = y_{i,t},\ \rexe_{i,t} = 1,\\
\;\;\;0 & d_{i,t} = \textsc{trust},\ \hat y_{i,t} = y_{i,t},\ \rexe_{i,t} = 0,\\
\;-\gamma & d_{i,t} = \textsc{trust},\ \hat y_{i,t} \neq y_{i,t},\\
\;-\kappa & d_{i,t} = \textsc{abstain},
\end{cases}
\label{eq:settle}
\end{equation}
where $\alpha, \gamma, \kappa > 0$ and equality is vector equality over both fields. Every inconclusive epistemic step scores zero, and so does a fully correct trusted crash forecast, which is also excluded from the anchor groups. Two asymmetries hold on the repository scaffold. A trusted \textsc{na} on a step the verifier decides counts as a miss, so a predicted \textsc{na} never scores $+\alpha$. Agreement between a predicted and a realized \textsc{na} is rewarded only through the prediction loss of Eq.~\ref{eq:lpred}, whose value tokens receive no policy gradient, because \textsc{na} is the one outcome the policy can produce at will (an obfuscated test command forces it).

\FloatBarrier
\section{Training setup and infrastructure}\label{app:hparams}

\begin{algorithm}[!t]
\caption{ANTE pipeline, expanded from Algorithm~\ref{alg:bapo-compact}}
\label{alg:bapo}
\footnotesize
\begin{algorithmic}[1]
\Require pool $\mathcal{D}$; $N$; $\pi_{\mathrm{ref}}$; $\psi$, $\chi$; $(\alpha, \gamma, \kappa)$; $\omega$, $\lambda$, $\eta$, $\eta_{\mathrm{lr}}$
\Statex \textcolor{palo!55!gray}{\textit{// BeliefTraj curation}}
\State collect verified solving trajectories in the containers
\State $y_{i,t} \gets \psi(o_{i,t+1}, \chi(a_{i,t}))$, written back as
       $\hat y_{i,t}$;\;\; label $d_{i,t}$ by rule
\Statex \textcolor{palo!55!gray}{\textit{// Stage~I: supervised fine-tuning on BeliefTraj}}
\State $\theta \gets \arg\min_\theta$ next-token CE on BeliefTraj
\Statex \textcolor{palo!55!gray}{\textit{// Stage~II: reinforcement learning with BAPO}}
\For{each iteration}
  \State sample $x \sim \mathcal{D}$; roll out $\tau_1, \dots, \tau_N \sim \pi_\theta(\cdot \mid x)$: each step emits $\beta_{i,t}$,
         then $a_{i,t}$, executed
  \State $y_{i,t} \gets \psi(o_{i,t+1}, \chi(a_{i,t}))$
  \State $r_{i,t} \gets \sigma(d_{i,t}, \hat y_{i,t}, y_{i,t})$ on $\mathcal{C}_i$, else $0$
         \Comment{Eq.~\ref{eq:settle}}
  \State $A_i \gets R_i - \tfrac{1}{N}\sum_j R_j$;\;\;
         $W_{i,t} \gets \sum_{s \ge t} r_{i,s}$
  \State $\Vhat(\tilde y) \gets \mathrm{mean}_{G(\tilde y)} W$;\;\;
         $\Qhat(\tilde y, d) \gets \mathrm{mean}_{G(\tilde y, d)} W$
  \State $\Delta_{i,t} \gets \omega\big(\tilde A^{\tilde y}_{i,t} +
          \tilde A^{\tilde y, d}_{i,t}\big)$ on anchored steps, else $0$;\;\; $\Delta \equiv 0$ on a task with no trusted conclusive step
  \State $A_{i,t} \gets A_i + \Delta_{i,t}$ \Comment{Eq.~\ref{eq:bapo-adv}}
  \State $\theta \gets \theta - \eta_{\mathrm{lr}} \nabla_\theta
          \mathcal{L}(\theta)$ \Comment{Eq.~\ref{eq:objective}}
\EndFor
\end{algorithmic}
\end{algorithm}

\subsection{Supervised fine-tuning records}
\label{app:sft-records}

\begin{figure}[h]
\centering
\includegraphics[width=\textwidth]{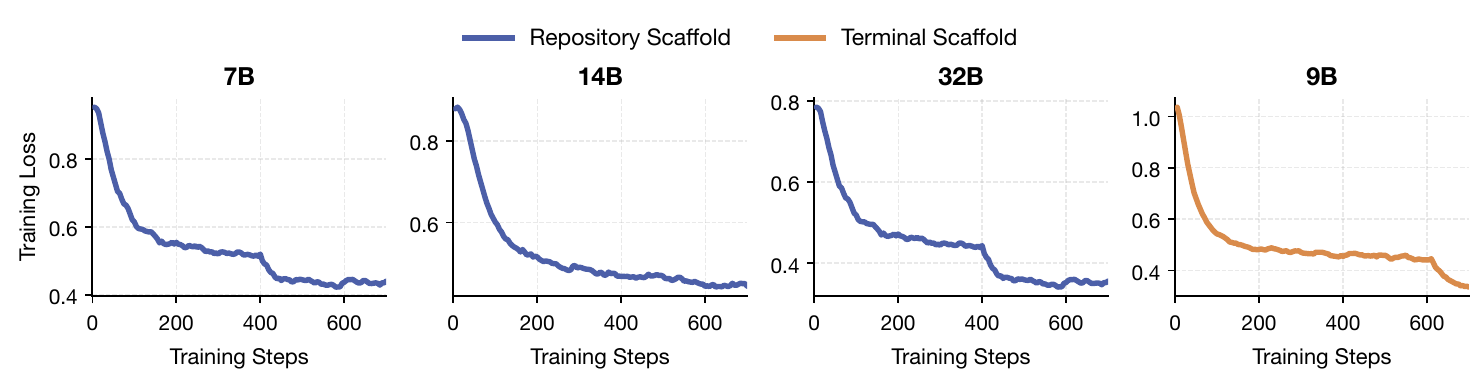}
\caption{\textbf{Supervised fine-tuning loss of ANTE.}}
\label{fig:sft9}
\end{figure}

Figure~\ref{fig:sft9} shows the supervised fine-tuning loss of the four ANTE runs: 7B, 14B, and 32B on the repository scaffold and 9B on the terminal scaffold.

\subsection{The optimization objective}
\begin{figure}[h]
\centering
\includegraphics[width=\textwidth]{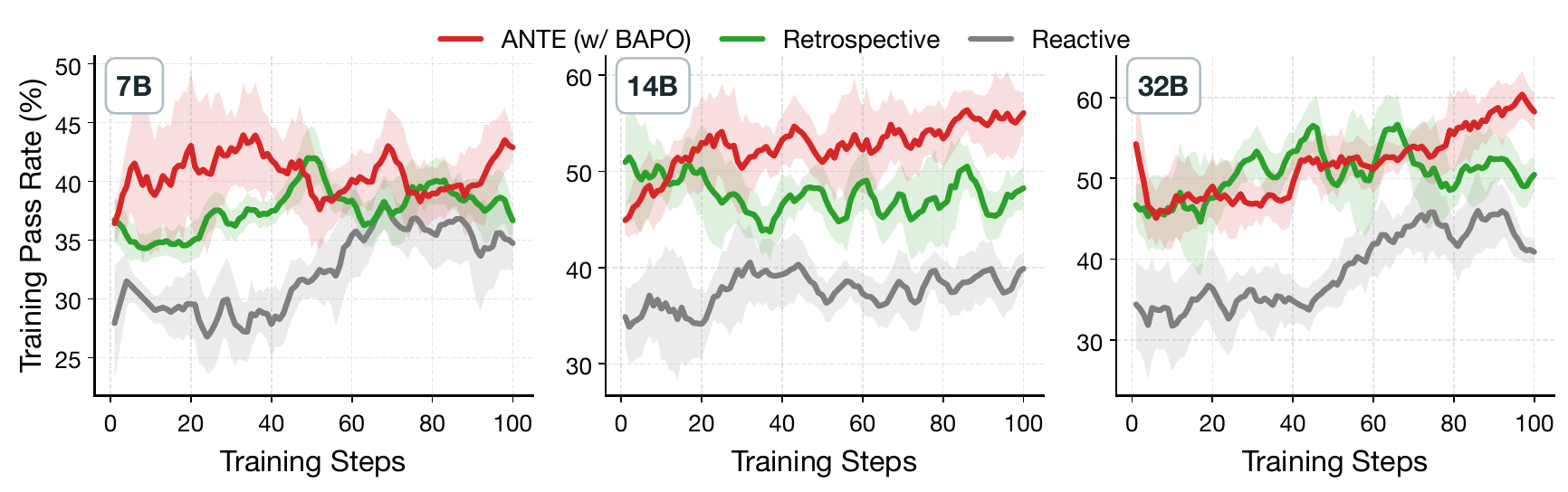}
\caption{\textbf{Training pass rate under RL.} Training pass rate (\%) for Reactive, Retrospective, and ANTE at 7B, 14B, and 32B on the RL task pool.}
\label{fig:dynamics}
\end{figure}
\paragraph{Group-relative objective.}
For a task $x$, the current policy samples $N$ rollouts $\tau_1, \dots, \tau_N$, each receives its task reward $R_i$, and GRPO \citep{shao2024deepseekmath} maximizes the clipped
surrogate
\begin{equation}
\mathcal{J}(\theta) = \mathbb{E}\!\left[\frac{1}{N}\sum_{i=1}^{N}
\frac{1}{|\tau_i|}\sum_{k=1}^{|\tau_i|}
\min\!\big(\varrho_{i,k}\hat A_i,\ \mathrm{clip}(\varrho_{i,k}, 1-\epsilon,
1+\epsilon)\hat A_i\big)\right] - \eta\, D_{\mathrm{KL}}\!\left(\pi_\theta
\,\|\, \pi_{\mathrm{ref}}\right),
\label{eq:grpo}
\end{equation}
where $\hat A_i = (R_i - \mathrm{mean}_j R_j)/\mathrm{std}_j R_j$ is the
group-normalized advantage and $\varrho_{i,k} = \pi_\theta / \pi_{\mathrm{old}}$
the importance ratio of the $k$-th generated token of $\tau_i$. Dr.GRPO
\citep{liu2025drgrpo} drops the per-sequence length normalization and the
standard-deviation scaling, and DAPO \citep{yu2025dapo} widens the upper
clip to $\epsilon_{\mathrm{hi}} > \epsilon_{\mathrm{lo}}$; ANTE adopts both, keeps the token-mean aggregation, and adds the step-level signals of \S\ref{sec:method:objective}: the prediction loss, the decision reward, and the anchored advantages. Every member of this family assigns one advantage per rollout, broadcast to every token of every step; that is the property BAPO changes.

Table~\ref{tab:hparams} lists the repository-task configuration of both
training stages; the remaining terminal-task values that differ from it (rollouts per task, step limit, response length, and batch geometry) are listed in Appendix~\ref{app:terminal-config}.

\begin{table}[h]
\centering
\caption{\textbf{Training configuration.} Settings are shared across model scales except where noted.}
\label{tab:hparams}
\footnotesize
\resizebox{\textwidth}{!}{\begin{tabular}{@{}lll@{}}
\toprule
 & Parameter & Value \\
\midrule
SFT & Objective & Next-token cross-entropy \\
 & Corpus & One variant of the paired corpora per training run \\
 & Epochs / cutoff & 3 / 32{,}768 tokens \\
 & Learning rate & $10^{-5}$, cosine, 5\% warmup \\
\midrule
RL & Rollouts per task & $N = 8$ (64 tasks per step) \\
 & Schedule & 103 steps \\
 & Sampling & temperature $1.0$, top-$p$ 0.8, top-$k$ 20 \\
 & Step limit / per-step cap & 50 steps / 4{,}096 tokens (32B: 8{,}192) \\
 & Context geometry & 4{,}096 prompt $+$ 34{,}816 response \\
 & Learning rate & $10^{-6}$ \\
 & Clipping $(\epsilon_{\mathrm{lo}}, \epsilon_{\mathrm{hi}})$ & $(0.2, 0.28)$ \\
 & KL coefficient $\eta$ & $0.01$; none on the terminal harness \\
 & Prediction weight $\lambda$ & $10$ against the token mean $Z$ \\
 & & $0.05$ against a per-position mean on the terminal harness \\
 & Reward coefficients $(\alpha, \gamma, \kappa)$ & $(0.05, 0.0625, 0.02)$ \\
 & Format term $r^{\mathrm{fmt}}_i$ & $0.02$ per step, clip $\pm 0.15$ (repository); rate, weight $1.0$ (terminal) \\
 & Advantage aggregation & Token mean; Episode advantage unnormalized \\
 & Group filtering & Uniform-outcome task groups dropped, not re-sampled \\
 & Step-stream normalization & Per-task std with floor $\alpha + \gamma$ \\
 & Step-stream weight $\omega$ & $0.5$ \\
 & Step-term clip & $\pm 2$ after standardization \\
\midrule
ANTE-TTO & Trigger & Abstention or predicted \textsc{fail} \\
 & Candidates ($K$) & 3 extra per trigger \\
 & Scoring & \textsc{pass} 3, \textsc{na} 1, \textsc{fail} 0, $+1$ clean exit \\
\bottomrule\end{tabular}}
\end{table}

\subsection{Break-even derivation of the reward coefficients}
\paragraph{Why the coefficients are derived quantities.}
Fixing the scale $\alpha > 0$, let $q$ denote the probability that a
trusted prediction is correct. On steps where a correct trusted forecast receives $+\alpha$, the expected score of trusting is
$q\alpha - (1 - q)\gamma$, which is positive iff $q > \gamma / (\alpha +
\gamma)$; choosing a target accuracy $q^\ast \in (0, 1)$ and setting $\gamma$ from it,
\begin{align}
\gamma \;&=\; \alpha\, \frac{q^\ast}{1 - q^\ast}
&&\text{(trusting is rewarded in expectation iff } q > q^\ast\text{)},
\label{eq:price-gamma}\\
q^\circ \;&=\; q^\ast - \frac{\kappa}{\alpha + \gamma}
&&\text{(indifference between trusting and abstaining, clean-exit forecasts)},
\label{eq:price-indiff}
\end{align}
places the trust-versus-abstain break-even point $q^\circ$ strictly below the target, since abstaining costs the flat $\kappa > 0$ regardless of the prediction. A policy whose predictions are becoming reliable can therefore drop the corpus's blanket abstention on epistemic steps before its accuracy reaches $q^\ast$, while trusting is rewarded in expectation only above $q^\ast$ itself.

With the coefficients of Table~\ref{tab:hparams}, the target accuracy is $q^{\ast}$ and the break-even point is $q^{\circ}$. The task reward adds to the binary resolve outcome the bounded train-time format term of \S\ref{sec:method:objective}. The term credits executed steps with a well-formed belief block and penalizes unsolved rollouts near the response budget.

\subsection{Compute and software stack}
\label{app:compute}

\paragraph{Training compute.}
We use NVIDIA A100 40GB GPUs for training: 64 GPUs at 7B, 14B, and 9B, and 128 GPUs at 32B.

\paragraph{Software stack.}
The repository-scaffold RL runs build on verl \citep{sheng2025hybridflow} and the terminal runs use the open-instruct GRPO trainer released with Tmax \citep{ivison2026tmax} on the Harbor. Rollout and evaluation inference use vLLM-based workers \citep{kwon2023vllm}. The repository scaffold is the official R2E-Gym tool interface (the base prompt of Appendix~\ref{app:prompts}).

\FloatBarrier
\section{ANTE-TTO: candidate score, algorithm, and examples}\label{app:tto}

\subsection{Scoring rule and inference algorithm}
\paragraph{Candidate score.}
ANTE-TTO scores a candidate by the belief it carries. For candidate $k$ with predicted check field $\ytst{}^{(k)}$ and exit field $\yexe{}^{(k)}$, the score is
\begin{equation}
s\big(\beta^{(k)}\big) \;=\; \phi\big(\ytst{}^{(k)}\big) + \mathbf{1}\big[\yexe{}^{(k)} = 1\big],
\qquad
\phi(\textsc{pass}) = 3,\quad \phi(\textsc{na}) = 1,\quad \phi(\textsc{fail}) = 0,
\label{eq:tto-score}
\end{equation}
and the selected candidate is $\tilde k = \argmax_{k \in \{0, 1, \dots, K\}} s(\beta^{(k)})$, with ties resolved toward $k = 0$, the action the policy
generated first. The ordering $\textsc{pass} \succ \textsc{na} \succ
\textsc{fail}$ ranks a predicted pass above an indeterminate
outcome above a predicted failure; the exit term favors candidates the
policy expects to execute cleanly. The rule consumes only the two value
tokens of the belief, so it adds no model and no execution. Exactly one action is executed per step.

\begin{algorithm}[h]
\caption{ANTE-TTO inference}
\label{alg:tto}
\footnotesize
\begin{algorithmic}[1]
\Require policy $\pi_\theta$; trigger set $\mathcal{E}$ (abstention or predicted \textsc{fail}); candidates per trigger
         $K$
\Ensure trajectory $\tau$
\For{$t = 1, 2, \dots$ until submission}
  \Statex \textcolor{palo!55!gray}{\textit{// Phase 1: generate and parse the belief}}
  \State $u_t = (v_t, \beta_t, a_t) \gets$ greedy decoding of $\pi_\theta(\cdot \mid h_t)$;\;\; $u^{(0)} \gets u_t$
  \Statex \textcolor{palo!55!gray}{\textit{// Phase 2: resample only where the belief signals uncertainty}}
  \If{$\beta_t \in \mathcal{E}$}
    \State sample $u^{(k)} = (v^{(k)}, \beta^{(k)}, a^{(k)}) \sim \pi_\theta(\cdot \mid h_t)$ for $k = 1, \dots, K$
    \State $\tilde k \gets \argmax_{k \in \{0, \dots, K\}} s(\beta^{(k)})$, ties toward $k = 0$ \Comment{Eq.~\ref{eq:tto-score}}
    \State $u_t \gets u^{(\tilde k)}$;\;\; $(v_t, \beta_t, a_t) \gets u_t$
  \EndIf
  \Statex \textcolor{palo!55!gray}{\textit{// Phase 3: execute the selected action}}
  \State $o_{t+1} \gets \mathrm{execute}(a_t)$;\;\; $h_{t+1} \gets (h_t, u_t, o_{t+1})$
\EndFor
\State \Return $\tau$
\end{algorithmic}
\end{algorithm}

The trigger set and $K$ take the values of Table~\ref{tab:hparams}. At a trigger step the draft stays greedy, and the three alternatives are sampled at temperature 1.0 under the same token cap and context, so an ANTE-TTO trajectory mixes greedy and sampled steps. Token budgets count prompt and generated tokens, discarded candidates included, and are measured from the recorded call counts of the same evaluation runs. Figure~\ref{fig:mtc} reports the marginal token cost of each setting and the averaged $B_{\mathrm{tok}}$ and $\mathrm{nMTC}$.

\subsection{Examples on different harnesses}
\begin{figure}[!t]
\centering
\includegraphics[width=\textwidth]{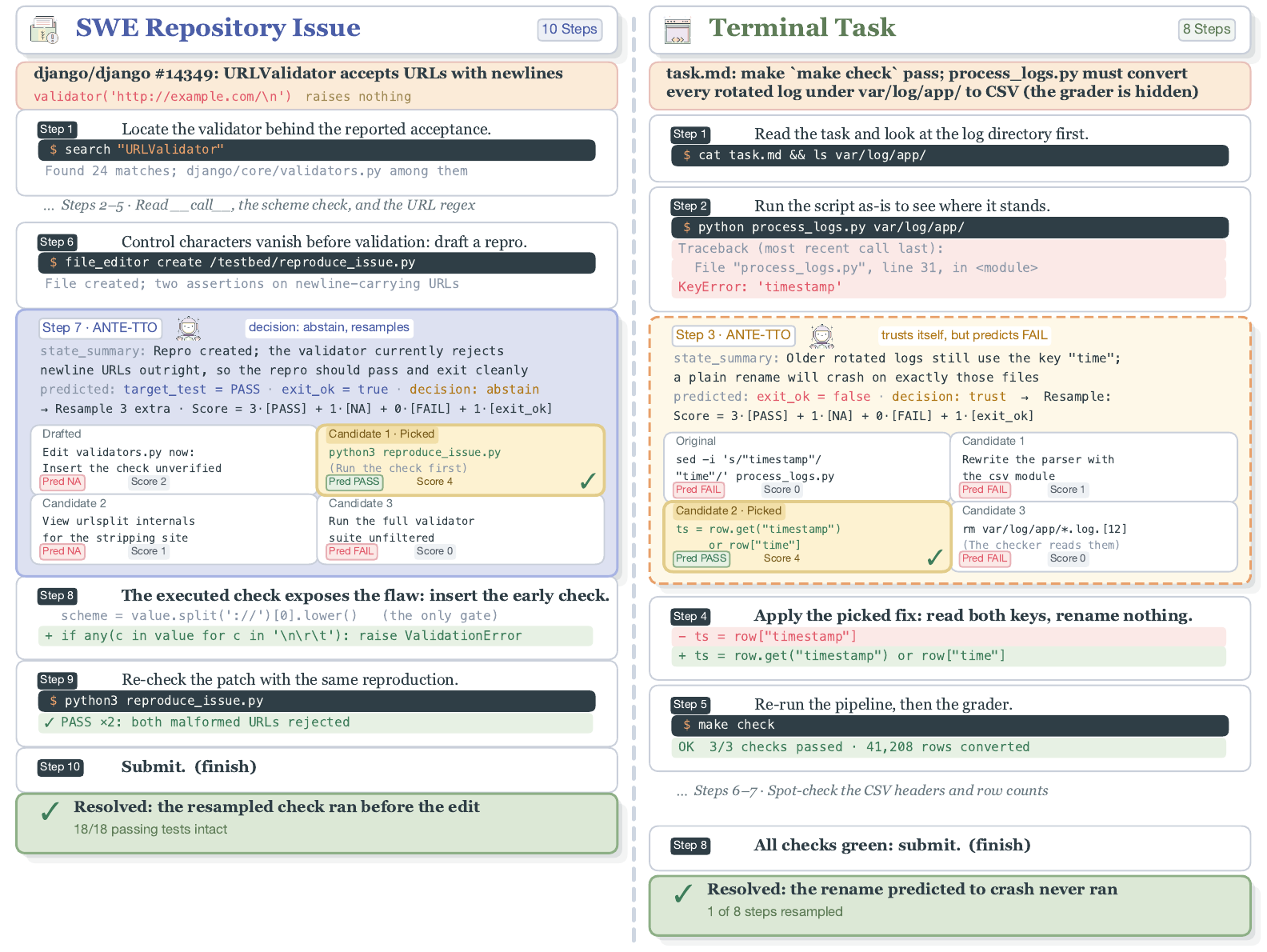}
\caption{\textbf{ANTE-TTO across the two harnesses.} At a trigger step
the policy resamples candidate actions and scores each by its own
belief; the command predicted to crash never touches the environment.}
\label{fig:case}
\end{figure}
On the repository scaffold (32B policy,
\texttt{django\_\_django-14349}), the step-7 abstention triggers
resampling: the four candidates score $\{4, 2, 1, 0\}$ and the winner replaces the drafted command (full trace in Appendix~\ref{app:case-studies}). On the terminal scaffold (Figure~\ref{fig:case}, right), the belief predicts a crash for the plain-rename fix, and the predicted failure triggers resampling. The read-both-keys fix wins with a score of 4, and the log-deleting shortcut scores zero.

\FloatBarrier
\section{A worked example of BAPO}
\label{app:worked}

We compute the BAPO assignment (\S\ref{sec:method:bapo}) on the
configuration of
Figure~\ref{fig:bapo-overview}: one task, three rollouts, and reward coefficients $(\alpha, \gamma, \kappa) = (0.05, 0.0625, 0.02)$. We report the belief-relative and decision-relative advantages, which the assignment combines at weight
$\omega = \tfrac{1}{2}$ after standardization. Conclusive steps are written as
[prediction, decision] with the decision reward to the right; F and P stand for the predicted \textsc{fail} and \textsc{pass}, a check mark means the prediction was correct, and a cross that it missed.

\begin{center}
\small
\begin{tabular}{@{}llll@{}}
\toprule
 & Step $t_a$ & Step $t_b$ & Step $t_c$ \\
\midrule
$\tau_1$ ($R{=}1$) & [F, abstain] $\;-0.02$ & [F, trust\,\ding{51}] $\;+0.05$ & [P, abstain] $\;-0.02$ \\
$\tau_2$ ($R{=}0$) & [F, abstain] $\;-0.02$ & [P, trust\,\ding{55}] $\;-0.0625$ & [P, abstain] $\;-0.02$ \\
$\tau_3$ ($R{=}0$) & [F, trust\,\ding{51}] $\;+0.05$ & [P, trust\,\ding{55}] $\;-0.0625$ & \\
\bottomrule
\end{tabular}
\end{center}

\begin{figure}[h]
\centering
\includegraphics[width=0.95\textwidth]{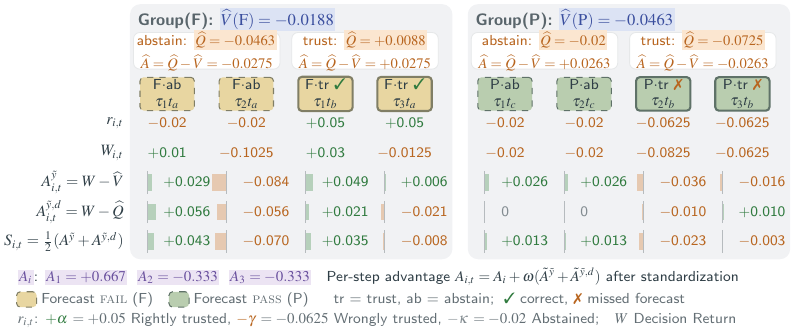}
\caption{\textbf{BAPO worked example.}}
\label{fig:app-worked}
\end{figure}

\paragraph{Episode advantages.} $R = [1, 0, 0]$, mean $1/3$, so
$A = [+0.667, -0.333, -0.333]$; every pragmatic step receives its trajectory's
$A_i$ unchanged.

\paragraph{Decision return.} $W$ is the suffix sum of the decision rewards within
each trajectory: for $\tau_1$, $W(t_c) = -0.02$, $W(t_b) = +0.03$,
$W(t_a) = +0.01$; for $\tau_2$, $W(t_c) = -0.02$, $W(t_b) = -0.0825$,
$W(t_a) = -0.1025$; for $\tau_3$, $W(t_b) = -0.0625$, $W(t_a) = -0.0125$.

\paragraph{Groups.} Belief groups ignore step positions:
$G(\mathrm{F}) = \{\tau_1 t_a, \tau_1 t_b, \tau_2 t_a, \tau_3 t_a\}$ and
$G(\mathrm{P}) = \{\tau_1 t_c, \tau_2 t_b, \tau_2 t_c, \tau_3 t_b\}$. Rollout $\tau_1$ contributes two members to $G(\mathrm{F})$, and $G(\mathrm{P})$ mixes steps from different positions. The group means are
$\Vhat(\mathrm{F}) = -0.0188$ and $\Vhat(\mathrm{P}) = -0.0463$.

\paragraph{Belief-relative advantages.} $A^{\tilde y} = W - \Vhat$ rewards steps whose decision returns exceed their group mean. The trusted correct prediction $\tau_1 t_b$ gets the largest belief-relative advantage ($+0.049$); the opening step of the rollout with the lowest decision return ($\tau_2 t_a$) gets the most negative ($-0.084$). In this example the signs of the final advantages are still set by $A_i$: a correct trust inside a failed trajectory ($\tau_3 t_a$) remains negative overall.

\paragraph{Decision advantage.} Splitting $G(\mathrm{P})$ by decision gives
$\Qhat(\mathrm{P}, \textsc{abstain}) = -0.02$ and
$\Qhat(\mathrm{P}, \textsc{trust}) = -0.0725$. The shared decision advantage $\Ahat(\mathrm{P}, d) = \Qhat - \Vhat$ is therefore $+0.0263$ for abstain and $-0.0263$
for trust: under a predicted \textsc{pass}, trusting carried negative expected reward in this batch. On
$G(\mathrm{F})$ the sign flips, $\Ahat(\mathrm{F}, \textsc{trust}) = +0.0275$, because in this batch the \textsc{fail} forecasts were the ones that proved correct, so trusting them was rewarded. The
decision is scored per belief, which is why the group key contains
$\tilde y$. Before standardization, the step term $\tfrac{1}{2}(A^{\tilde y} + A^{\tilde y, d})$ of the BAPO assignment is largest for the abstained \textsc{fail} forecast that opens the winning rollout ($\tau_1 t_a$, $+0.043$) and most negative for the same opening step of $\tau_2$ ($\tau_2 t_a$, $-0.07$). The correct trust inside the failed rollout ($\tau_3 t_a$) keeps a positive belief-relative advantage ($A^{\tilde y} = +0.006$) even as
its total advantage stays negative.
The decomposition $A^{\tilde y} = \Ahat + A^{\tilde y, d}$ (Eq.~\ref{eq:decomp}) holds for every member listed above.

\FloatBarrier
\section{Terminal instantiation and configuration differences}
\label{app:terminal-config}

We redefine the outcome space, the anchor key, and the decision reward since the outcomes the terminal verifier can read are different. In the Harbor task format the hidden test runs after the episode, so the harness returns two signals. The exit status of each command is forecast by the exit field and settled by the next observation. The episode outcome is forecast by the check field and settled once by the verifier. Belief groups are keyed on the predicted exit field, the coordinate that recurs across rollouts, and the decision reward takes the episode outcome. Eq.~\ref{eq:lpred} is instantiated with the exit field as the supervised target at every step and a per-position denominator (Table~\ref{tab:hparams}).

\begin{figure}[!t]
\centering
\includegraphics[width=0.53\textwidth]{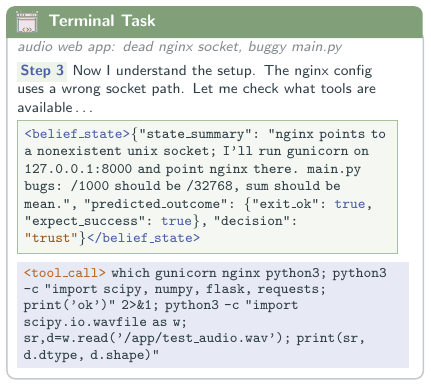}
\caption{\textbf{Belief state on the terminal harness.}}
\label{fig:belief-state}
\end{figure}

\begin{figure}[!t]
\centering
\includegraphics[width=\textwidth]{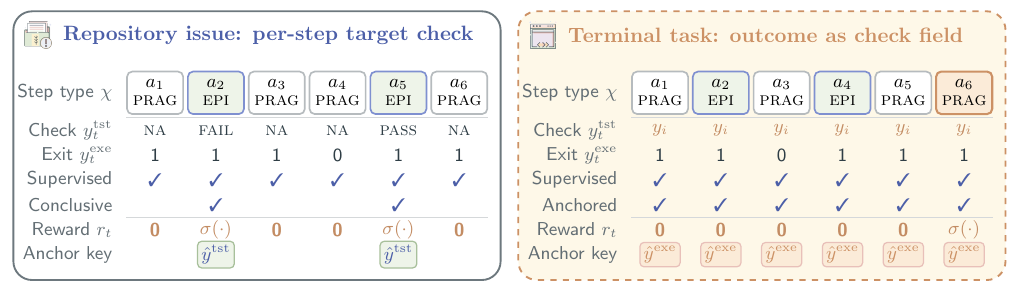}
\caption{\textbf{Terminal instantiation.} Left: repository issues. Right: terminal tasks.}
\label{fig:app-terminal}
\end{figure}

\paragraph{Instantiation.}
The check field predicts the hidden verifier's outcome, and the exit field predicts whether each command exits with code zero (Figure~\ref{fig:belief-state}). The exit field is supervised on every step against the next observation's exit code. The check field carries the episode outcome and enters credit through the decision reward rather than the prediction loss. The check coordinate of the realized outcome is constant within an episode, while the exit coordinate is read per step from the next
observation. Credit is per-step as well: steps are anchored on their
predicted exit field, and belief-decision groups split each belief group exactly as on repository issues. The hidden verifier evaluates the episode once, so the check field is scored after the last step (Table~\ref{tab:parser-rules}): $r_{i,t} = 0$ for $t < T_i$, $r_{i,T_i} = \sigma(d_{i,T_i}, \hat y_{i,T_i}, y_{i,T_i})$, and $W_{i,t} = r_{i,T_i}$. Here $\mathcal{C}_i := \{T_i\}$, and the belief groups are $G(\tilde y) := \{(j, s) : m_{j,s} = 1,\ \yexe_{j,s} = \tilde y\}$ for $\tilde y \in \{0, 1\}$. Steps of one rollout therefore share a return but not a baseline. The group of a step is set by its own predicted exit field, so a rollout's steps are measured against different $\Vhat$ and $\Qhat$ according to what each predicted. A step whose block does not parse is not anchored (Figure~\ref{fig:app-terminal}). At inference, the ANTE-TTO trigger of \S\ref{sec:method:validator} reads the predicted exit field in place of the check field.

\paragraph{Terminal training configuration.}
The terminal variants train on the Tmax agent on Harbor as follows. The RL pool holds 3{,}558 tasks of the 14{,}601-task tmax-15k pool \citep{ivison2026tmax}. Each variant starts from its own fine-tuned checkpoint and uses its own prediction, format, and anchoring coefficients. Each episode has a 48K-token generation budget and a 120-step cap, and each training step uses 16 tasks with $N = 16$ rollouts each, 256 rollouts in all.

\FloatBarrier
\section{Evaluation against concurrent methods on SWE-bench Verified}
\label{app:concurrent}

Table~\ref{tab:sota-full} places ANTE among open-weight methods on SWE-bench Verified at all three scales. At the two smaller scales, ANTE-7B reaches 27.2 against 24.1 for SWE-AGILE-8B \citep{kde2026sweagile}, the best task-trained agent below 10B, and ANTE-14B reaches 46.2 against 30.1 for SWE-AGILE-14B, the best between 10B and 30B. Every entry is a pass@1 value, and external rows are as reported under each method's own scaffold. The ANTE evaluation runs with a 64K context and a 200-step cap.

\begin{table}[!htbp]
\centering
\caption{Open-weight methods on SWE-bench Verified (pass@1, \%) at all three scales. \textbf{Bold} marks the best and \underline{underline} the second-best task-trained method per scale.}
\label{tab:sota-full}
\fontsize{8pt}{9.6pt}\selectfont
\setlength{\tabcolsep}{5pt}
\renewcommand{\arraystretch}{1.08}
\begin{tabular}{@{}llllc@{}}
\toprule
Method & Backbone & Training & Scaffold & Resolved (\%) \\
\midrule
\multicolumn{5}{@{}l}{\itshape $<$10B} \\
Qwen2.5-Coder-7B & -- & -- & OpenHands & 1.8 \\
Qwen3-8B & -- & -- & ProRL Agent & 9.6 \\
SWE-Dev-7B & Qwen2.5-Coder-7B & SFT & OpenHands & 23.4 \\
R2E-Gym-7B & Qwen2.5-Coder-7B & SFT & R2E-Gym & 19.0 \\
Lingma-SWE-GPT-7B & Qwen2.5-Coder-7B & SFT & SWESynInfer & 18.2 \\
SWE-agent-LM-7B & Qwen2.5-Coder-7B & SFT & SWE-agent & 15.2 \\
SkyRL-Agent-7B-v0 & Qwen2.5-Coder-7B & SFT+RL & OpenHands & 14.6 \\
SWE-Gym-7B & Qwen2.5-Coder-7B & SFT & OpenHands & 10.6 \\
SWE-AGILE-8B & Qwen3-8B & SFT+RL & R2E-Gym & \underline{24.1} \\
ProRL Agent-8B & Qwen3-8B & RL & ProRL Agent & 18.0 \\
\rowcolor{beliefbg}
ANTE-7B & Qwen2.5-Coder-7B & SFT+RL & R2E-Gym & \textbf{27.2} \\
\midrule
\multicolumn{5}{@{}l}{\itshape 10--30B} \\
Qwen3-14B & -- & -- & ProRL Agent & 15.4 \\
Qwen2.5-Coder-14B & -- & -- & OpenHands & 4.0 \\
SWE-AGILE-14B & Qwen3-14B & SFT & R2E-Gym & \underline{30.1} \\
R2E-Gym-14B & Qwen2.5-Coder-14B & SFT & R2E-Gym & 26.8 \\
ProRL Agent-14B & Qwen3-14B & RL & ProRL Agent & 23.6 \\
SkyRL-Agent-14B-v0 & Qwen3-14B & RL & OpenHands & 21.6 \\
SWE-Gym-14B & Qwen2.5-Coder-14B & SFT & OpenHands & 16.4 \\
\rowcolor{beliefbg}
ANTE-14B & Qwen2.5-Coder-14B & SFT+RL & R2E-Gym & \textbf{46.2} \\
\midrule
\multicolumn{5}{@{}l}{\itshape $>$30B} \\
Qwen3-32B & -- & -- & R2E-Gym & 25.0 \\
Qwen3-32B & -- & -- & ReAct & 24.4 \\
Qwen3-32B & -- & -- & OpenHands & 23.2 \\
Qwen2.5-Coder-32B & -- & -- & OpenHands & 7.0 \\
SWE-World-32B & Qwen2.5-Coder-32B & SFT+RL & R2E-Gym & \underline{55.0} \\
FrogBoss-32B & Qwen3-32B & SFT & R2E-Gym & 54.6 \\
Meta-CWM-32B & Scratch & SFT+RL & CWM & 53.9 \\
SWE-Lego-32B & Qwen3-32B & SFT & OpenHands & 52.6 \\
DeepSWE-32B & Qwen3-32B & RL & R2E-Gym & 42.2 \\
SWE-agent-LM-32B & Qwen2.5-Coder-32B & SFT & SWE-agent & 40.2 \\
Skywork-SWE-32B & Qwen2.5-Coder-32B & SFT & OpenHands & 38.0 \\
OpenHands-LM-32B & Qwen2.5-Coder-32B & SFT & OpenHands & 37.2 \\
SWE-Dev-32B & Qwen2.5-Coder-32B & SFT & OpenHands & 36.6 \\
R2E-Gym-32B & Qwen2.5-Coder-32B & SFT & R2E-Gym & 34.4 \\
SWE-Gym-32B & Qwen2.5-Coder-32B & SFT & OpenHands & 20.6 \\
Kimi-Dev-72B & Qwen2.5-72B & SFT+RL & SWE-agent & 48.6 \\
SWE-Fixer-72B & Qwen2.5-72B & SFT & Retrieve-and-edit & 32.8 \\
Lingma-SWE-GPT-72B & Qwen2.5-72B & SFT & SWESynInfer & 30.2 \\
\rowcolor{beliefbg}
ANTE-32B & Qwen3-32B & SFT+RL & R2E-Gym & \textbf{57.2} \\
\bottomrule
\end{tabular}
\end{table}

Sources for Table~\ref{tab:sota-full}: SWE-AGILE \citep{kde2026sweagile},
SWE-Dev \citep{wang2025swedev}, R2E-Gym
\citep{jain2025r2egym}, Lingma SWE-GPT \citep{ma2024lingma},
SWE-agent-LM \citep{yang2025swesmith}, SWE-Gym \citep{pan2024swegym},
SkyRL-Agent \citep{novasky2025skyrl}, SWE-World \citep{sun2026sweworld},
FrogBoss \citep{sonwane2025bugpilot}, Meta-CWM \citep{meta2025cwm}, Kimi-Dev
\citep{kimi2025kimidev}, DeepSWE \citep{luo2025deepswe}, Skywork-SWE
\citep{zeng2025skywork}, OpenHands LM \citep{openhands2025lm},
SWE-Fixer \citep{xie2025swefixer}, ProRL Agent
\citep{zhang2026prorl}, and SWE-Lego \citep{wang2026swelego}. Scaffolds: SWE-agent \citep{yang2024sweagent}, OpenHands \citep{wang2024openhands}, and ReAct \citep{yao2022react}. Terminal rows (Table~\ref{tab:term-market}): Nemotron-Terminal \citep{pi2026data}, TerminalTraj
\citep{wu2026termitraj}, ECHO \citep{shrivastava2026echo}; the Qwen3.5-9B base row with the Tmax agent on Harbor is our own measurement. The terminal base-model rows are from ECHO \citep{shrivastava2026echo} (Qwen3-8B and Qwen3-14B, five attempts at temperature 0.6) and TerminalTraj \citep{wu2026termitraj} (the Qwen2.5-Coder-7B, 14B, and 32B readings, four runs); Nemotron-Terminal \citep{pi2026data} reports 2.5 and 4.0 for the two Qwen3 bases without a trial count, and the OpenThinker-Agent model card \citep{openthinker2026agent} reports 0.0 for Qwen3-8B. The base-model scores on SWE-bench Verified are from \citet{pan2024swegym} (the
zero-shot Qwen2.5-Coder-7B, 14B, and 32B readings), \citet{zhang2026prorl}
(Qwen3-8B and 14B), and, for the three Qwen3-32B rows,
\citet{sonwane2025bugpilot}, \citet{skyrl2025agent}, and
\citet{wang2026swelego}. Table~\ref{tab:sota-full} is restricted to methods
that report publicly released weights.

\paragraph{Variation across runs.}
Three-run standard errors are the $\pm$ entries of Tables~\ref{tab:main} and~\ref{tab:term-results}.

\FloatBarrier
\section{Terminal evaluation protocol}\label{app:term}

This section introduces the protocol behind Tables~\ref{tab:term-results} and~\ref{tab:term-market}. External rows are as reported by each method under its own protocol
and scaffold, and the run counts they state differ: TerminalTraj averages four runs \citep{wu2026termitraj}, Endless Terminals averages five \citep{gandhi2026endless}, the published ECHO rows report pass@1 over five sampled attempts with Terminus-2 on Qwen3 backbones \citep{shrivastava2026echo}, and the remaining rows state none. Our ECHO reproduction implements ECHO's objective on the SkyRL trainer \citep{skyrl2025agent} with Qwen3.5-9B and our RL task pool, with 16 prompts and 16 samples per step, and is evaluated exactly like our own rows.

Table~\ref{tab:term-market}
places ANTE-9B among published open-weight terminal agents:
it resolves 24.7\% of Terminal-Bench 2.0, ahead of every task-trained
agent below 20B in the table, including ECHO at 8B and 14B and our
ECHO reproduction on the same backbone (21.4).

\begin{table}[!htbp]
\centering
\caption{\textbf{Open-weight terminal agents on Terminal-Bench 2.0
(Resolved, \%).} \textbf{Bold} marks the best and \underline{underline} the second-best task-trained method within each size band. $\dagger$: our reproduction of ECHO (backbone Qwen3.5-9B, trained on our RL task pool).}
\label{tab:term-market}
\fontsize{8pt}{9.6pt}\selectfont
\setlength{\tabcolsep}{4pt}
\renewcommand{\arraystretch}{1.08}
\begin{tabular}{@{}llllc@{}}
\toprule
Method & Backbone & Training & Scaffold & Resolved (\%) \\
\midrule
\multicolumn{5}{@{}l}{\itshape Base models} \\
Qwen2.5-Coder-7B & -- & -- & Terminus-2 & 0.0 \\
Qwen2.5-Coder-14B & -- & -- & Terminus-2 & 1.2 \\
Qwen2.5-Coder-32B & -- & -- & Terminus-2 & 4.5 \\
Qwen3-8B & -- & -- & Terminus-2 & 1.6 \\
Qwen3.5-9B & -- & -- & Tmax & 16.9 \\
Qwen3-14B & -- & -- & Terminus-2 & 4.3 \\
\midrule
\multicolumn{5}{@{}l}{\itshape $<$10B} \\
TerminalTraj-7B & Qwen2.5-Coder-7B & SFT & Terminus-2 & 10.1 \\
Nemotron-Terminal-8B & Qwen3-8B & SFT & Terminus-2 & 13.0 \\
EndlessTerminals-8B \citep{gandhi2026endless} & Qwen3-8B & SFT+RL & Minimal loop & 6.7 \\
ECHO-8B & Qwen3-8B & RL & Terminus-2 & 5.2 \\
OpenThinker-Agent-8B \citep{openthinker2026agent} & Qwen3-8B & SFT+RL & Terminus-2 & 4.9 \\
\rowcolor{beliefbg}
ANTE-9B & Qwen3.5-9B & SFT+RL & Tmax & \textbf{24.7} \\
ECHO-9B$^{\dagger}$ & Qwen3.5-9B & RL & Tmax & \underline{21.4} \\
\midrule
\multicolumn{5}{@{}l}{\itshape 10--30B} \\
Nemotron-Terminal-14B & Qwen3-14B & SFT & Terminus-2 & \textbf{20.2} \\
TerminalTraj-14B & Qwen2.5-Coder-14B & SFT & Terminus-2 & \underline{19.1} \\
ECHO-14B & Qwen3-14B & RL & Terminus-2 & 10.8 \\

\bottomrule
\end{tabular}
\end{table}

\FloatBarrier
\section{Additional analyses of the belief under RL}
\label{app:analyses}

We also analyze whether the belief protocol holds under RL and how well the emitted forecast judges the trajectory. We find that the protocol holds, and that the forecast judges the trajectory about as well as a dedicated reward model given the gold patch as a reference.

\begin{figure}[h]
\centering
\includegraphics[width=0.82\textwidth]{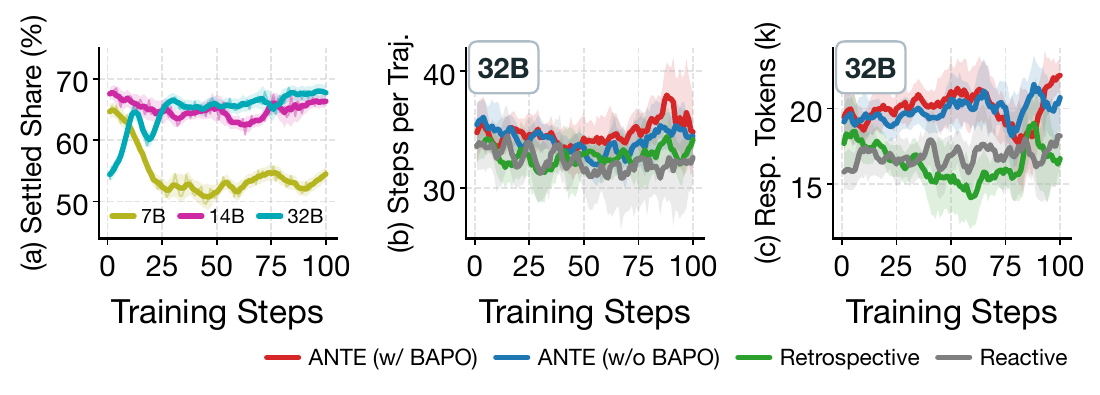}
\caption{\textbf{Belief protocol stability.} (a) The settled share of predictions, the steps whose forecast is scored, for ANTE at 7B, 14B, and 32B; (b) steps and (c) response tokens per trajectory at 32B for the four variants.}
\label{fig:protocol}
\end{figure}

\paragraph{Belief protocol stability.}
At 32B under RL (Figure~\ref{fig:protocol}), the settled share of predictions holds, and so do the steps and response tokens per trajectory.

\paragraph{Outcome judging against a reward model.}
A separately trained reward model is the standard tool for judging whether a trajectory resolves its task. We apply SWE-World's 32B reward model (SW-RM) \citep{sun2026sweworld}, with and without the gold patch, to ANTE-32B's trajectories and compare its verdicts with the outcome the belief forecasts (Table~\ref{tab:swr}). ANTE outperforms SW-RM and performs comparably even to SW-RM with the gold patch as reference.

\begin{table}[!htbp]
\centering
\caption{\textbf{Trajectory-outcome judging.} SW-RM: SWE-World's 32B reward model, with or without the gold patch. Covered Acc.\ and MCC (Matthews correlation coefficient) count only judged trajectories, and Overall Acc.\ also counts the rest.}
\label{tab:swr}
\small
\setlength{\tabcolsep}{5pt}
\begin{tabular}{@{}lrrr@{}}
\toprule
Judge & Covered Acc.\ (\%) & MCC & Overall Acc.\ (\%) \\
\midrule
SW-RM (w/o gold) & 65.5 & $+$0.359 & 58.4 \\
SW-RM (w.\ gold) & 72.3 & $+$0.454 & 63.2 \\
\rowcolor{beliefbg}
ANTE & \textbf{73.2} & $\bm{+}$\textbf{0.482} & \textbf{73.2} \\
\bottomrule
\end{tabular}
\end{table}

\FloatBarrier
\section{Failure modes without foresight}
\label{app:failures}

\begin{figure}[h]
\centering
\includegraphics[width=\textwidth]{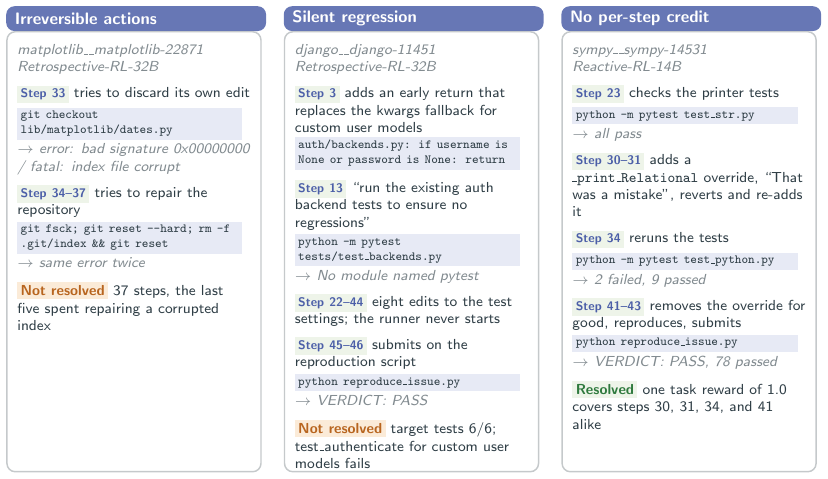}
\caption{\textbf{Failure modes without foresight.}}
\label{fig:app-failures}
\end{figure}

This section provides one episode for each failure mode of an agent that acts without a scored forecast (Figure~\ref{fig:app-failures}).

\paragraph{Irreversible actions.}
An agent that tries to discard its own edit with \texttt{git checkout} finds the repository's index corrupt and spends its last five steps repairing it (\texttt{matplotlib\_\_matplotlib-22871}). The exit field puts a forecast on every command's outcome, and a predicted clean run that crashes is a supervised miss (Eq.~\ref{eq:lpred}).

\paragraph{Silent regressions.}
On \texttt{django\_\_django-11451} the patch passes every target test but breaks authentication for custom user models. The agent's attempts to run the authentication tests fail, first on a missing module and then on the test settings, so it submits on its reproduction script, and only the final evaluation surfaces the damage. Fourteen further episodes of the same run end the same way. A per-step forecast, checked at every executed test, can surface such a regression at the next check.

\paragraph{No per-step credit.}
On \texttt{sympy\_\_sympy-14531} one resolved trajectory adds a printer override at step 30, breaks two tests with it, and removes it at step 41. The single task reward covers the step that broke the build and the step that fixed it alike.

\FloatBarrier
\section{Case studies}\label{app:case-studies}

We provide case studies from evaluation, training, and off-scaffold runs, each illustrating a different finding.

\subsection{Foresight guides test-time optimization}
On \texttt{django\_\_django-14349} the policy abstains at steps 7 and 9. At step 7, ANTE-TTO samples three more candidate actions, scores all four by their beliefs, and executes the highest-scoring one instead of the original action. The executed check refutes its abstained forecast and exposes the issue. The run finishes with the task resolved and every previously passing test intact.

\begin{trajstep}{paloalto}{Steps 1--3 ~\textperiodcentered~ locating the validator}
\begin{trajbelief}
<belief\_state>\{"state\_summary": "The URL validation bug likely lives
in the URLValidator class under django/core/validators.py.",
"predicted\_outcome": \{"target\_test": "NA", "exit\_ok": true\},
"decision": "trust"\}\allowbreak</belief\_state>
\end{trajbelief}
\texttt{\footnotesize <function=search> URLValidator}
\begin{trajobs}
Found 24 matches; ./django/core/validators.py among them
\end{trajobs}
\begin{trajbelief}
<belief\_state>\{"state\_summary": "core/validators.py holds the URL
validation logic; its condensed structure locates the class.",
"predicted\_outcome": \{"target\_test": "NA", "exit\_ok": true\},
"decision": "trust"\}\allowbreak</belief\_state>
\end{trajbelief}
\texttt{\footnotesize <function=file\_editor> view django/core/validators.py}
\begin{trajobs}
Condensed view of the file structure
\end{trajobs}
\begin{trajbelief}
<belief\_state>\{"state\_summary": "URLValidator.\_\_call\_\_ likely
mishandles a scheme or host case; locating its definition.",
"predicted\_outcome": \{"target\_test": "NA", "exit\_ok": true\},
"decision": "trust"\}\allowbreak</belief\_state>
\end{trajbelief}
\texttt{\footnotesize <function=search> def \_\_call\_\_ (validators.py)}
\begin{trajobs}
7 matches; the URLValidator's at line 101
\end{trajobs}
\end{trajstep}

\begin{trajstep}{paloalto}{Steps 4--6 ~\textperiodcentered~ reading the validator, writing a script that triggers the bug}
\begin{trajbelief}
<belief\_state>\{"state\_summary": "\_\_call\_\_ validates the scheme and
then the full URL; inspecting the body around line 101.",
"predicted\_outcome": \{"target\_test": "NA", "exit\_ok": true\},
"decision": "trust"\}\allowbreak</belief\_state>
\end{trajbelief}
\texttt{\footnotesize <function=file\_editor> view validators.py:101--124}
\begin{trajobs}
Scheme check at 105; full-URL validation below
\end{trajobs}
\begin{trajbelief}
<belief\_state>\{"state\_summary": "The URL regex at lines 86--92 defines
the accepted structure; checking it against the reported case.",
"predicted\_outcome": \{"target\_test": "NA", "exit\_ok": true\},
"decision": "trust"\}\allowbreak</belief\_state>
\end{trajbelief}
\texttt{\footnotesize <function=file\_editor> view validators.py:86--92}
\begin{trajobs}
The compiled URL regex, scheme validated separately
\end{trajobs}
\begin{trajbelief}
<belief\_state>\{"state\_summary": "Newer Python strips control characters
before URL splitting, so URLs containing newlines pass validation;
writing a reproduction script for both reported cases.",
"predicted\_outcome": \{"target\_test": "NA", "exit\_ok": true\},
"decision": "trust"\}\allowbreak</belief\_state>
\end{trajbelief}
\texttt{\footnotesize <function=file\_editor> create /testbed/reproduce\_issue.py}
\begin{trajobs}
File created; two assertions on newline-carrying URLs
\end{trajobs}
\end{trajstep}

\begin{trajstep}{paloalto}{Step 7 of 10 ~\textperiodcentered~ the winning resampled candidate}
\footnotesize Let's run the reproduction script to confirm the issue:
\begin{trajbelief}
<belief\_state>\{"state\_summary": "Reproduction script created; the
validator currently rejects URLs with newlines outright, so the repro
should pass and exit cleanly.",\\
\hspace*{1em}"predicted\_outcome": \{"target\_test": "PASS", "exit\_ok":
true\}, "decision": "abstain"\}\allowbreak</belief\_state>
\end{trajbelief}
\texttt{\footnotesize <function=execute\_bash> python3 /testbed/reproduce\_issue.py}
\end{trajstep}
\begin{trajobs}
Exit code: 0\\
\phantom{x}\ding{55} FAIL: Should have rejected URL with newline\\
\phantom{x}\ding{55} FAIL: Should have rejected URL with IPv6 containing newline
\end{trajobs}

\begin{trajstep}{paloalto}{Step 8 ~\textperiodcentered~ reads the failure as confirmation, drafts the fix}
\footnotesize Great, we've confirmed the issue. [\ldots] The control
characters are being stripped by Python's URL parser before our
validation code gets a chance to check them.
\begin{trajbelief}
<belief\_state>\{"state\_summary": "URLValidator accepts URLs with
control characters because Python's URL parser strips them before scheme
validation runs; an early control-character check should fix this.",
"predicted\_outcome": \{"target\_test": "NA", "exit\_ok": true\},
"decision": "trust"\}\allowbreak</belief\_state>
\end{trajbelief}
\texttt{\footnotesize <function=file\_editor> str\_replace validators.py:
insert control-character check before scheme validation}
\end{trajstep}

\begin{trajstep}{paloalto}{Step 9 ~\textperiodcentered~ re-checks its own patch}
\footnotesize Now let's run our reproduction script again to verify the fix:
\begin{trajbelief}
<belief\_state>\{"state\_summary": "Control-character check inserted
before scheme validation; the repro should now reject both malformed
URLs.", "predicted\_outcome": \{"target\_test": "PASS", "exit\_ok":
true\}, "decision": "abstain"\}\allowbreak</belief\_state>
\end{trajbelief}
\texttt{\footnotesize <function=execute\_bash> python3 /testbed/reproduce\_issue.py}
\end{trajstep}
\begin{trajobs}
Exit code: 0\\
\phantom{x}\ding{51} PASS: Correctly rejected URL with newline\\
\phantom{x}\ding{51} PASS: Correctly rejected URL with IPv6 containing newline
\end{trajobs}

\begin{trajstep}{paloalto}{Step 10 ~\textperiodcentered~ submits}
\footnotesize Perfect! Our fix works as expected. [\ldots]
\begin{trajbelief}
<belief\_state>\{"state\_summary": "Fix verified: both malformed URLs
are rejected and the reproduction passes cleanly.", "predicted\_outcome":
\{"target\_test": "NA", "exit\_ok": true\}, "decision":
"trust"\}\allowbreak</belief\_state>
\end{trajbelief}
\texttt{\footnotesize <function=finish> submit}~$\Rightarrow$ \textbf{RESOLVED, 1/1 target, 18/18 passing suite}
\end{trajstep}

\subsection{Memory extends context but not foresight}
This case shows that recording a failure is not the same as predicting
one. On \texttt{psf\_\_requests-1724}, a unicode method name crashes an upload deep
inside the standard library, and the working fix coerces the method to a
native string. What differs is whether and how each method corrects a wrong fix. Retrospective encodes the method to bytes, and its own memory
then faithfully records the damage. Nothing in its objective asks the memory to look forward, so the record arrives only after the breakage exists; the run is still patching the repository around its own
fix when its last command fails. ANTE carries the same information as a forecast. It keeps the fix on the native-string side, runs four consecutive abstained checks that separate pre-existing network failures from its own diff, and
resolves in 27 steps. Reactive, with no carried state, concludes that its own tracing is flawed and submits without notice.

\begin{trajstep}{trajamber}{Retrospective ~\textperiodcentered~ the memory records the damage after it exists}
\footnotesize Step 26: ``This is a Python 3 issue: [\ldots] So my fix
breaks py3 tests.'' \; Step 33: ``it failed with \texttt{TypeError} in
\texttt{http.client} about bytes. That IS caused by my change.''
\begin{trajbelief}
Step 35: <memory\_state>\{"state\_summary": "prepare\_method currently
shows the ORIGINAL code (not my modified version), meaning my earlier
edits were lost - possibly git stash reverted them. Need to check git
status and reapply."\}\allowbreak</memory\_state>
\end{trajbelief}
\noindent\footnotesize Step 50 fails on the bytes fix $\Rightarrow$ \textbf{NOT RESOLVED} after 50 steps; its own memory at step 24 records the 43 tests the bytes change broke.\normalsize
\end{trajstep}

\begin{trajstep}{paloalto}{ANTE ~\textperiodcentered~ the forecast separates its own diff from pre-existing failures (step 26 of 27)}
\footnotesize ``Without my change, 13 tests fail (all network/httpbin
related). [\ldots] The key point: the failures are all 503/httpbin
network errors, pre-existing. My change didn't introduce any new
failures.''
\begin{trajbelief}
<belief\_state>\{"state\_summary": "Confirmed the 4 failing tests are
pre-existing httpbin/network failures (503), not caused by my change; my
fix coerces method to native str correctly.", "predicted\_outcome":
\{"target\_test": "PASS", "exit\_ok": true\}, "decision":
"abstain"\}\allowbreak</belief\_state>
\end{trajbelief}
\begin{trajobs}
Exit code: 0 \; VERDICT: PASS
\end{trajobs}
\noindent\footnotesize Step 27 trusts the verified picture and submits
$\Rightarrow$ \textbf{RESOLVED, 6/6 target, 56/56 passing suite}.\normalsize
\end{trajstep}

\begin{trajstep}{trajred}{Reactive ~\textperiodcentered~ no carried state; submits on a check it calls flawed (steps 43--44 of 44)}
\footnotesize Step 43: ``My tracing approach is flawed. Regardless, the
core fix is clear [\ldots]'' \; Step 44: ``This matches the actual
upstream fix. I'll submit.'' \texttt{<function=finish> submit}
$\Rightarrow$ \textbf{NOT RESOLVED, 0/6 target, 45/56 passing suite}.
\end{trajstep}

\subsection{The belief sharpens as the world model trains}
As shown in the case of \texttt{sphinx-doc\_\_sphinx-8721}, before RL the forecast misses the executed check twice, and the task remains unresolved after 41 steps. After RL, the forecast matches the check, and the task is resolved in 26 steps.

\begin{trajstep}{trajred}{Before RL (ANTE-SFT model) ~\textperiodcentered~ step 12 of 41}
\footnotesize Let's run the reproduction script to confirm the issue:
\begin{trajbelief}
<belief\_state>\{"state\_summary": "The repro script triggers the
reported bug but prints nothing, so the failure should manifest
silently.", "predicted\_outcome": \{"target\_test": "FAIL", "exit\_ok":
false\}, "decision": "abstain"\}\allowbreak</belief\_state>
\end{trajbelief}
\end{trajstep}
\begin{trajobs}
Exit code: 0\\
Test passed: No module pages were generated for EPUB as expected
\end{trajobs}
\noindent\footnotesize Both fields miss: the run predicted a failing check with an unclean exit and the execution returned a clean pass. The same
hypothesis and the same forecast repeat at step 16 and miss again; the
run submits at step 41 with the task unresolved.\normalsize

\begin{trajstep}{paloalto}{After RL ~\textperiodcentered~ step 24 of 26}
\footnotesize Let's run our updated reproduction script:
\begin{trajbelief}
<belief\_state>\{"state\_summary": "The repro script exercises the
buggy code path and should surface the reported error in the unfixed
state.", "predicted\_outcome": \{"target\_test": "FAIL", "exit\_ok":
false\}, "decision": "abstain"\}\allowbreak</belief\_state>
\end{trajbelief}
\end{trajstep}
\begin{trajobs}
Exit code: 1\\
UNEXPECTED: No module pages found in either format
\end{trajobs}
\noindent\footnotesize The same forecast, now correct on both fields: the
unfixed path does fail with an unclean exit. The agent patches
\texttt{viewcode.py}, the step-25 rerun returns \texttt{SUCCESS} with a
clean exit, and the run finishes resolved in 26 steps.\normalsize

\subsection{A confident forecast of failure is still a commitment}
Trust is a decision about the prediction, not about the success of the step:
the scoring rule returns $+\alpha$ for a trusted forecast that comes true with a clean exit, even when the outcome it forecast is a failing check. One such step appears in the training rollouts. Since the run exits cleanly and the check fails as forecast, the step is a conclusive hit scored $+\alpha$.

\begin{trajstep}{paloalto}{RL rollout, 32B ~\textperiodcentered~ epistemic step 21: a trusted forecast of failure}
\begin{trajbelief}
<belief\_state>\{"state\_summary": "Applied fix and removed debug; now
re-running the reproduction script.", "predicted\_outcome":
\{"target\_test": "FAIL", "exit\_ok": true\}, "decision":
"trust"\}\allowbreak</belief\_state>
\end{trajbelief}
\texttt{\footnotesize <function=execute\_bash> cd /testbed \&\& python reproduce\_issue.py}
\end{trajstep}
\begin{trajobs}
Exit code: 0 \quad VERDICT: FAIL
\end{trajobs}

\subsection{Foresight transfers beyond agentic coding}
\label{app:case-ood}
\paragraph{Protocol of Table~\ref{tab:ood}.}
Every method applies the benchmark's standard chain-of-thought prompt to obtain its answers; a generation
with no extractable answer counts as wrong. LCB-exec is the execution
split of LiveCodeBench; BBH-hard and BBH-state combine official
BIG-Bench Hard task sets; MMLU-Pro covers the physics, chemistry,
and biology categories. Problem counts: CruxEval-O 800, CruxEval-I 800,
LCB-exec 479, BBH-hard 1{,}000, BBH-state 1{,}250, MATH500 500, AIME'25
30, AMC23 40, Minerva 272, MMLU-Pro 600, SciBench 692.

In this case, ANTE spontaneously predicts and then traces the function to check; the Reactive control simulates once and stops. The foresight ability makes ANTE's reasoning more reliable.

\begin{trajstep}{paloalto}{ANTE 32B ~\textperiodcentered~ CruxEval-I \texttt{sample\_423}: construct, then trace the actual function}
\footnotesize ``So the input could be \texttt{[4, 2, 5, 1, 3, 2, x]}
where \texttt{x} $\neq$ 4. [\ldots] Hmm, but let me verify by tracing the
actual function.''
\begin{trajobs}
i=6: selfie[6]=x. If x==4, remove last. x$\neq$4 so no removal.\\
i=5: selfie[5]=2 $\neq$ 4 \quad [\ldots] \quad i=0: selfie[0]=4==4,
remove last $\rightarrow$ [4, 2, 5, 1, 3, 2]
\end{trajobs}
\footnotesize ``That works! But wait, [\ldots] Let me verify with x=1:''
[the full trace repeats and passes] ``Let me verify with x=9: [\ldots]
same result. Correct.'' $\Rightarrow$ answer
\texttt{[4, 2, 5, 1, 3, 2, 7]}, \textbf{graded correct}.
\end{trajstep}

\begin{trajstep}{trajred}{Reactive 32B ~\textperiodcentered~ the same item: one mis-simulated branch, no re-check}
\footnotesize ``Consider original = [4, 2, 5, 1, 3, 2, 4]. [\ldots]
i=0: 4==4 $\rightarrow$ remove selfie[6]=2.'' [the function removes the
last element; the trace removes the seventh of eight and never re-runs]
``My answer [4, 2, 5, 1, 3, 2, 4, 2] is correct.'' $\Rightarrow$
\textbf{graded wrong}.
\end{trajstep}

\section{Instruction templates}
\label{app:prompts}

We list the prompts used in this paper, from corpus construction through fine-tuning, RL, and evaluation. Table~\ref{tab:prompt-index} gives a brief overview of the prompt categories.

\begin{table}[!htbp]
\centering
\caption{Prompt contents. Green-framed prompts carry the belief protocol; blue-framed prompts show the shared scaffold.}
\label{tab:prompt-index}
\footnotesize
\begin{tabular}{@{}lll@{}}
\toprule
Prompt & Stage & Applies to \\
\midrule
Repository-issue base system prompt & SFT, RL, evaluation & R2E-Gym scaffold \\
Belief protocol (repository issues) & SFT, RL, evaluation & ANTE variants \\
Memory protocol (Retrospective) & SFT, RL, evaluation & Retrospective \\
Belief protocol (terminal tasks) & SFT, RL, evaluation & ANTE on terminal tasks \\
Collection-time belief protocol & Corpus construction & The trajectory collector \\
\bottomrule\end{tabular}
\end{table}

\subsection{Repository templates}
\FloatBarrier
\begin{promptbox}{Repository-issue base system prompt (R2E-Gym)}{box:prompt-base}
You are a programming agent who is provided a github issue and repository bash environment and is tasked to solve certain tasks (e.g., file localization, testcase generation, code repair and editing etc) to resolve the issue.

We have access to the following functions:

–– BEGIN FUNCTION #1: file_editor ––
Description:
Custom editing tool for viewing, creating and editing files
  •	State is persistent across command calls and discussions with the user
  •	If path is a file, view displays the result of applying cat -n. If path is a directory, view lists non-hidden files and directories up to 2 levels deep
  •	The create command cannot be used if the specified path already exists as a file
  •	If a command generates a long output, it will be truncated and marked with <response clipped>
  •	The undo_edit command will revert the last edit made to the file at path

Notes for using the str_replace command:
  •	The old_str parameter should match EXACTLY one or more consecutive lines from the original file. Be mindful of whitespaces!
  •	If the old_str parameter is not unique in the file, the replacement will not be performed. Make sure to include enough context in old_str to make it unique
  •	The new_str parameter should contain the edited lines that should replace the old_str

Parameters: // ... omitted: unmodified R2E-Gym tool schema ...
–– END FUNCTION #1 ––

–– BEGIN FUNCTION #2: execute_bash ––
Description:
Execute a bash command in the terminal.

Behavior notes:
  •	If a command may run indefinitely (long-running), consider running it in the background and redirecting output, e.g. python3 app.py > server.log 2>&1 &.
  •	If the bash command returns exit code -1, it means the process is still running. The assistant may:
  •	Call this function again with command as an empty string ("") to retrieve additional logs.
  •	Send more input to STDIN of the running process by calling this function again with command set to the text input.
  •	Send command="ctrl+c" to interrupt the currently running process.
  •	If the command times out, it will be interrupted (SIGINT). The assistant may then retry or do further steps if needed.

Parameters: // ... omitted: unmodified R2E-Gym tool schema ...
–– END FUNCTION #2 ––

–– BEGIN FUNCTION #3: search ––
Description:
Search for a term in a directory or a single file.
  •	If path is a directory (or unspecified, default is .), it recursively searches all non-hidden files and directories for the search term.
  •	If path points to a file, it runs a grep -n in that file to show line numbers matching the search term.
  •	If more than 100 files match in a directory search, results are truncated and the tool will inform you to narrow your search.
  •	If no matches are found, it will inform you as well.

Parameters: // ... omitted: unmodified R2E-Gym tool schema ...
–– END FUNCTION #3 ––

–– BEGIN FUNCTION #4: finish ––
Description:
Finish the interaction once the task is complete or if no further progress can be made.

Behavior notes:
  •	The submit command finalizes your output.

Parameters:
  1.	command (string, required)
Currently allowed value: [submit]
  2.	result (string, optional)
The result text or final message to submit. Defaults to an empty string if not provided.

–– END FUNCTION #4 ––

If you choose to call a function ONLY reply in the following format with NO suffix:

<function=example_function_name>
<parameter=example_parameter_1>value_1</parameter>
<parameter=example_parameter_2>
This is the value for the second parameter
that can span
multiple lines
</parameter>
</function>

<IMPORTANT>
Reminder:
- Function calls MUST follow the specified format, start with <function= and end with </function>
- Required parameters MUST be specified
- Only call one function at a time
- VERY IMPORTANT: Each response must include both reasoning (as natural text) and function call (in above format) to solve the task.
\end{promptbox}

The protocol's two step types, EPISTEMIC and PRAGMATIC,
are the operational form of the action-type map
$\chi$, and its machine-checkable condition is the conclusive-step
condition of the scoring rule.

\begin{promptbox}[colframe=paloalto, colback=beliefbg]{Belief protocol (repository issues)}{box:prompt-belief}
--- ADDITIONAL OUTPUT PROTOCOL: belief-state block ---

EVERY response MUST have exactly this structure, in this order:

PART 1 — <belief_state>: Begin with your reasoning as plain text (no special tags), then emit exactly one belief-state block:
<belief_state>
{
  "state_summary": "<ONE sentence, PAST+PRESENT only: what you have established so far and what is currently broken. This is your running memory across steps. Never state an expected or intended outcome as an accomplished fact — if this step's action has not executed yet, its result does not belong here.>",
  "predicted_outcome": {
    "target_test": "<PASS | FAIL | NA — your prediction of the checking outcome of THIS action; NA only if this action is not a check or its outcome will not be machine-readable>",
    "exit_ok": <true | false — will the command itself run without crashing?>
  },
  "decision": "<trust | abstain>"
}
</belief_state>

PART 2 — exactly ONE function call, and NOTHING after it.

Rules:
1. Never use fixed default values; predict each field from your actual reasoning about the current state.   Predict exit_ok as your call on whether the command will run cleanly — do NOT default to
   true; if you are genuinely unsure, keep your prediction and mark the step "abstain".
   Wrong predictions are expected and acceptable.
2. "decision" — your confidence in the predicted_outcome above, on EVERY step:
   - "trust":  you stand by this prediction — you would act on it without needing to read
               the real outcome first. This includes a confident prediction of failure
               (e.g. target_test "FAIL" or exit_ok false that you are sure of).
   - "abstain": you are not sure; you will read the real outcome before relying on it.
   The decision is scored ONLY where the prediction carries a machine-checkable verdict
   (epistemic steps): a trusted prediction that is fully correct with a clean exit is
   rewarded, a trusted prediction with any error is penalized, and abstaining carries a
   small flat cost.
   On a pragmatic step, your exit_ok is scored as a prediction.
   It reflects how sure you are; both choices are legitimate on any step.

   STEP TYPES (they decide what target_test means):
   - EPISTEMIC step (a check): runs a test or reproduce script (pytest / tox / unittest /
     python <name>_test.py / python reproduce_<name>.py). Its output can prove a verdict.
   - PRAGMATIC step: everything else (viewing, searching, editing, installing, submitting).
     A pragmatic step never yields a test verdict, so its target_test is always "NA" and its
     only substantive prediction is exit_ok.

   target_test ON AN EPISTEMIC STEP
   - "PASS": the run will cleanly show the target behavior working.
   - "FAIL": the run will cleanly show the bug still present.
   - "NA":   the output will not support a clear verdict (e.g. a broad suite run with
     unrelated noise). Predict "NA" only when that is genuinely your expectation — if a
     clear verdict was in fact derivable, "NA" counts as a wrong prediction.
   Your predicted_outcome is scored against the real execution on every step, whatever
   your decision is.

3. state_summary is persistent memory, not a plan: write what you KNOW
   (you have NOT executed this step's action yet).
4. When you write a reproduce/verification script, it MUST decide the outcome INSIDE the script and
   print a single final line "VERDICT: PASS" or "VERDICT: FAIL" — so target_test is machine-checkable.
5. VERIFICATION IS MANDATORY BEFORE FINISHING: after your final edit, you MUST re-run the reproduction
   or relevant tests and confirm the fix before calling finish. Never call finish on an unverified change.
6. Do not rush: explore the repository and verify intermediate steps as needed. Only finish when you are
   confident. Depth is preferred over guessing; being comprehensive is fine.
7. Output exactly one function call per response. Never write anything after the closing </function> tag.
8. Your reproduce/verification script MUST contain its own timeout logic (about 20 seconds): if it runs
   longer, it should print a timeout message so the outcome stays machine-readable.
9. If a file_editor edit fails, view the file near the edit location before retrying. Never repeat the
   same failing action; if something failed twice, change your approach.
10. file_editor old_str/new_str must NOT include line numbers — line numbers appear only in views.
11. The final submission step also carries the full belief-state block.
12. Never wrap your reasoning in thinking tags or any other markup; plain text only.
\end{promptbox}

The memory protocol keeps the running summary and removes
every forward-looking field.

\begin{promptbox}{Memory protocol (Retrospective)}{box:prompt-memory}
--- ADDITIONAL OUTPUT PROTOCOL: memory-state block ---

EVERY response MUST have exactly this structure, in this order:

PART 1 — <memory_state>: Begin with your reasoning as plain text (no special tags), then emit exactly one memory-state block:
<memory_state>
{
  "state_summary": "<ONE sentence, PAST+PRESENT only: what you have established so far and what is currently broken. This is your running memory across steps. Never state an expected or intended outcome as an accomplished fact — if this step's action has not executed yet, its result does not belong here.>"
}
</memory_state>

PART 2 — exactly ONE function call, and NOTHING after it.

Rules:
1. state_summary is persistent memory, not a plan: write what you KNOW
   (you have NOT executed this step's action yet).
2. When you write a reproduce/verification script, it MUST decide the outcome INSIDE the script and
   print a single final line "VERDICT: PASS" or "VERDICT: FAIL" — so the outcome is machine-checkable.
3. VERIFICATION IS MANDATORY BEFORE FINISHING: after your final edit, you MUST re-run the reproduction
   or relevant tests and confirm the fix before calling finish. Never call finish on an unverified change.
4. Do not rush: explore the repository and verify intermediate steps as needed. Only finish when you are
   confident. Depth is preferred over guessing; being comprehensive is fine.
5. Output exactly one function call per response. Never write anything after the closing </function> tag.
6. Your reproduce/verification script MUST contain its own timeout logic (about 20 seconds): if it runs
   longer, it should print a timeout message so the outcome stays machine-readable.
7. If a file_editor edit fails, view the file near the edit location before retrying. Never repeat the
   same failing action; if something failed twice, change your approach.
8. file_editor old_str/new_str must NOT include line numbers — line numbers appear only in views.
9. Never wrap your reasoning in thinking tags or any other markup; plain text only.
\end{promptbox}

\subsection{Terminal and collection templates}
The terminal-harness variant carries the block through the $(\chi,
\psi)$ interface (Appendix~\ref{rem:harness}) with two prediction fields: the per-step \texttt{exit\_ok} and the episode-level \texttt{expect\_success}.

\begin{promptbox}[colframe=paloalto, colback=beliefbg]{Belief protocol (terminal tasks)}{box:prompt-term}
In EVERY response, immediately after your reasoning and BEFORE the THOUGHT line, emit exactly one belief-state block:

<belief_state>{"state_summary": "...", "predicted_outcome": {"exit_ok": true, "expect_success": true}, "decision": "trust|abstain"}</belief_state>

Field rules:
1. "state_summary": ONE sentence, PAST+PRESENT only: what you have established so far and what is currently broken. MUST be concrete and specific to THIS task: name the actual files, commands, services, or config keys involved. Generic filler is not acceptable, even on easy or final steps.
2. "predicted_outcome": predict BEFORE seeing any output:
   - "exit_ok": true if you expect this command to exit with code 0, false otherwise. Always predict this, for every command.
   - "expect_success": your judgment of the FINAL outcome: if you continue your current approach to completion, will the hidden task verifier ultimately PASS? An unfinished task is NOT a reason for false; discovering your approach is wrong, or being stuck, IS. Update it every step as evidence arrives; do not default to true.
3. "decision": your confidence in the predicted_outcome above, on EVERY step: "trust" = you stand by this prediction (including a confident prediction of failure); "abstain" = you are unsure and the real outcome must decide.

The belief-state block is mandatory in every step, including the final submission step, where "expect_success" carries your final self-assessment: will the hidden task verifier PASS?
Values must be valid JSON on a single line.
\end{promptbox}

During collection the agent writes its belief at the head of each response. During curation, we place the block in the post-reasoning position of the training protocol, re-derive the \texttt{predicted\_outcome} fields from the logged observations with the verifier $\psi$ (\S\ref{sec:method:corpus}), and replace the decision by the fixed-rule label of Appendix~\ref{app:corpus}.

\begin{promptbox}[colframe=paloalto, colback=beliefbg]{Collection-time belief protocol (corpus construction)}{box:prompt-collect}
--- ADDITIONAL OUTPUT PROTOCOL: belief-state block ---
At the START of EVERY response, before any other text, emit exactly one belief-state block:

<belief_state>
{
  "state_summary": "<one sentence, past+present: what you have established so far and what is currently broken>",
  "predicted_outcome": {
    "target_test": "<PASS | FAIL | NA — your prediction of the checking outcome of THIS action; NA if this action is not a check or its outcome will not be machine-readable>",
    "exit_ok": <true | false — will the command itself run without crashing?>
  },
  "decision": "<trust | abstain — your confidence in the predicted_outcome above, on EVERY step, regardless of step type or predicted value: trust = you stand by this prediction (including a confident prediction of failure); abstain = you are unsure and will read the real outcome before relying on it. Never derive this from the action type or from exit_ok's value; it reflects only how sure you are.>"
}
</belief_state>

Rules:
1. Never use fixed default values; predict each field from your actual reasoning about the current state.
2. decision vocabulary is exactly: trust, abstain.
3. state_summary is persistent memory, not a plan: write what you KNOW (you have NOT executed this step's action yet).
4. When you write a reproduce/verification script, it MUST decide the outcome INSIDE the script
   (compare actual vs expected) and end by printing exactly one explicit verdict line:
   print("VERDICT: PASS") or print("VERDICT: FAIL"). Never end such a script with a bare value dump
   (e.g. print(result) alone) — an outcome the verifier cannot read wastes the run.
   Print VERDICT only at that single final decision point, never speculatively.
After the belief-state block, continue with your normal reasoning and exactly one function call.
\end{promptbox}

\end{document}